\documentclass{article} 
\usepackage{iclr2026_conference,times}

\usepackage{amsmath,amsfonts,bm}

\def\eqref#1{equation~\ref{#1}}

\def\1{\bm{1}}

\DeclareMathAlphabet{\mathsfit}{\encodingdefault}{\sfdefault}{m}{sl}
\SetMathAlphabet{\mathsfit}{bold}{\encodingdefault}{\sfdefault}{bx}{n}

\usepackage{hyperref}
\usepackage{url}

\usepackage[utf8]{inputenc} 
\usepackage[T1]{fontenc}    
\usepackage{hyperref}       
\usepackage{url}            
\usepackage{booktabs}       
\usepackage{amsfonts}       
\usepackage{nicefrac}       
\usepackage{microtype}      
\usepackage{xcolor}         
\usepackage{amsfonts}       
\usepackage{amsmath}
\usepackage{amssymb}
\usepackage{graphicx, subfig}
\usepackage{nicefrac} 
\usepackage[ruled, vlined, linesnumbered]{algorithm2e}
\usepackage{enumerate}
\usepackage{enumitem}
\usepackage{geometry}
\usepackage{tikz}
\usetikzlibrary{shapes.geometric}
\usepackage{arydshln}
\usepackage{xspace}
\usepackage{amssymb}
\usepackage{bbding}
\usepackage{tcolorbox}
\usepackage{multirow}
\usepackage{threeparttable}
\usepackage{soul}
\usepackage{color}

\usepackage[capitalize,noabbrev]{cleveref}
\usepackage{makecell}
\usepackage{wrapfig}
\usepackage{caption}
\usepackage{subcaption}
\usepackage{pifont}
\usepackage{tikz}
\usepackage{hyperref}
\usepackage[thicklines]{cancel}
\usepackage{algorithmic}
\usepackage{float}
\usepackage{bookmark}
\usepackage{titletoc}

\title{GDGB: A Benchmark for Generative Dynamic Text-Attributed Graph Learning}

\author{
  Jie Peng$^{1}$ \quad
  Jiarui Ji$^{1}$ \quad
  Runlin Lei$^{1}$ \quad 
  \textbf{Zhewei Wei}$^{1}\thanks{Zhewei Wei and Yongchao Liu are the corresponding authors.}$ \quad
  \textbf{Yongchao Liu}$^{2*}$ \quad
  \textbf{Chuntao Hong}$^{2}$
  \\[1.5ex] 
  $^{1}$Renmin University of China \quad
  $^{2}$Ant Group
  \\[1ex] 
  \texttt{\{peng\_jie, jijiarui, runlin\_lei, zhewei\}@ruc.edu.cn} \\
  \texttt{\{yongchao.ly, chuntao.hct\}@antgroup.com}  
}

\iclrfinalcopy 
\begin{document}

\maketitle

\setcounter{footnote}{0}
\begin{abstract}

Dynamic Text-Attributed Graphs (DyTAGs), which intricately integrate structural, temporal, and textual attributes, are crucial for modeling complex real-world systems.
However, most existing DyTAG datasets exhibit poor textual quality, which severely limits their utility for generative DyTAG tasks requiring semantically rich inputs.
Additionally, prior work mainly focuses on discriminative tasks on DyTAGs, resulting in a lack of standardized task formulations and evaluation protocols tailored for DyTAG generation.
To address these critical issues, we propose \underline{G}enerative \underline{D}yTA\underline{G} \underline{B}enchmark (GDGB), which comprises eight meticulously curated DyTAG datasets with high-quality textual features for both nodes and edges, overcoming limitations of prior datasets.
Building on GDGB, we define two novel DyTAG generation tasks: Transductive Dynamic Graph Generation (TDGG) and Inductive Dynamic Graph Generation (IDGG). 
TDGG transductively generates a target DyTAG based on the given source and destination node sets, while the more challenging IDGG introduces new node generation to inductively model the dynamic expansion of real-world graph data. 
To enable holistic evaluation, we design multifaceted metrics that assess the structural, temporal, and textual quality of the generated DyTAGs.
We further propose GAG-General, an LLM-based multi-agent generative framework tailored for reproducible and robust benchmarking of DyTAG generation.
Experimental results demonstrate that GDGB enables rigorous evaluation of TDGG and IDGG, with key insights revealing the critical interplay of structural and textual features in DyTAG generation. 
These findings establish GDGB as a foundational resource for advancing generative DyTAG research and unlocking further practical applications in DyTAG generation. 
The dataset and source code are available at \url{https://github.com/Lucas-PJ/GDGB-ALGO}.

\end{abstract}

\section{Introduction}
\label{sec:introduction}
Dynamic graph-structured data is ubiquitous, spanning various domains, such as social networks \citep{huang2022ttergm, sun2022aligning}, recommendation systems \citep{zhang2022dynamic, tang2023dynamic}, and citation networks \citep{skarding2021foundations, geng2022modeling}. 
Notably, real-world dynamic evolution scenarios inherently contain rich (textual) features. 
For example, user posts and interactions such as comments and reposts on social networks, and consumer reviews of products on e-commerce platforms. 
This motivates the construction of dynamic text-attributed graphs (DyTAGs), which systematically integrate structural/temporal dynamics and textual attributes \citep{DTGB}.


Existing representative DyTAG benchmarks, such as DTGB \citep{DTGB}, demonstrate that state-of-the-art Dynamic Graph Neural Networks (DGNNs) achieve significant performance gains in discriminative tasks (e.g., link prediction, node retrieval, and edge classification) by leveraging DyTAGs' textual features. 
Growing interest in generative tasks spans computer vision (CV) \citep{wang2021generative, raut2024generative} and natural language processing (NLP) \citep{xu2024large, blease2024generative}, with increasing demand for graph generation in domains such as drug discovery \citep{luo20213d, martinelli2022generative}, graph augmentation \citep{ding2022data, meng2023generative}, and privacy-preserving frameworks \citep{miao2023graph, hetemporal2025}. However, DyTAG generation remains underexplored.
For instance, DTGB focuses narrowly on simplistic textual relation generation while neglecting the rich interplay of textual, structural, and temporal dynamics in graph evolution \citep{DTGB}. 
This gap underscores the increasing criticality of generating high-quality DyTAGs.

The primary obstacle hindering progress in DyTAG generation lies in the absence of high-quality, text-rich DyTAG benchmarks. 
Current datasets and benchmarks face two critical limitations:
\emph{\textbf{1)} Lack of high-quality (textual) attributes}:
Traditional dynamic graph datasets fundamentally lack node/edge features, relying solely on topological and temporal information \citep{DGB}. 
This scarcity of attributes poses substantial challenges for generative models that require rich feature inputs. 
Meanwhile, existing DyTAG datasets from DTGB exhibit poor textual quality, with (source) node texts typically limited to usernames or email addresses, lacking semantic richness \citep{DTGB}, which severely restricts the development of DyTAG learning.
\emph{\textbf{2)} Lack of standardized DyTAG generative task formulations and evaluation protocols}:
Current dynamic graph generative models mainly rely on structural and temporal information \citep{TIGGER,G2A2}, necessitating the development of new task formulations and holistic evaluation metrics tailored to DyTAG's textual characteristics. 
Furthermore, most approaches prioritize direct generation of the final target graph \citep{Dymond,VRDAG}, diverging from the incremental and expansive growth patterns observed in real-world dynamic graph generation scenarios \citep{GAG}. 
Hence, to address these issues, we propose the \textbf{G}enerative \textbf{D}yTA\textbf{G} \textbf{B}enchmark (\textbf{GDGB}) from the following three aspects: 1) dataset construction, 2) task \& metric definition, and 3) generative framework design.



\textbf{Datasets.} Our proposed GDGB comprises eight carefully selected and rigorously processed DyTAG datasets covering domains such as e-commerce recommendations, social networks, biographies of celebrities, citation networks, and movie collaboration networks. 
All datasets include nodes and edges endowed with rich semantic textual information. 
Thus, the newly proposed datasets resolve the key problems associated with previous dynamic graph datasets, such as poor quality of textual features, which subsequently 
results in the inability to support DyTAG generation tasks.



\textbf{Tasks \& Metrics.}
Based on our GDGB datasets, we introduce two novel DyTAG generation tasks: \emph{Transductive Dynamic Graph Generation (TDGG)} and \emph{Inductive Dynamic Graph Generation (IDGG)}. 
TDGG generates a target DyTAG based on the given source and destination node sets, maintaining the transductive assumption that all nodes are known as prior knowledge.
Different from TDGG, the more challenging IDGG extends the transductive setting by introducing inductive modeling of new node generation during graph evolution, thereby successfully modeling the dynamic expansion of real-world graphs.
To holistically assess DyTAG generation, we design multifaceted evaluation protocols that jointly consider topological patterns, temporal dynamics, and semantic quality, including: \emph{1) Graph Structural Metric}, \emph{2) Textual Quality Metric}, and \emph{3) Graph Embedding Metric}.

\textbf{Framework.} Given the text-rich nature of DyTAGs, we propose GAG-General, an LLM-based multi-agent framework tailored for DyTAG generation tasks, extending the implementation of GAG \citep{GAG}. 
Unlike the bipartite-graph-specific GAG, GAG-General demonstrates superior generalization across diverse DyTAG scenarios, without requiring domain-specific customization. 
Furthermore, we implement both TDGG and IDGG tasks within GAG-General, integrating the proposed holistic evaluation metrics to ensure reproducible DyTAG generation and robust benchmarking.

Experimental results of GDGB demonstrate that the proposed datasets, tasks, evaluation protocols, and GAG-General framework collectively enable robust benchmarking for DyTAG generation. 
Key findings highlight the critical interplay between structural and textual features in DyTAG generation, as well as the practical applicability of DyTAG generation in domains such as e-commerce and social network analysis. 
While the GAG-General framework achieves competitive performance in both TDGG and IDGG, the generative pipeline still warrants further refinement to address challenges in both structural fidelity and attribute richness, ultimately enabling high-quality DyTAG generation.
Our contributions could be summarized as follows:
\begin{itemize}[itemsep=4pt, topsep=0pt, partopsep=0pt, parsep=0pt]
    \item \textbf{First Generative DyTAG Benchmark.} 
    To the best of our knowledge, GDGB is the first generative DyTAG benchmark, including eight high-quality datasets tailored for DyTAG generation.
    
    \item \textbf{Novel Generative Tasks, Metric, and Framework.} 
    We introduce two novel DyTAG generation tasks, TDGG and IDGG, along with multifaceted metrics for holistic evaluation.
    Furthermore, we propose GAG-General to ensure reproducible and robust benchmarking.
    
    \item \textbf{Empirical Insights.} 
    We find that both structural and textual features are crucial for DyTAG generation, guiding future generative framework refinement and practical applications.
\end{itemize}

\section{Related Work}
\label{sec:related_work}

{\textbf{Discriminative Dynamic Graph Learning.}}
Recent years have witnessed significant advances in dynamic graph learning over discrete-time dynamic graphs (DTDGs) and continuous-time dynamic graphs (CTDGs). 
Substantial progress has been made through various DGNNs~\citep{TGN,peng2025tidformer}. 
These models demonstrate strong performance in discriminative tasks, including dynamic link prediction, node retrieval, etc.  
Concurrently, the community develops diverse benchmarks to standardize evaluation~\citep{DGB,TGB}.
Notable initiatives include DyGLib~\citep{DyGLib} benchmarking on classical dynamic graph datasets, TGB~\citep{TGB} introducing large-scale graphs for scalability challenges, and TGB-Seq~\citep{TGB-seq} focusing on complex sequential patterns. 
However, current benchmarks primarily provide basic structural-temporal information, with limited datasets containing simplistic statistical attributes for nodes or edges \citep{DTGB}. 
To address this, DTGB \citep{DTGB} proposes DyTAG datasets and incorporates textual features through BERT embeddings~\citep{BERT}, showing measurable performance gains across multiple discriminative tasks~\citep{DTGB}. 
Nevertheless, the information bottleneck caused by transformed BERT embeddings inevitably compromises original semantic richness, leaving open the challenge of effectively utilizing textual semantics for enhanced DyTAG learning.  

\textbf{Generative Dynamic Graph Learning.}  
Compared to its discriminative counterpart, generative dynamic graph learning is still in its nascent stages.
Early approaches primarily focus on DTDGs, including \citep{Dymond, taggen, G2A2}. 
Recent advances have shifted toward CTDG generation, enabling more refined temporal modeling that better aligns with real-world application needs \citep{TG-GAN, TIGGER}. 
Contemporary approaches begin exploring feature-supportive generation: VRDAG \citep{VRDAG} implements node attribute generation through graph-based variational autoencoders (VAEs), while DG-Gen \citep{DGGEN} models edge attributes via joint conditional probability distributions.
More recently, GAG \citep{GAG}, an LLM-based multi-agent framework, enables simulation-based generation for social bipartite DyTAGs.
However, this field significantly lacks standardized benchmarks for comprehensive and effective evaluation of dynamic graph generation tasks. 
First, in terms of task design, existing methods prioritize direct generation of the final target graph \citep{Dymond,taggen,VRDAG}, which deviates from the expansive growth patterns observed in real-world dynamic graph generation scenarios \citep{GAG}.
Second, the definition of evaluation metrics lacks multidimensional, holistic assessments of graph structure, temporal dynamics, and attribute coherence \citep{taggen,VRDAG,GAG}. 
Therefore, when extending generative tasks to DyTAG generation, establishing a robust benchmark tailored for DyTAG's dynamic and text-rich characteristics becomes a critical need.



\section{Proposed Datasets}
\label{sec:dataset}

To enable high-quality DyTAG generation, this task necessitates datasets where nodes and edges are accompanied by rich textual attributes and evolving structure-temporal information over time.

\begin{table}[t] \small
\vspace{-6mm}
\caption{Statistics of proposed DyTAG datasets of GDGB. See \cref{app:dataset_details} for more details.} 
\vspace{-2mm}
\centering \scalebox{0.75}{
\begin{tabular}{c|ccccccc}
    \toprule
    \textbf{Dataset} & \textbf{Nodes} & \textbf{Edges} & \textbf{Edge Labels} & \textbf{Timestamps} & \textbf{Domain} & \textbf{Text Attributes} & \textbf{Bipartite} \\
    \midrule
    \textbf{Sephora}&210,357/2,274&801,234&5&5,314& E-commerce &Node \& Edge & \ding{51}\\
    \textbf{Dianping}&158,541/88,118&1,990,409&5&745,151& E-commerce & Node \& Edge & \ding{51}\\
    \textbf{WikiRevision}&75,622/3,204&2,778,732&2&2,766,153&  Web Interaction&Node \& Edge & \ding{51}\\
    \textbf{WikiLife}&406,148/54,513&1,996,520&24&1,810& Celebrity Biography&Node \& Edge & \ding{51}\\
    \textbf{IMDB}&125,714&1,534,162&20&122& Movie Collaboration&Node \& Edge & \ding{55}\\
    \textbf{WeiboTech}&20,767&109,345&2&79,925& Social Network&Node \& Edge & \ding{55}\\
    \textbf{WeiboDaily}&66,500&354,098&2&293,662& Social Network&Node \& Edge & \ding{55}\\
    \textbf{Cora}&48,797&110,788&5&8,274& Citation&Node \& Edge & \ding{55}\\
\bottomrule
\end{tabular}}
\label{tab:datasets}
\vspace{-5mm}
\end{table}

\begin{wrapfigure}{r}{0.39\textwidth}  
  \vspace{-1mm}
  \centering
  \includegraphics[width=1.0\linewidth]{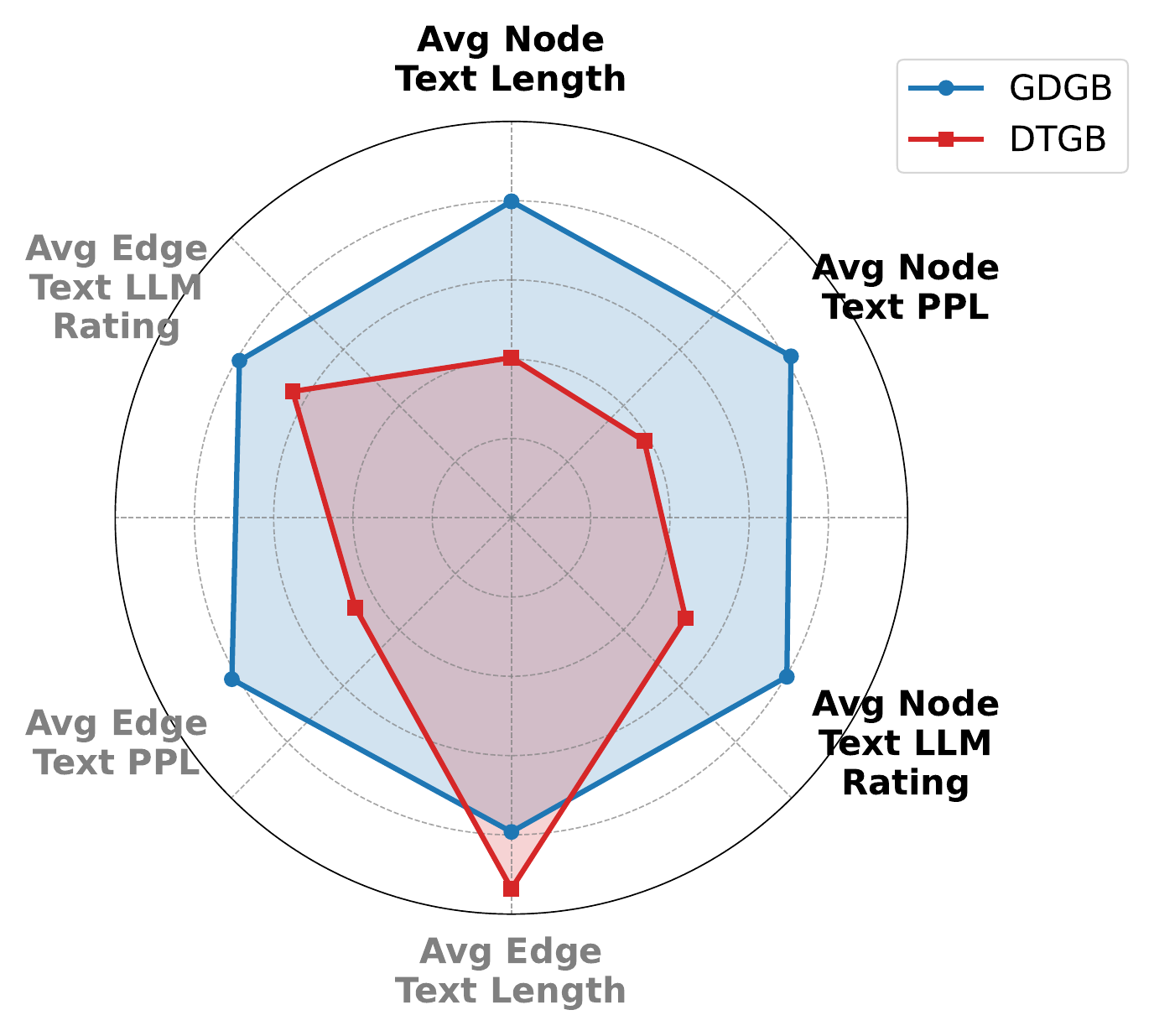}
  \vspace{-7mm}
  \caption{Comparison of node and edge texts across GDGB and DTGB datasets in terms of length, perplexity (PPL (Reversed)), and LLM-based rating.}
  \label{fig:gdgb_dtgb_radar}
  \vspace{-5mm}
\end{wrapfigure}

\textbf{Limitations of Existing Datasets.}
Existing dynamic graph datasets, though valuable for structural and temporal modeling, suffer from critical limitations in textual attributes that hinder high-quality DyTAG generation. 
DTGB \citep{DTGB} represents the first effort to incorporate both node and edge texts. 
However, as shown in \cref{fig:gdgb_dtgb_radar}, the node and edge texts in DTGB are notably short and semantically shallow in terms of length, perplexity (PPL), and LLM-based rating. 
GDGB achieves significant quality advantages over DTGB in five out of six dimensions (see detailed results for each dimension in \cref{app:comparison_with_dtgb}).
For instance, in six of DTGB’s eight datasets, (source) node texts are typically mere identifiers such as email addresses or usernames (\cref{fig:node_edge_text_len}, left). 
While the two Stack-platform datasets attempt to enrich node texts with user locations and bios, half of these fields remain empty or meaningless. 
Moreover, edge texts in datasets like GDELT and ICEWS1819 are excessively brief (\cref{fig:node_edge_text_len}, right), providing insufficient context for text generation. 
Despite DTGB having longer average edge text length, the semantic quality of DTGB's texts—as indicated by PPL and LLM-based rating—is significantly lower than GDGB's.
These limitations of current datasets impede progress in DyTAG generation research. 
A comprehensive comparison of additional existing datasets is provided in \cref{app:comparision_dataset}.

\textbf{Our Proposed High-Quality DyTAG Datasets.}
To overcome the aforementioned textual quality issues and enable DyTAG generation tasks, we propose Generative DyTAG Benchmark, GDGB, consisting of eight carefully curated DyTAG datasets. 
These datasets are sourced from diverse real-world domains, including e-commerce, social networks, biography networks, etc., encompassing both bipartite (4 datasets) and non-bipartite (4 datasets) graphs. 
Detailed statistics are presented in \cref{tab:datasets}. 
A key focus during the collection and preprocessing of these datasets is to ensure that all source/destination nodes and interaction edges possess rich, semantic textual attributes. 
For example, Sephora includes: 1) user node texts detailing user appearances and review histories, 2) product node texts describing product brands, ingredients, and ratings, and 3) edge texts containing user detailed reviews. 
The textual completeness of GDGB provides a robust foundation for developing DyTAG generation models capable of producing both realistic structures and coherent text.

{\textbf{Generative Task Performance Validation.}}  
To assess the impact of textual quality on dynamic graph generation, we evaluate two latest feature-supportive models—VRDAG \citep{VRDAG} (node features) and DG-Gen \citep{DGGEN} (edge features)—using their public implementations. Following DTGB's setup \citep{DTGB}, we encode raw texts with BERT-base-uncased \citep{BERT} and measure structural fidelity via Degree/Spectra MMD \citep{MMD} to quantify distributional differences between generated and ground-truth graphs. Lower MMD indicates better structural quality when textual features are effective.
Results in \cref{fig:vrdag_degree_spectra_mmd,fig:dggen_degree_spectra_mmd} show that, on GDGB, incorporating node or edge texts significantly reduces MMD, improving structural similarity. In contrast, on DTGB, textual features degrade performance in half the cases—especially for VRDAG—due to poor text quality. This highlights inherent limitations in DTGB. Beyond these experiments, we further validate GDGB's advantages through applying our generative framework (proposed in \cref{sec:framework}) to downstream generation tasks, providing a comprehensive demonstration of the advantages and distinctions of GDGB over DTGB in \cref{app:comparison_with_dtgb}.

\section{Generative Tasks, Metrics, and Framework}  
\label{sec:framework_task_metrics}  

\begin{figure}[!t]
  \centering
  \vspace{-14mm}
  \includegraphics[width=0.78\linewidth]{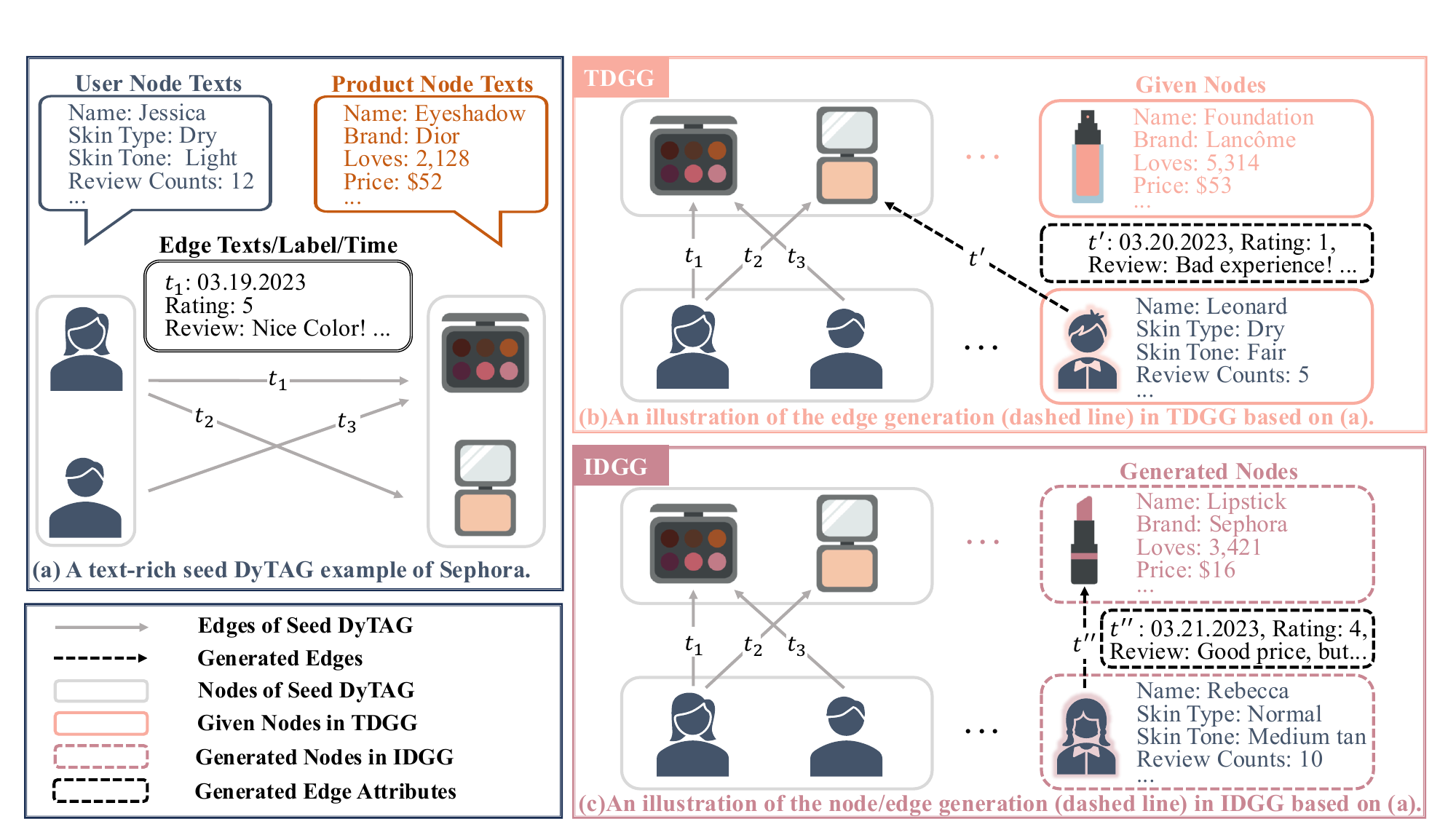}
  \vspace{-2mm}
  \caption{A case study on TDGG and IDGG in the Sephora product reviews scenario.}
  \vspace{-5mm}
  \label{fig:pipeline}
\end{figure}

\subsection{Generative Task Design}  
\label{sec:task}  
We consider a DyTAG $\mathcal{G} = (\mathcal{N}, \mathcal{E}, \mathcal{T})$, where $\mathcal{N}$ denotes the set of nodes, $\mathcal{E}$ denotes the set of edges, and $\mathcal{T}$ denotes the set of timestamps. 
Let $\mathcal{N}^{\text{text}}$, $\mathcal{E}^{\text{text}}$, and $\mathcal{L}$ represent the sets of node texts, edge texts, and edge labels, respectively. 
For any node $n \in \mathcal{N}$, its associated node text is denoted by $\mathcal{N}_n^{\text{text}}$. 
Each edge $(u,v) \in \mathcal{E}$ is attached with an edge text $\mathcal{E}_{u,v}^{\text{text}}$, an edge label $\mathcal{L}_{u,v}$, and an interaction timestamp $\mathcal{T}_{u,v}$. 
Leveraging our proposed GDGB datasets, we introduce two novel DyTAG generation tasks: Transductive Dynamic Graph Generation (TDGG) and Inductive Dynamic Graph Generation (IDGG). 

\textbf{Transductive Dynamic Graph Generation.} 
Starting with an initial seed DyTAG $\mathcal{G}_0$, TDGG generates a final DyTAG $\mathcal{G}_K$ based on the given source and destination node sets, maintaining the transductive assumption that all nodes are known as prior knowledge.
Thus, the goal of TDGG is to conduct destination node selection and edge generation, which naturally integrates traditional discriminative tasks (e.g., node retrieval and edge classification) in a generative paradigm.
The generated DyTAG $\mathcal{G}_K$ in TDGG is expected to maintain structural, temporal, and textual similarity to the ground-truth DyTAG, reflecting high-quality generation.
The illustration of TDGG is shown in \cref{fig:pipeline} (b), demonstrating that the user will review the powder at $t^{\prime}$, along with a generated negative review and a rating of 1.

\textbf{Inductive Dynamic Graph Generation.} 
Different from TDGG, the more challenging IDGG extends the transductive setting by introducing inductive modeling of new node generation during graph evolution. 
Starting with an initial seed DyTAG $\mathcal{G}_0$, IDGG generates a final DyTAG $\mathcal{G}_K$ based on the dynamically evolving source and destination node sets with newly generated nodes.
Due to the incorporation of inductive node generation, the generated DyTAG $\mathcal{G}_K$ in IDGG is expected to preserve the structural properties of the ground-truth DyTAG, while ensuring that all newly added nodes and edges contain high-quality, semantically coherent textual attributes, thereby successfully modeling the dynamic expansion of real-world graph data.
The illustration of IDGG is shown in \cref{fig:pipeline} (c), showcasing that the newly generated user will review the newly generated lipstick at $t^{\prime\prime}$ with a generated positive review and a rating of 4.


\subsection{Proposed Generative Metrics}  
\label{sec:metric}  
To ensure a comprehensive assessment of graph structure, temporal dynamics, and textual quality of the generated DyTAG on TDGG and IDGG, we propose multifaceted evaluation protocols for DyTAG generation, integrating structural, textual, and graph embedding-based metrics.
Detailed formulations and implementation specifics of the following metrics are provided in \cref{app:metrics}.

{\textbf{Graph Structural Metric.}}  
We evaluate the structural validity of generated DyTAGs using two classical approaches:
\textbf{1) Degree/Spectra MMD}: We utilize the maximum mean discrepancy (MMD) with a radial basis function (RBF) kernel to measure the distribution distance between generated and ground-truth graphs (e.g., degree/spectral properties)\citep{MMD,graphrnn,EDGE}.
\textbf{2) Power-law Analysis}: We assess power-law behavior in degree distributions using the Kolmogorov-Smirnov distance $D_k$ and the power-law exponent $\alpha$ \citep{powerlaw}. 
Specifically, we assess the presence of power law validity by evaluating whether $D_k<0.15$ and $\alpha \in[2,3]$ for the generated graphs.

{\textbf{Textual Quality Metric.}}  
Inspired by the recent role-playing agent research \citep{RoleBench,SocialBench,charbox}, we employ the \textbf{LLM-as-Evaluator} framework to assess the text quality in the generated DyTAGs. 
Specifically, the evaluation framework is conducted with five scoring criteria 
(\emph{Contextual Fidelity, Personality Depth, Dynamic Adaptability, Immersive Quality, Content Richness}) tailored for DyTAG generation, scored on a 1–5 scale. 
Compared to embedding-based metrics used in prior DyTAG benchmarks (e.g., BERTScore \citep{bertscore} in DTGB \citep{DTGB}), this evaluation framework offers two distinct advantages: 1) Unified multidimensional evaluation across diverse textual attributes, and 2) Preservation of semantic fidelity by avoiding information compression inherent in embedding representations. 

{\textbf{Graph Embedding Metric.}} 
To enable a holistic and coupled evaluation of structural, temporal, and textual quality in DyTAG generation, we extend the JL-Metric \citep{jlmetric}, designed for dynamic graph generation, to a better adaptation on DyTAGs by integrating textual node/edge features, resulting in a \textbf{graph embedding-based indicator} that quantifies the generation quality across three critical dimensions: topological patterns, temporal dynamics, and node/edge textual attributes. 
This metric condenses DyTAGs into a unified embedding space for the pairwise similarity computation between graph embeddings. 
The resulting scalar score quantifies global fidelity between generated and ground-truth DyTAGs by jointly measuring structural, temporal, and textual characteristics.

\subsection{Generative Framework}  
\label{sec:framework}  
\textbf{Limitations of Previous Methods in DyTAG Generation.}
The novelty of our proposed TDGG and IDGG tasks presents unique challenges, as no existing method can directly address their requirements. 
For instance, the latest feature-supportive dynamic graph generation models like VRDAG \citep{VRDAG} and DG-Gen \citep{DGGEN} focus solely on generating node or edge representation features, but lack the capability to handle textual contents in DyTAGs. 
As far as we know, while GAG \citep{GAG} is the only existing baseline that incorporates textual attributes, it is specifically designed for bipartite social graph generation and lacks generalizability to diverse DyTAG structures and domains. 
This gap motivates the development of a more universal framework tailored for DyTAG generation tasks.

\textbf{GAG-General.} Given the text-rich nature of DyTAGs, we propose GAG-General, an LLM-based multi-agent framework designed for DyTAG generation tasks, building upon the implementation of GAG \citep{GAG}. 
Leveraging the text understanding and generation capabilities of LLMs, our framework extends the original GAG with three key enhancements:
\textbf{1) Generalization}: Support both bipartite and non-bipartite graph structures universally.
\textbf{2) Multi-domain Compatibility}: Abstract the DyTAG generation pipeline to enable seamless adaptation across diverse interaction scenarios without domain-specific customization.
\textbf{3) Standardization}: Establish standardized task formulations for TDGG and IDGG and incorporate holistic evaluation metrics, ensuring reproducible DyTAG generation. 
See more details of GAG-General in \cref{app:gag-general}.

\textbf{Framework Details.}
Following GAG \citep{GAG}, GAG-General employs carefully designed LLM-based agents for source and destination nodes, with each node agent maintaining a memory module to record historical neighbor interactions. 
This memory module effectively captures structural and temporal dynamics from the DyTAG by preserving contextual information about past interactions. 
We further incorporate an optional memory reflection mechanism, which leverages LLMs to distill the node memories into valuable summaries, akin to the message aggregation process in GNNs \citep{kipf2016semi,peng2024beyond}. 
For TDGG and IDGG, GAG-General implements an iterative, expansive DyTAG generation pipeline: in each iteration, the source node agent selects destination nodes based on its memory and contextual features. 
In the case of IDGG, new nodes are first generated and then updated to the node set before this selection step, ensuring the DyTAG's inductive expansion. 
Subsequently, edges are generated between the selected node pairs. 
This iterative process continues until the final DyTAG is produced, seamlessly integrating node/edge generation and DyTAG evolution. 
See more details and pseudocode regarding the implementation of TDGG and IDGG based on GAG-General in \cref{app:tdgg_idgg_details}.

\section{Experiment}
\label{sec:experiment}

\subsection{Experimental Settings}

\textbf{Baselines.} \textbf{1)} As our proposed GAG-General is an LLM-based multi-agent generative framework, for the DyTAG generation in TDGG and IDGG, we benchmark three open-source LLMs: DeepSeek-R1-Distill-Qwen-32B \citep{deepseek} (referred to as DeepSeek), Llama-3-70B-Instruct \citep{llama3} (Llama), and Qwen2.5-72B-Instruct \citep{qwen2.5} (Qwen), as well as a closed-source model: GPT-4o-Mini \citep{gpt4omini} (GPT).
\textbf{2)} For TDGG, which integrates traditional discriminative tasks from DGNNs in a generative paradigm, we compare against five state-of-the-art baselines, including JODIE \citep{JODIE}, TGN \citep{TGN}, CAWN \citep{CAWN}, GraphMixer \citep{Graphmixer}, and DyGFormer \citep{DyGLib}. 
Performance is assessed on discriminative tasks such as node retrieval (link prediction) and edge classification to benchmark GAG-General against conventional DGNNs.  
\textbf{3)} For IDGG, due to its novelty, the challenge of new node generation, and the text-rich nature of DyTAGs, we select two of the latest dynamic graph generation models that support node/edge feature modeling and generation: DG-Gen \citep{DGGEN}, VRDAG \citep{VRDAG}, and {TIGGER-I \citep{TIGGER}}. See more details in \cref{app:supplementary_implement_details}.

\textbf{Implementation Details.} Both TDGG and IDGG share basic settings: the first 1,000 edges and the corresponding nodes serve as the initial seed DyTAG, and we set the edge generation size per round as 50. 
We employ the metrics introduced in \cref{sec:metric} for evaluating the DyTAG generation quality on TDGG and IDGG, including graph structural metrics, textual quality scores, and a graph embedding-based indicator for joint structural-temporal-textual assessment. 
See more implementation details in \cref{app:supplementary_implement_details}.

\subsection{Transductive Dynamic Graph Generation}

\textbf{Results of TDGG.}  
In TDGG, GAG-General generates DyTAGs on GDGB with high structural fidelity, as shown in \cref{tab:tdgg_structural_gpt}. The generated graphs exhibit small deviations from ground-truths—most Degree/Spectra MMDs are below 0.3, and six of eight satisfy the power-law validity criterion, indicating high-quality generation. To assess the role of structural and textual information from historical neighbors, we ablate three configurations: w/o node memory, w/ node memory, and w/ node memory \& reflection mechanism. As shown in \cref{tab:tdgg_textual_average_4datasets,tab:tdgg_graph_embedding_4datasets}, both node memory and the reflection mechanism significantly improve textual quality and graph embedding-based metrics across most LLM backbones, due to effective integration and aggregation of historical interaction information. 
Moreover, we experimentally verify the direct usability of generated graphs from TDGG in downstream tasks compared to the original graphs; experimental details are provided in \cref{TDGG-downstream}.
Scalability analysis for TDGG is provided in \cref{app:scalability}.


\begin{table}[!t]
\vspace{-8mm}
\centering \caption{The results on Degree MMD, Spectra MMD, $D_k$, $\alpha$, and power-law validity under TDGG with GPT as the LLM backbone. Full results on the other three LLM backbones are available in \cref{tab:tdgg_structural_deepseek}, \ref{tab:tdgg_structural_llama}, and \ref{tab:tdgg_structural_qwen}.} 
\vspace{-2mm}
\label{tab:tdgg_structural_gpt}
\resizebox{0.9\linewidth}{!}{
    \begin{tabular}{l|cccccccc}
    \toprule
        Dataset & \textbf{Sephora} & \textbf{Dianping} & \textbf{WikiRevision} & \textbf{WikiLife} & \textbf{IMDB} & \textbf{WeiboTech} & \textbf{WeiboDaily} & \textbf{Cora} \\
        \midrule
        Degree MMD$\downarrow$        & 0.023 & 0.055 & 0.108 & 0.181 & 0.278 & 0.243 & 0.247 & 0.128 \\
        Spectra MMD$\downarrow$       & 0.011 & 0.328 & 0.156 & 0.223 & 0.316 & 0.297 & 0.493 & 0.156 \\
        $D_k$            & 0.143 & 0.041 & 0.056 & 0.099 & 0.135 & 0.030 & 0.048 & 0.049 \\
        $\alpha$         & 2.993 & 2.234 & 2.041 & 2.204 & 1.720 & 2.011 & 1.845 & 2.378 \\
        Power-law Validity& \ding{51}     & \ding{51}     & \ding{51}     & \ding{51}     & \ding{55}     & \ding{51}     & \ding{55}     & \ding{51} \\
    \bottomrule
    \end{tabular}}
\vspace{-2.5mm}
\end{table}

\begin{table}[!t]
\centering \caption{The results on average textual quality scores under TDGG. M. and R. denote node memory and reflection mechanism, respectively. Full results on each scoring criterion are available in \cref{tab:tdgg_textual_average}-\ref{tab:tdgg_textual_cora}. 
{The LLM (GPT) used in evaluation is independent of the LLM backbone used by GAG-General itself. See \cref{app:Implementation Details} for more details.}
The best and the runner-up scores are highlighted in bold and underlined fonts.
} 
\vspace{-2mm}
\label{tab:tdgg_textual_average_4datasets}
\resizebox{1.0\linewidth}{!}{
\begin{tabular}{l|ccc|ccc|ccc|ccc}\toprule
 & \multicolumn{3}{c|}{\textbf{DeepSeek}} & \multicolumn{3}{c|}{\textbf{Llama}} & \multicolumn{3}{c|}{\textbf{Qwen}} & \multicolumn{3}{c}{\textbf{GPT}}\\
 & \multicolumn{1}{c}{w/o M.} & \multicolumn{1}{c}{w/ M.} & \multicolumn{1}{c|}{w/ M.R.} & \multicolumn{1}{c}{w/o M.} & \multicolumn{1}{c}{w/ M.} & \multicolumn{1}{c|}{w/ M.R.} & \multicolumn{1}{c}{w/o M.} & \multicolumn{1}{c}{w/ M.} & \multicolumn{1}{c|}{w/ M.R.} & \multicolumn{1}{c}{w/o M.} & \multicolumn{1}{c}{w/ M.} & \multicolumn{1}{c}{w/ M.R.} \\ 
 \midrule
Sephora& 4.09 & \underline{4.10} & \textbf{4.37} & 4.57 & \underline{4.58} &\textbf{4.66} & 4.61 & \underline{4.64} & \textbf{4.70} & 4.69 & \underline{4.69} & \textbf{4.77} \\
{Dianping} & {4.29} & {\underline{4.34}} & {\textbf{4.41}} & {4.13} & {\underline{4.18}} & {\textbf{4.46}} & {\underline{4.14}} & {4.13} & {\textbf{4.43}} & {4.32} & {\underline{4.34}} & {\textbf{4.71}} \\
{IMDB} & {3.65} & {\underline{3.82}} & {\textbf{3.99}} & {3.97} & {\underline{4.02}} & {\textbf{4.32}} & {4.10} & {\underline{4.18}} & {\textbf{4.33}} & {3.91} & {\underline{4.02}} & {\textbf{4.44}} \\
WeiboTech& 3.88 & \underline{3.89} & \textbf{3.92} & 4.49 & \underline{4.56} & \textbf{4.86} & \underline{4.93} & 4.91 & \textbf{4.96} & 4.84 & \underline{4.88} & \textbf{4.97} \\
\bottomrule
\end{tabular}}
\vspace{-3mm}
\end{table}

\begin{table}[!t]
\vspace{-0mm}
\centering \caption{The results on the graph embedding metric under TDGG. See \cref{tab:tdgg_graph_embedding} for full results.} 
\vspace{-2mm}
\label{tab:tdgg_graph_embedding_4datasets}
\resizebox{1.0\linewidth}{!}{
\begin{tabular}{l|ccc|ccc|ccc|ccc}\toprule
 & \multicolumn{3}{c|}{\textbf{DeepSeek}} & \multicolumn{3}{c|}{\textbf{Llama}} & \multicolumn{3}{c|}{\textbf{Qwen}} & \multicolumn{3}{c}{\textbf{GPT}}\\
 & \multicolumn{1}{c}{w/o M.} & \multicolumn{1}{c}{w/ M.} & \multicolumn{1}{c|}{w/ M.R.} & \multicolumn{1}{c}{w/o M.} & \multicolumn{1}{c}{w/ M.} & \multicolumn{1}{c|}{w/ M.R.} & \multicolumn{1}{c}{w/o M.} & \multicolumn{1}{c}{w/ M.} & \multicolumn{1}{c|}{w/ M.R.} & \multicolumn{1}{c}{w/o M.} & \multicolumn{1}{c}{w/ M.} & \multicolumn{1}{c}{w/ M.R.} \\ 
 \midrule
Sephora        & \underline{0.715} & \textbf{0.758} & 0.679 & \underline{0.675} & \textbf{0.681} & 0.648 & 0.579 & \underline{0.630} & \textbf{0.740} & 0.588 & \textbf{0.671} & \underline{0.654} \\
{Dianping} & {0.637} & {\underline{0.666}} & {\textbf{0.722}} & {0.368} & {\underline{0.386}} & {\textbf{0.408}} & {\underline{0.408}} & {0.390} & {\textbf{0.413}} & {0.351} & {\underline{0.368}} & {\textbf{0.369}} \\
{IMDB} & {\underline{0.534}} & {\textbf{0.586}} & {0.532} & {0.396} & {\underline{0.482}} & {\textbf{0.533}} & {0.487} & {\underline{0.493}} & {\textbf{0.519}} & {\underline{0.432}} & {0.420} & {\textbf{0.435}} \\
WeiboTech      & 0.415 & \textbf{0.577} & \underline{0.501} & 0.335 & \underline{0.417} & \textbf{0.420} & 0.296 & \underline{0.353} & \textbf{0.364} & 0.327 & \underline{0.488} & \textbf{0.698} \\

\bottomrule
\end{tabular}}
\vspace{-3mm}
\end{table}

\textbf{Results of Discriminative Tasks in TDGG.} 
For node retrieval, results on Hit@1 and Hit@10 are reported in \cref{tab:node_retrieval_all_dgnn}; for edge classification, in \cref{tab:edge_classification_dgnn}. 
Although existing DGNNs employ sophisticated modules to model dynamic interactions and achieve strong performance on discriminative tasks \citep{DyGLib}, their effectiveness diminishes when training on only 1,000 edges. Notably, as shown in \cref{tab:edge_classification_dgnn}, GAG-General outperforms DGNNs on most datasets, while gains on node retrieval are less pronounced (\cref{tab:node_retrieval_all_dgnn}). This demonstrates GAG-General’s superior ability to leverage structural, temporal, and textual information in DyTAGs, whereas DGNNs exhibit limited generalization due to their reliance on large-scale training data. These results further confirm GAG-General’s effectiveness in modeling DyTAGs, particularly in capturing dependencies across both discriminative and generative tasks.

\begin{table}[!b]
\vspace{-2mm}
\centering \caption{The results on Degree MMD, Spectra MMD, $D_k$, $\alpha$, and power-law validity under IDGG with GPT as the LLM backbone. Full results on the other three LLM backbones are available in \cref{tab:idgg_structural_deepseek}, \ref{tab:idgg_structural_llama}, and \ref{tab:idgg_structural_qwen}.} 
\vspace{-2mm}
\label{tab:idgg_structural_gpt}
\resizebox{0.9\linewidth}{!}{
    \begin{tabular}{l|cccccccc}
    \toprule
    Dataset & \textbf{Sephora} & \textbf{Dianping} & \textbf{WikiRevision} & \textbf{WikiLife} & \textbf{IMDB} & \textbf{WeiboTech} & \textbf{WeiboDaily} & \textbf{Cora} \\
    \midrule
    Degree MMD$\downarrow$          & 0.454 & 0.300 & 0.267 & 0.083 & 0.321 & 0.268 & 0.268 & 0.136 \\
    Spectra MMD$\downarrow$         & 0.189 & 0.453 & 0.229 & 0.208 & 0.423 & 0.201 & 0.439 & 0.244 \\
    $D_k$              & 0.147 & 0.069 & 0.045 & 0.099 & 0.252 & 0.064 & 0.156 & 0.091 \\
    $\alpha$           & 2.057 & 2.430 & 2.050 & 2.204 & 1.746 & 1.867 & 1.732 & 2.250 \\
    Power-law Validity & \ding{51}     & \ding{51}     & \ding{51}     & \ding{51}     & \ding{55}     & \ding{55}     & \ding{55}     & \ding{51} \\
    \bottomrule
    \end{tabular}}
\end{table}

\subsection{Inductive Dynamic Graph Generation}
\label{sec:exp_idgg}
\textbf{Results of IDGG.}  
In IDGG, structural quality results are summarized in \cref{tab:idgg_structural_gpt}. Generated DyTAGs show greater deviation from ground-truths than in TDGG—due to the added complexity of new node generation—yet retain key structural properties. For example, most Degree/Spectra MMDs exceed 0.2, but five of eight graphs still satisfy the power-law criterion, indicating reasonable fidelity. Similar to TDGG, ablation studies (\cref{tab:idgg_textual_4datasets,tab:idgg_graph_embedding_4datasets}) confirm that node memory and reflection enhance textual and structural quality, underscoring the importance of summarizing historical neighbor information. Moreover, {\cref{tab:idgg_vrdag_dggen_4datasets}} compares our framework with existing dynamic graph generation models, revealing their significant limitations: generated graphs exhibit notably lower structural fidelity and poorer node/edge attribute richness. These results highlight the need for dedicated DyTAG generation methods. 
Additionally, we demonstrate that generated graphs from IDGG can serve as data augmentation to enhance model performance on inductive new nodes; see details in \cref{IDGG-inductive}. 
Scalability analysis for IDGG is also included in \cref{app:scalability}.

\textbf{Hub Node Analysis in IDGG.}
The generated DyTAGs in IDGG demonstrate a critical evolutionary pattern: while maintaining structural congruence with ground-truth graphs, they develop highly connected hub nodes with divergent textual attributes. 
Specifically, for generating a DyTAG with 2,000 edges on Sephora, the top three newly generated hub nodes in the generated DyTAG include products like \emph{Vitamin C Serum}, \emph{Eye Cream}, and \emph{Neck Cream}. 
In contrast, the ground-truth graph’s top three hub nodes consist of products like \emph{Acne Control Clarifying Cleanser}, \emph{Rosebud Perfume}, and \emph{Moisturizing Lotion}. 
The visualization of the hub node structures in the ground-truth and generated graphs is shown in \cref{fig:hub-nodes-gt,fig:hub-nodes-gen}.
This divergence arises because IDGG mimics real-world graph evolution dynamics, ensuring structural fidelity while enabling the creation of new nodes with reasonable attributes that align with the underlying generative patterns of the real world.  
For instance, in recommendation systems, the hub nodes in the generated DyTAG can represent \textbf{emerging products} with high potential for virality, while those in the ground-truth graph correspond to \textbf{established bestsellers}. 
This capability positions DyTAG generation as a strategic tool for proactive decision-making in e-commerce and digital marketing. 
By identifying potential future hubs, platforms can prioritize resource allocation for product promotion, optimize advertising strategies, and anticipate market trends before they manifest in real-world data.

\begin{table}[!t]
\centering \caption{The results on average textual quality scores under IDGG. Full results are available in \cref{tab:idgg_textual}-\ref{tab:idgg_textual_cora}. } 
\vspace{-2mm}
\label{tab:idgg_textual_4datasets}
\resizebox{1.0\linewidth}{!}{
\begin{tabular}{l|ccc|ccc|ccc|ccc}\toprule
 & \multicolumn{3}{c|}{\textbf{DeepSeek}} & \multicolumn{3}{c|}{\textbf{Llama}} & \multicolumn{3}{c|}{\textbf{Qwen}} & \multicolumn{3}{c}{\textbf{GPT}}\\
 & \multicolumn{1}{c}{w/o M.} & \multicolumn{1}{c}{w/ M.} & \multicolumn{1}{c|}{w/ M.R.} & \multicolumn{1}{c}{w/o M.} & \multicolumn{1}{c}{w/ M.} & \multicolumn{1}{c|}{w/ M.R.} & \multicolumn{1}{c}{w/o M.} & \multicolumn{1}{c}{w/ M.} & \multicolumn{1}{c|}{w/ M.R.} & \multicolumn{1}{c}{w/o M.} & \multicolumn{1}{c}{w/ M.} & \multicolumn{1}{c}{w/ M.R.} \\ 
 \midrule
Sephora& 4.63 & \underline{4.73} & \textbf{4.77} & 4.58 & \underline{4.65} &\textbf{4.74} & 4.58 & \underline{4.68} & \textbf{4.78} & 4.58 & \underline{4.77} & \textbf{4.87} \\
{Dianping} & {4.29} & {\underline{4.44}} & {\textbf{4.65}} & {4.29} & {\underline{4.51}} & {\textbf{4.87}} & {4.04} & {\underline{4.50}} & {\textbf{4.68}} & {4.56} & {\underline{4.71}} & {\textbf{4.86}} \\
{IMDB} & {4.13} & {\underline{4.28}} & {\textbf{4.39}} & {4.22} & {\underline{4.31}} & {\textbf{4.43}} & {4.23} & {\underline{4.36}} & {\textbf{4.51}} & {4.19} & {\underline{4.29}} & {\textbf{4.49}} \\
WeiboTech& 4.60 & \underline{4.75} & \textbf{4.85} & 3.94 & \underline{4.00} & \textbf{4.75} & 4.56 & \underline{4.74} &\textbf{4.93} & 4.60 & \underline{4.71} & \textbf{4.93} \\
\bottomrule
\end{tabular}}
\vspace{-3mm}
\end{table}

\begin{table}[!t]
\centering \caption{The results on the graph embedding metric under IDGG. See \cref{tab:idgg_graph_embedding} for full results.} 
\vspace{-2mm}
\label{tab:idgg_graph_embedding_4datasets}
\resizebox{1.0\linewidth}{!}{
\begin{tabular}{l|ccc|ccc|ccc|ccc}\toprule
 & \multicolumn{3}{c|}{\textbf{DeepSeek}} & \multicolumn{3}{c|}{\textbf{Llama}} & \multicolumn{3}{c|}{\textbf{Qwen}} & \multicolumn{3}{c}{\textbf{GPT}}\\
 & \multicolumn{1}{c}{w/o M.} & \multicolumn{1}{c}{w/ M.} & \multicolumn{1}{c|}{w/ M.R.} & \multicolumn{1}{c}{w/o M.} & \multicolumn{1}{c}{w/ M.} & \multicolumn{1}{c|}{w/ M.R.} & \multicolumn{1}{c}{w/o M.} & \multicolumn{1}{c}{w/ M.} & \multicolumn{1}{c|}{w/ M.R.} & \multicolumn{1}{c}{w/o M.} & \multicolumn{1}{c}{w/ M.} & \multicolumn{1}{c}{w/ M.R.} \\ 
 \midrule
Sephora        & 0.621& \textbf{0.661} & \underline{0.634} & 0.602& \underline{0.612} & \textbf{0.647} & 0.569& \underline{0.587} & \textbf{0.616} & 0.601& \underline{0.603} & \textbf{0.628} \\
{Dianping} & {0.739} & {\underline{0.749}} & {\textbf{0.769}} & {0.426} & {\underline{0.457}} & {\textbf{0.512}} & {0.531} & {\underline{0.542}} & {\textbf{0.578}} & {\underline{0.521}} & {\textbf{0.527}} & {0.505} \\
{IMDB} & {0.601} & {\underline{0.616}} & {\textbf{0.729}} & {0.511} & {\underline{0.522}} & {\textbf{0.568}} & {0.502} & {\underline{0.557}} & {\textbf{0.588}} & {0.521} & {\underline{0.535}} & {\textbf{0.563}} \\
WeiboTech      & \textbf{0.604}& \underline{0.591} & 0.562 & 0.601& \underline{0.629} & \textbf{0.695} & 0.531& \underline{0.547} & \textbf{0.549} & 0.501& \underline{0.514} & \textbf{0.536} \\
\bottomrule
\end{tabular}}
\vspace{-5mm}
\end{table}

\begin{table}[!h]
\vspace{-2mm}
\centering 
\caption{The results on the graph structural and the graph embedding metrics under IDGG of our framework and current feature-supportive dynamic graph generation models. Ours correspond to the best results of our proposed GAG-General among four LLM backbones. See \cref{tab:idgg_vrdag_dggen} for full results.} 
\vspace{-2mm}
\label{tab:idgg_vrdag_dggen_4datasets}
\resizebox{1.0\linewidth}{!}{
\begin{tabular}{l|cccc|cccc|cccc}\toprule
 & \multicolumn{4}{c|}{\textbf{Sephora}} & \multicolumn{4}{c|}{\textbf{Dianping}} & \multicolumn{4}{c}{\textbf{Cora}} \\
 & \multicolumn{1}{c}{Ours} & \multicolumn{1}{c}{VRDAG} & \multicolumn{1}{c}{DG-Gen} & \multicolumn{1}{c|}{{TIGGER-I}} 
 & \multicolumn{1}{c}{Ours} & \multicolumn{1}{c}{VRDAG} & \multicolumn{1}{c}{DG-Gen} & \multicolumn{1}{c|}{{TIGGER-I}} 
 & \multicolumn{1}{c}{Ours} & \multicolumn{1}{c}{VRDAG} & \multicolumn{1}{c}{DG-Gen} & \multicolumn{1}{c}{{TIGGER-I}} \\ 
\midrule
Degree MMD $\downarrow$ 
& \textbf{0.370} & 0.795 & 0.422 & {0.622} 
& \textbf{0.150} & 0.887 & 0.167 & {0.446} 
& \textbf{0.073} & 0.877 & 0.212 & {0.372} \\
Spectra MMD $\downarrow$ 
& \textbf{0.189} & 0.847 & 0.274 & {0.687} 
& 0.351 & 0.808 & \textbf{0.245} & {0.341} 
& \textbf{0.181} & 0.760 & 0.365 & {0.389} \\
Power-law Validity 
& \ding{51} & \ding{55} & \ding{55} & {\ding{55}} 
& \ding{51} & \ding{55} & \ding{51} & {\ding{51}} 
& \ding{51} & \ding{55} & \ding{55} & {\ding{55}} \\
Graph Embedding $\uparrow$ 
& \textbf{0.661} & 0.011 & 0.228 & {0.085} 
& \textbf{0.769} & 0.024 & 0.517 & {0.580} 
& \textbf{0.828} & 0.053 & 0.056 & {0.197} \\
\bottomrule
\end{tabular}}
\vspace{-5mm}
\end{table}

\section{Future Work}
\label{sec:future_work}

\textbf{Generative Framework Optimization.}  
Leveraging our high-quality, text-rich GDGB datasets, the proposed generative framework achieves groundbreaking performance in the novel DyTAG generation tasks of TDGG and IDGG. 
Despite this progress, key aspects of the pipeline—especially in IDGG—require further refinement. 
Future work should focus on improving node generation strategies and selection mechanisms, such as incorporating adaptive node sampling to balance diversity and fidelity, and leveraging existing DGNN techniques to enhance the accuracy and efficiency of candidate node retrieval and ranking. 
These advancements, grounded in GDGB, will strengthen DyTAG models and provide a solid foundation for generative graph foundation models, addressing challenges in both structural fidelity and attribute richness.

\textbf{Further Applications.}  
Beyond model optimization, DyTAG generation holds substantial practical value. 
As shown in \cref{sec:exp_idgg}, identifying generated future hub nodes—like potential hit products in e-commerce—demonstrates its utility in node-level predictive modeling. 
At the edge and graph levels, DyTAG generation serves as effective data augmentation tools for sparse graphs, producing synthetic graphs that preserve structure while enriching textual content. 
This capability extends applications to e-commerce (modeling generative recommendation), social networks (forecasting misinformation spread), and urban planning (predicting infrastructure usage). By integrating dynamic graph analysis with forward-looking textual modeling, DyTAG generation can drive impactful real-world solutions across industries.
\section{Conclusion}
\label{sec:conclusion}


In this work, we propose \textbf{G}enerative \textbf{D}yTA\textbf{G} \textbf{B}enchmark (\textbf{GDGB}), which resolves existing limitations in datasets, task formulations, and evaluation metrics for DyTAG generation.
GDGB comprises 8 meticulously curated datasets with semantically rich node/edge attributes, enabling rigorous evaluation of DyTAG generation.
By defining two novel tasks—TDGG (transductive generation with fixed node sets) and IDGG (inductive generation with new node generation)—we effectively model the dynamic expansion of real-world graphs.
Additionally, we propose GAG-General, an LLM-based multi-agent framework that generalizes across diverse DyTAG structures and integrates multifaceted metrics for holistic evaluation in DyTAG generation.
Experimental results demonstrate that GDGB establishes robust benchmarks for DyTAG generation, revealing the interplay between structural and textual features as a cornerstone for effective generative models.
In summary, GDGB lays the groundwork for future research on advancing DyTAG generation.

\section*{Acknowledgments}
The work was partially done at Gaoling School of Artificial Intelligence, Beijing Key Laboratory of Research on Large Models and Intelligent Governance, Engineering Research Center of Next-Generation Intelligent Search and Recommendation, MOE, and Pazhou Laboratory (Huangpu), Guangzhou, Guangdong 510555, China. 
This research was supported in part by National Natural Science Foundation of China (No. 92470128, No. U2241212), by Beijing Outstanding Young Scientist Program No.BJJWZYJH012019100020098, by the National Key Research and Development Plan of China (2023YFB4502305) and Ant Group through CCF-Ant Research Fund. 
We also wish to acknowledge the support provided by the fund for building world-class universities (disciplines) of Renmin University of China, by Engineering Research Center of Next-Generation Intelligent Search and Recommendation, Ministry of Education, by Intelligent Social Governance Interdisciplinary Platform, Major Innovation \& Planning Interdisciplinary Platform for the “Double-First Class” Initiative, Public Policy and Decision-making Research Lab, and Public Computing Cloud, Renmin University of China.

\bibliography{reference}
\bibliographystyle{iclr2026_conference}

\newpage

\appendix
\startcontents[appendices]
\addcontentsline{toc}{section}{Appendix: Contents} 

\printcontents[appendices]{}{1}{}

\newpage

\section{Ethics Statement}
\label{app:ethics_statement}
Our proposed GDGB offers both potential benefits and challenges. Positively, it advances DyTAG generation research, enabling applications in e-commerce (e.g., personalized recommendations), social networks (e.g., community governance), and urban planning frameworks. These advancements could enhance data-driven decision-making while promoting DyTAG generation development through standardized evaluation protocols.
Negatively, the misuse of generated DyTAGs might risk generating misleading information or amplifying biases in dynamic network scenarios (e.g., illegal user interactions). However, these risks are mitigated under proper usage and regulatory frameworks—such as data anonymization, access control, and ethical guidelines—which ensure responsible deployment in real-world applications. By adhering to these principles, GDGB’s technical contributions can safely drive innovation in DyTAG learning.

\section{Reproducibility Statement}
To ensure the reproducibility of our work, we have made the following efforts: 
Our proposed GDGB comprises eight carefully selected and rigorously processed DyTAG datasets, hosted through a sharing link (\url{https://www.kaggle.com/datasets/gdgbdataset/gdgb-a-benchmark-for-generative-dytag-learning}). 
The proposed GDGB benchmark and our generative framework GAG-General are fully described in this paper, including detailed dataset construction processes and model implementation specifications. 
All code and data preprocessing scripts are publicly available in our code repository (\url{https://github.com/Lucas-PJ/GDGB-ALGO}), which includes comprehensive instructions for reproducing the TDGG and IDGG experiments. 
The repository also provides evaluation scripts of our proposed multifaceted metrics to facilitate verification of the reported results.
The GDGB website is \url{https://gdgb-algo.github.io/}. The website allows researchers to submit methods and track performance via leaderboards on TDGG and IDGG.

\section{Dataset}
\subsection{Dataset Details}
\label{app:dataset_details}
\underline{Sephora}\footnote{\url{https://www.kaggle.com/datasets/nadyinky/sephora-products-and-skincare-reviews/}} is a dataset collected from Kaggle, documenting user reviews of beauty and skincare products on the Sephora e-commerce platform. 
The temporal span of the dataset ranges from August 28, 2008, to March 21, 2023. 
Specifically, the dataset includes rich textual information about users, such as skin tone, skin type, hair color, eye color, and historical review statistics. 
Notably, it also contains detailed textual features of beauty products from the Sephora online store, including product and brand names, prices, ingredients, ratings, and all associated attributes, which support the construction of textual features for nodes in this DyTAG. 
For user reviews of beauty and skincare products, the dataset provides review ratings (ranging from 1 to 5) as edge labels, and detailed textual reviews as edge content. 
Consequently, the Sephora dataset is represented as a bipartite DyTAG, where users and beauty products serve as nodes with textual features, and an edge represents a user's rating and textual review of a product at a given time.

\underline{Dianping}\footnote{\url{http://www.dianping.com}} \citep{dianping1,dianping2} is a business review dataset derived from Dianping, a prominent platform for business recommendations (e.g., restaurants), spanning from July 3, 2009, to February 8, 2012. 
The dataset records Dianping users' historical review statistics, frequently reviewed cities, and detailed business information, including store names, addresses, cuisine styles, average costs, and historical scores. 
User reviews of businesses include multi-dimensional ratings (e.g., flavor, environment, service) in addition to overall scores (ranging from 0 to 5) as edge labels. 
Reviews also contain rich textual content and supplementary evaluations for specific items such as recommended dishes or special services, providing extensive edge textual features. 
The dataset is structured as a bipartite DyTAG, where Dianping users and businesses are nodes with textual features, and an edge represents a user's detailed textual review of a business at a given time.

\underline{WikiRevision}\footnote{\url{https://dumps.wikimedia.org/enwiki/20241201/}} is a dataset cleaned and processed from Wikipedia's official data dumps, recording user revisions and modifications to Wikipedia pages with the selected dumps corresponding to December 1, 2024. 
The temporal span ranges from January 30, 2001, to December 3, 2024. 
The dataset includes Wikipedia users' usernames, historical editing statistics, frequently edited pages, and all revised Wikipedia page titles. 
To enrich textual features for page nodes, we crawl the first paragraph of each corresponding Wikipedia page from Wikipedia. 
User revisions are categorized into two edge classes (minor vs. non-minor edits), with the accompanying revision comments serving as edge textual features. 
The dataset is structured as a bipartite DyTAG, where Wikipedia users and pages are nodes with textual features, and an edge represents a user's textual revision summary on a page at a given time.

\underline{WikiLife} \citep{wikilife} is a dataset derived from Wikipedia's official data dumps, recording historical records of notable individuals being physically present at specific locations. 
The temporal span ranges from 202 to 2024. 
The original dataset \citep{wikilife} is extracted from English Wikipedia's biography pages as person-time-location triplets. 
We extend this by crawling the first paragraphs of corresponding Wikipedia pages for persons and locations as extra textual features. 
Additionally, we utilize 24 life trajectory activity categories (e.g., birth/death, education, career) from \citep{wikilife} as edge labels and map the original Wikipedia text of triplets to describe the specific activities of individuals at locations. 
The WikiLife dataset is structured as a bipartite DyTAG, where persons and locations are nodes with textual features, and an edge represents a person's textual life trajectory activity in a location at a given time.

\underline{IMDB}\footnote{\url{https://datasets.imdbws.com/}} is a dataset based on IMDB's official data, documenting actor/actress collaboration networks. 
The temporal span ranges from 1988 to 2031 (including collaboration information of official future films like Avatar 3). 
The dataset includes actors' and actresses' names, birth/death years, primary professions, and extra biographical information crawled from Wikipedia to enrich node textual features. 
Edge categories correspond to 20 types of movie genres (e.g., comedy, drama, crime, action), with edge texts comprising the title of the collaborated movie and roles played by actors/actresses, respectively. 
The IMDB dataset is a non-bipartite DyTAG, where actors/actresses are nodes with textual features, and an edge represents a movie collaboration relationship between them at a given time.

\underline{WeiboTech}\citep{sagraph} is a dataset collected from the Weibo social platform, recording user interactions (comments and reposts). 
The temporal span ranges from December 29, 2023, to January 5, 2024. 
The original dataset \citep{sagraph} focuses on advertising strategies for electric toothbrushes. 
We reprocess the raw data to extract temporal information and restructure it into the DyTAG format. 
The dataset includes user profiles (e.g., usernames, gender, regions, follower/followee counts, self-introductions) and interaction edges categorized as comment or repost, with edge texts comprising source post content and destination user comments/reposts. 
Named WeiboTech due to its focus on technology-related topics (e.g., electronics, automobiles), the dataset is a non-bipartite DyTAG, where Weibo users are nodes with textual features, and an edge represents user interactions at a given time.

\underline{WeiboDaily}\citep{sagraph} is also a dataset collected from the Weibo social platform, recording user interactions (comments and reposts). 
Different from WeiboTech, the temporal span ranges from December 1, 2023, to December 31, 2023, and the original dataset \citep{sagraph} focuses on advertising strategies for ABCReading. 
Thus, WeiboDaily spans a full month and targets daily life topics (e.g., lifestyle sharing), enabling continuous modeling analysis. 
Similarly, the dataset includes user profiles (e.g., usernames, gender, regions, follower/followee counts, self-introductions) and interaction edges categorized as comment or repost, with edge texts comprising source post content and destination user comments/reposts. 
The dataset is a non-bipartite DyTAG, where Weibo users are nodes with textual features, and an edge represents user interactions at a given time.

\underline{Cora} is an extended version of the classic citation network dataset Cora \citep{cora}, documenting academic paper citation relationships. 
The temporal span ranges from February 1, 1985, to September 1, 2024. 
The original Cora dataset records citation relationships and node categories \citep{cora}. 
We expand the dataset by crawling first-order and second-order paper information via references, enriching the size of the citation network. 
Node textual features include paper titles, abstracts, authors, and citation counts. 
Edge texts are extracted from the exact sentence in the citing paper that references the cited paper.
With regards to the edge labels, based on the section name of each citation in the citing paper, they are mapped to five categories: 1) Intro \& Background, 2) Tech \& Methodology, 3) Experiment \& Conclusion, 4) Topic-specific, and 5) Others. 
The Cora dataset is a non-bipartite DyTAG, where papers are nodes with textual features, and an edge indicates a citation relationship at a given time.

\subsection{Dataset Licenses}
\label{app:data_license}
In this section, we provide dataset licenses for each proposed DyTAG.

\underline{Sephora}: CC BY 4.0 license (Creative Commons Attribution 4.0 International License).
The original dataset can be found \href{https://www.kaggle.com/datasets/nadyinky/sephora-products-and-skincare-reviews/}{here}.

\underline{Dianping}: CC BY-SA license (Creative Commons Attribution-ShareAlike License). 
The original dataset can be found \href{http://yongfeng.me/dataset/}{here}.

\underline{WikiRevision}: GFDL (GNU Free Documentation License) and CC BY-SA license (Creative Commons Attribution-ShareAlike License).
The original dataset can be found \href{https://dumps.wikimedia.org/enwiki/20241201/}{here}.

\underline{WikiLife}: GFDL (GNU Free Documentation License) and CC BY-SA license (Creative Commons Attribution-ShareAlike License).
The original dataset can be found \href{https://dumps.wikimedia.org/}{here}.

\underline{IMDB}: Subject to your compliance with these Conditions of Use and your payment of any applicable fees, IMDB or its content providers grants you a limited, non-exclusive, non-transferable, non-sublicenseable license to access and make personal and non-commercial use of the IMDB Services, including digital content available through the IMDB Services, and not to download (other than page caching) or modify this site, or any portion of it, except with express written consent of IMDB. Additional license terms may be found in the Terms. The IMDB Services or any portion of such services may not be reproduced, duplicated, copied, sold, resold, visited, or otherwise exploited for any commercial purpose without the express written consent of IMDB. This license does not include any resale or commercial use of any IMDB Service or its contents or any derivative use of this site or its contents. All licenses are non-exclusive, and all rights not expressly granted to you in these Conditions of Use or any applicable Terms are reserved and retained by IMDB or its licensors, suppliers, publishers, rightsholders, or other content providers. You will use all IMDB Services in compliance with all applicable laws.
The original dataset can be found \href{https://datasets.imdbws.com/}{here}.

\underline{WeiboTech}: CC-BY 4.0 license (Creative Commons Attribution 4.0 International License).
The original dataset can be found \href{https://github.com/xiaoqzhwhu/SAGraph}{here}.

\underline{WeiboDaily}: CC-BY 4.0 license (Creative Commons Attribution 4.0 International License).
The original dataset can be found \href{https://github.com/xiaoqzhwhu/SAGraph}{here}.

\underline{Cora}: CC BY-SA license (Creative Commons Attribution-ShareAlike).
The original dataset can be found \href{https://github.com/CurryTang/Graph-LLM}{here}.

\subsection{Dataset Analysis}
\begin{figure}[h]
  \centering
  \vspace{-3mm}
  \includegraphics[width=0.90\linewidth]{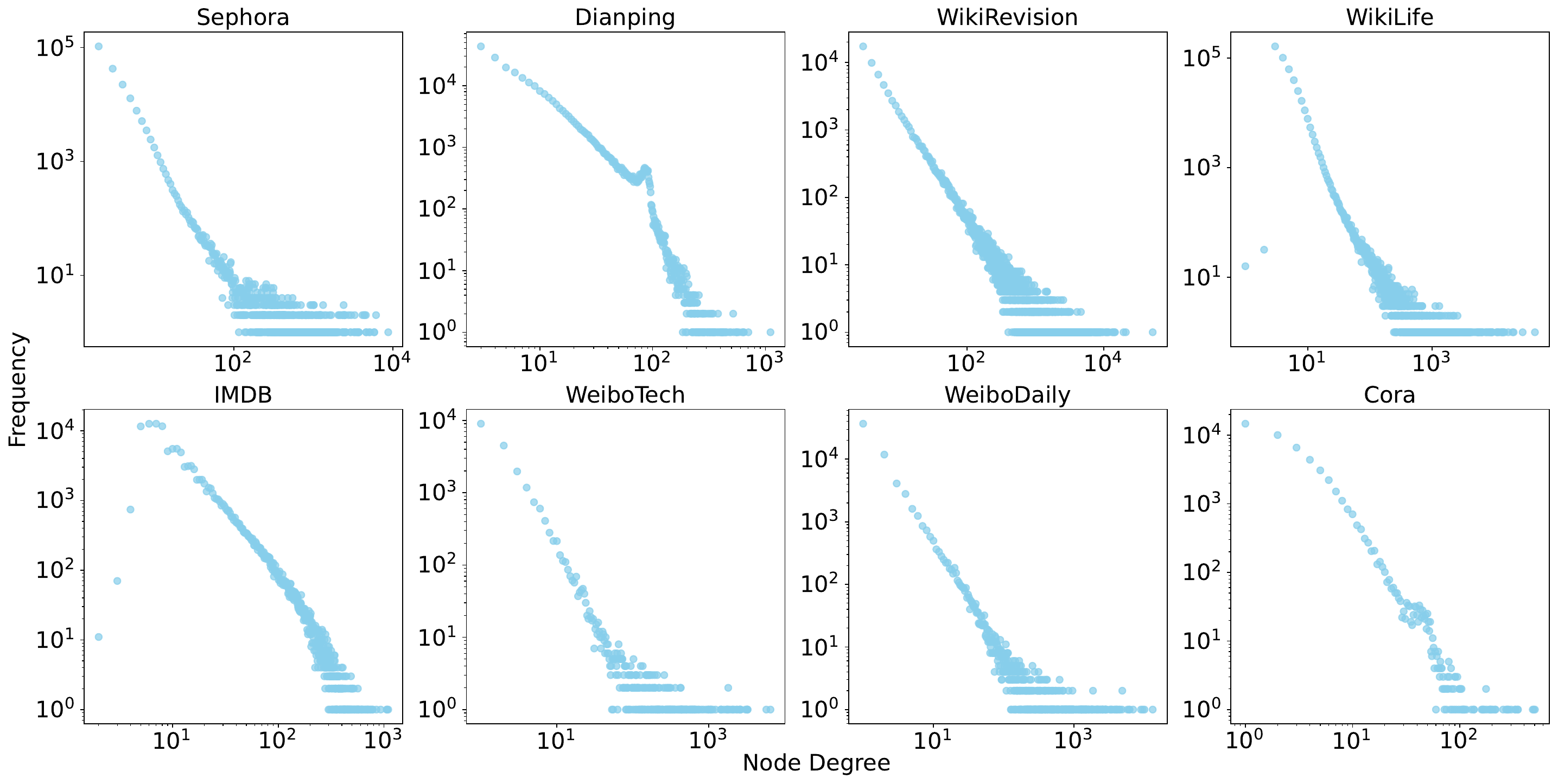}
  \caption{Distribution of node degree on GDGB datasets.}
  \vspace{-3mm}
  \label{fig:node_degree}
\end{figure}

To provide a comprehensive overview of our GDGB datasets, we present the node degree distribution in \cref{fig:node_degree}. 
Most datasets exhibit a long-tailed distribution, which aligns with the real-world graph growth patterns governed by power-law principles \citep{powerlaw}. 
This characteristic reflects the uneven connectivity observed in natural networks, such as social networks or citation graphs, where a few nodes dominate high degrees while the majority remain sparsely connected.

\begin{figure}[h]
  \centering
  \includegraphics[width=0.90\linewidth]{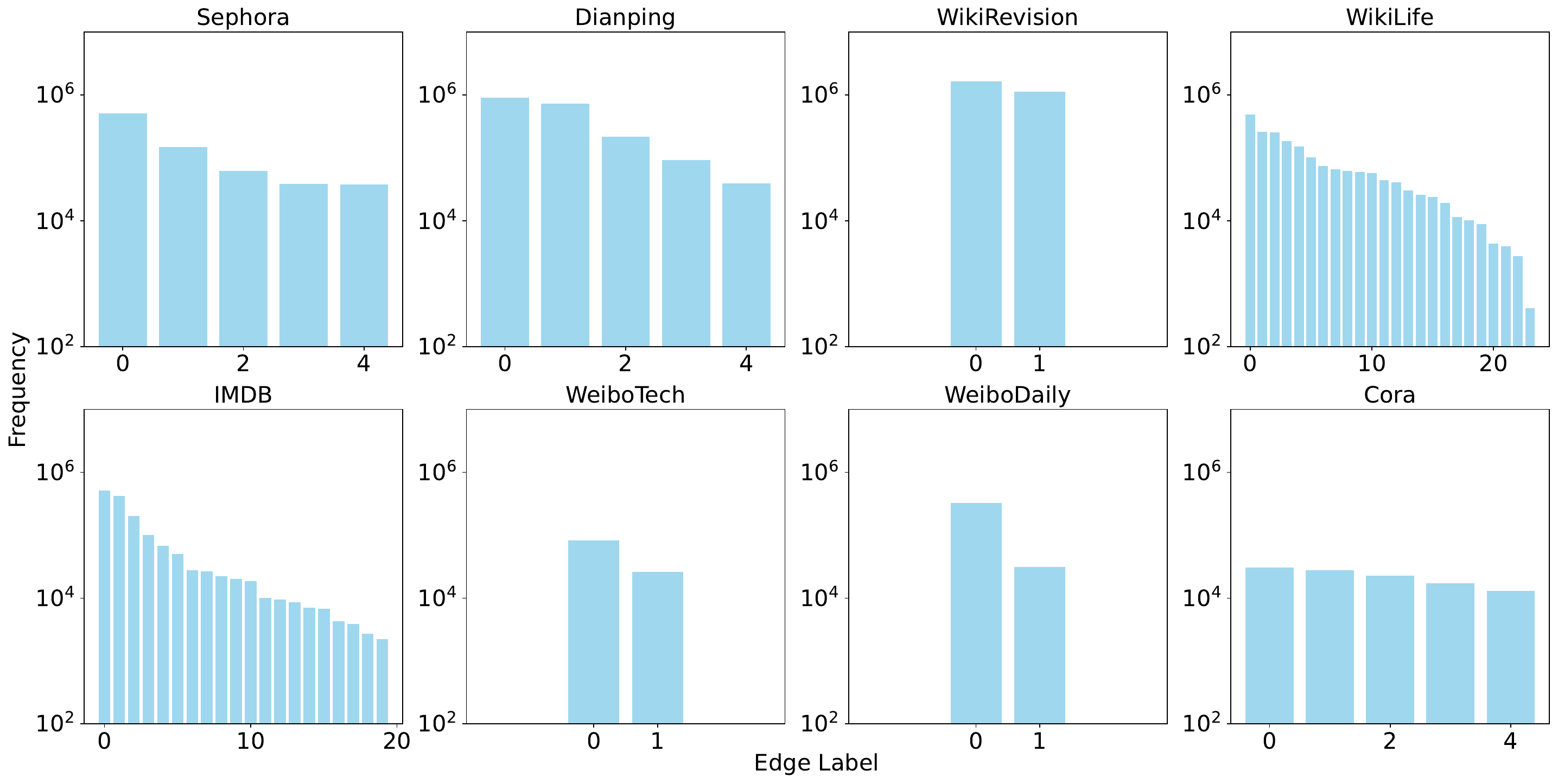}
  \caption{Distribution of the number of edges for each label on GDGB datasets.}
   \vspace{-3mm}
  \label{fig:edge_label}
\end{figure}

Furthermore, \cref{fig:edge_label} illustrates the edge label distribution of the GDGB datasets. The results reveal that most datasets maintain a balanced label distribution, mitigating the risk of performance degradation in edge classification tasks caused by extreme class imbalances. 
However, a subset of datasets exhibits relatively skewed label distributions, intentionally introducing realistic challenges akin to practical applications (e.g., rare but critical edge types in life trajectory or movie collaboration scenarios). 
This dual distribution pattern ensures both diversity and practical relevance, enhancing the dataset's utility for robust model evaluation.

\subsection{Comparison with Existing Datasets for Graph Generation.}
\label{app:comparision_dataset}
In graph generation research, while significant progress has been made in molecular graph generation, non-molecular graph generation has received comparatively less attention \citep{DBLP:journals/access/FaezOBR21}. 
Consequently, existing models for non-molecular graph generation are often trained on synthetic datasets such as Grid \citep{Bigg}, Community, and SBM \citep{graphrnn, DBLP:journals/pami/GuoZ23}, or on static, topology-only graphs, such as Ego and Polblogs \citep{EDGE}.
Further, we observe a lack of focus on dynamic graph generation, despite the fact that real-world non-molecular graphs inherently exhibit temporal evolution in both topology and node attributes \citep{DBLP:journals/pami/GuoZ23}. 
Among the most commonly used continuous-time dynamic graph datasets, such as LastFM and MOOC \citep{DGGEN, TIGGER, DGB}, both are derived from interaction networks \citep{DBLP:conf/nips/PoursafaeiHPR22} that inherently include textual attributes of users/items and timestamped interactions. However, these datasets suffer from critical limitations: MOOC provides only numerical features extracted from raw text, while LastFM discards attribute information entirely.

Although widely adopted, existing datasets face notable challenges in modeling the co-evolution of temporal, structural, and textual attributes in graphs. 
First, static graph datasets lack temporal annotations, making them incompatible with dynamic graph generation tasks. 
Even discrete-time dynamic graph datasets like Bitcoin-Alpha \citep{bitcoin,TIGGER}, which usually represent graphs as discrete daily snapshots, suffer from sparse temporal distributions of edges, complicating the fine-grained modeling of structural and temporal dynamics. 
Second, most dynamic graph datasets omit raw textual data, relying instead on preprocessed numeric or categorical features \citep{DGGEN}, which limits the exploration of rich textual attributes in real-world scenarios. 
Finally, more recently, datasets such as the DyTAGs from DTGB \citep{DTGB}, while providing edge-level text features, often lack sufficient and high-quality node-level textual information.
For instance, six out of eight datasets of DTGB contain (source) node texts that are typically just identifiers like email addresses or username.
This is a critical limitation, as real-world applications frequently require detailed node-wise text attributes to disambiguate entities beyond IDs.
The statistics of the above mentioned datasets and our proposed datasets are shown in \cref{tab:all_generative_datasets}.

\begin{table}[t] \small
\caption{Statistics of our proposed datasets and comparison with existing datasets for generative tasks. $|V|_{\max}$ and $|E|_{\max }$ represent the maximum number of nodes and maximum number of edges in the graph list of the dataset, respectively.} 
\centering \scalebox{0.71}{
\begin{tabular}{c||c|ccccccc}
    \toprule
    & {Dataset} & Nodes & Edges & Edge Categories & Timestamps & Domain & Text Attributes & Bipartite \\
    \midrule
    \midrule
    \multirow{4}{1.2cm}{\centering Previous Static Graphs} & \textbf{Grid} & $|V|_{\max }=361$ & $|E|_{\max }=684$
    & \textbackslash & \textbackslash & Synthetic & \textbackslash & \ding{55}\\
    & \textbf{Community} & $|V|_{\max}=160$ & $|E|_{\max }=1945$
    & \textbackslash & \textbackslash & Synthetic & \textbackslash & \ding{55}\\
    & \textbf{Ego} & $|V|_{\max }=399$ & $|E|_{\max }=1071$
    & \textbackslash & \textbackslash & Social Network & \textbackslash & \ding{55}\\
    & \textbf{Polblogs} & 1,222 & 16,714 & \textbackslash & \textbackslash & Interaction & \textbackslash & \ding{55}\\
    \midrule
    \midrule
    \multirow{5}{1.2cm}{\centering Previous Dynamic Graphs}
    & \textbf{MOOC} & 7,047/97 & 411,749
    & \textbackslash & 345,600 & Interaction & \textbackslash & \ding{51}\\
    & \textbf{LastFM} & 980/1,000 & 1,293,103 & \textbackslash & 1,283,614 & Interaction & \textbackslash & \ding{51}\\
    & \textbf{Reddit} &  10,000/984 & 672,447
    & \textbackslash &  669,065 & Social & \textbackslash & \ding{51}\\
    & \textbf{Wikipedia} & 8,227/1,000 & 157,474
    & \textbackslash & 152,757 & Interaction &  \textbackslash & \ding{55}\\
    & \textbf{Bitcoin-Alpha} & 3,783 & 24,186
    & \textbackslash & 191 & Financial Network & \textbackslash & \ding{55}\\

    \midrule
    \midrule
    \multirow{8}{1.2cm}{\centering Previous DyTAGs (DTGB)}  
    &\textbf{Stack elec} &67,155/330,547 &1,262,225 &2 &5,224 & Multi-round dialogue &Node \& Edge & \ding{51} \\
    &\textbf{Stack ubuntu} &180,261/493,987 &1,497,006 &2 &4,972 & Multi-round dialogue &Node \& Edge & \ding{51} \\
    &\textbf{Googlemap CT} &83,796/27,372 &1,380,623 &5 &55,521 & E-commerce &Node \& Edge & \ding{51} \\
    &\textbf{Amazon movies} &233,459/60,107 &3,217,324 &5 &7,287 & E-commerce &Node \& Edge & \ding{51} \\
    &\textbf{Yelp} &1,987,896/150,346 &6,990,189 &5 &6,036 & E-commerce  &Node \& Edge & \ding{51} \\
    &\textbf{Enron} &42,711 &797,907 &10 &1,006 & E-mail &Node \& Edge & \ding{55} \\
    &\textbf{GDELT} &6,786 &1,339,245 &237 &2,591 & Knowledge graph & Node \& Edge & \ding{55} \\
    &\textbf{ICEWS1819} &31,796 &1,100,071 &266 &730 & Knowledge graph &Node \& Edge & \ding{55} \\
    \midrule
    \midrule
    \multirow{8}{*}{GDGB} & \textbf{Sephora}&210,357/2,274&801,234&5&5,314& E-commerce &Node \& Edge & \ding{51}\\
    &\textbf{Dianping}&158,541/88,118&1,990,409&5&745,151& E-commerce & Node \& Edge & \ding{51}\\
    &\textbf{WikiRevision}&75,622/3,204&2,778,732&2&2,766,153&  Web Interaction&Node \& Edge & \ding{51}\\
    &\textbf{WikiLife}&406,148/54,513&1,996,520&24&1,810& Celebrity Biography&Node \& Edge & \ding{51}\\
    &\textbf{IMDB}&125,714&1,534,162&20&122& Movie Collaboration&Node \& Edge & \ding{55}\\
    &\textbf{WeiboTech}&20,767&109,345&2&79,925& Social Network&Node \& Edge & \ding{55}\\
    &\textbf{WeiboDaily}&66,500&354,098&2&293,662& Social Network&Node \& Edge & \ding{55}\\
    &\textbf{Cora}&48,797&110,788&5&8,274& Citation&Node \& Edge & \ding{55}\\
\bottomrule
\end{tabular}}
\label{tab:all_generative_datasets}
\end{table}

\subsection{Comparison Between GDGB and DTGB datasets} \label{app:comparison_with_dtgb}
We provide a comprehensive analysis and empirical comparison between our proposed \textbf{GDGB} and the existing \textbf{DTGB}. 
While both benchmarks contain dynamic text-attributed graphs, we demonstrate that GDGB is fundamentally distinct in construction philosophy, data quality, and utility for generative tasks.

\begin{figure}[t]
  \centering
        \includegraphics[width=0.2845\linewidth]{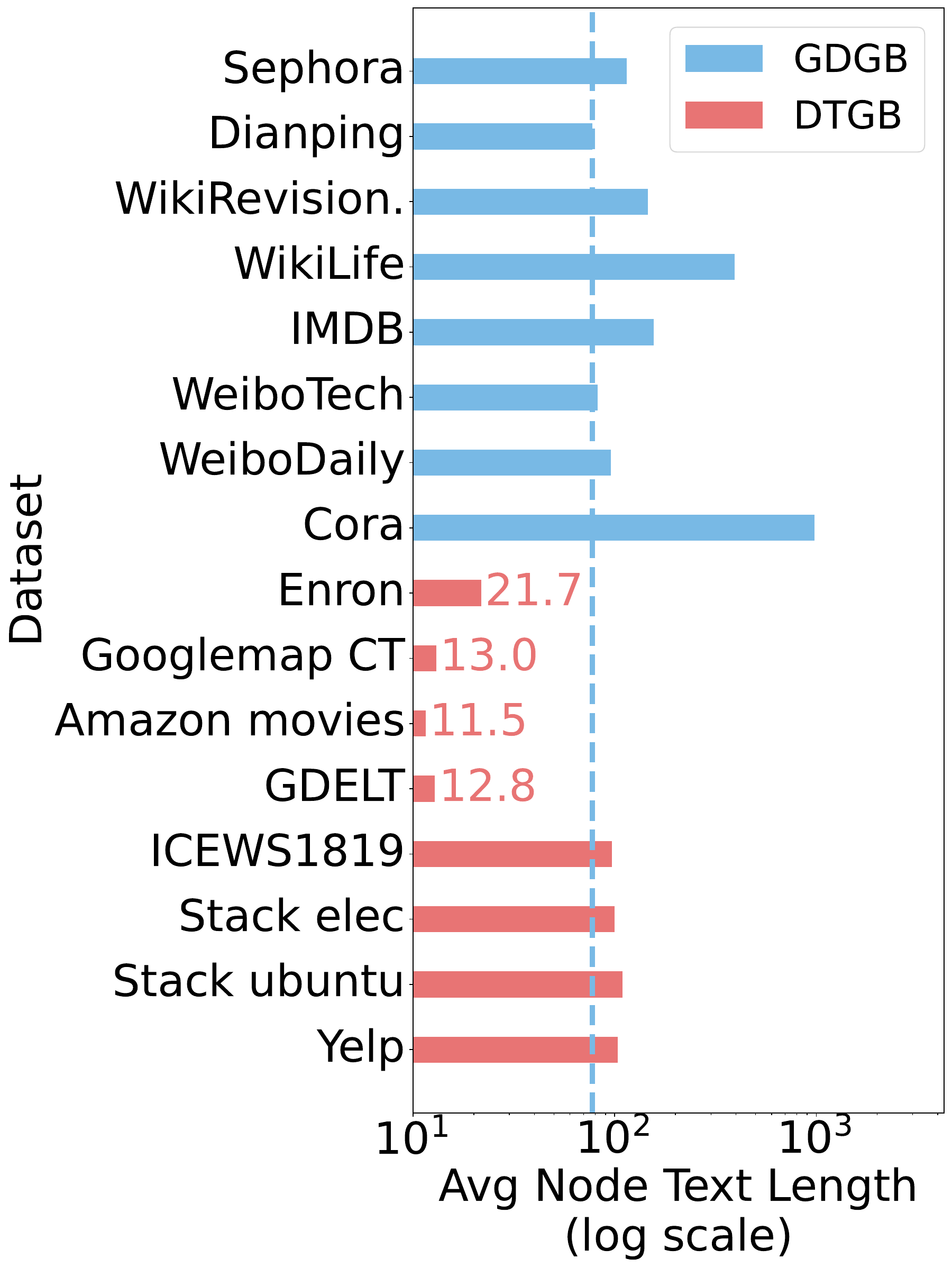}
        \includegraphics[width=0.185\linewidth]{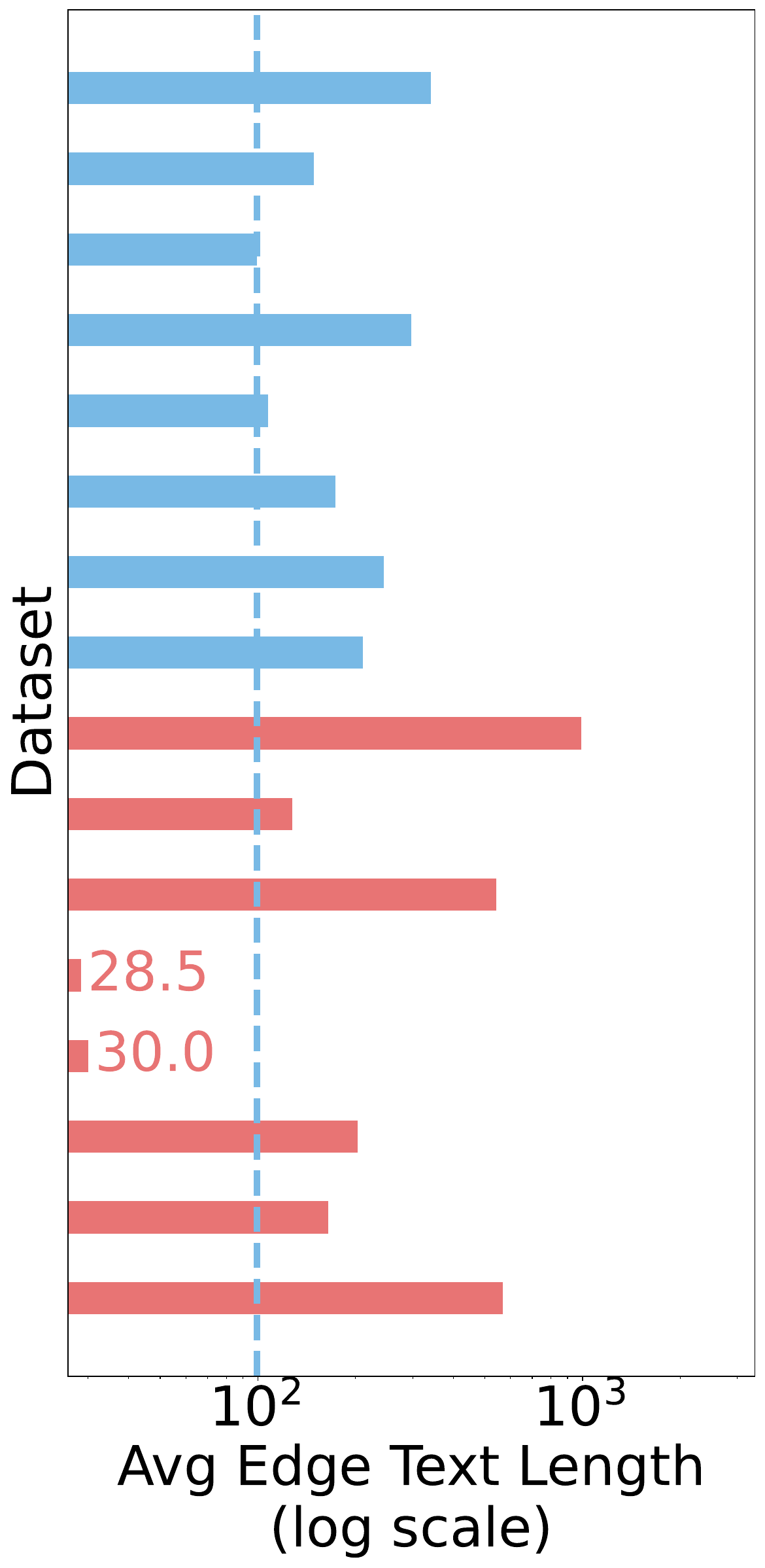}
        \caption{\textbf{Left}: Average node text lengths on GDGB and DTGB datasets. In non-bipartite graphs, text lengths are averaged across all nodes. For bipartite graphs, averages are calculated for source nodes. \textbf{Right}: Average edge text lengths on each dataset.} 
        \label{fig:node_edge_text_len}
\end{figure}

\begin{figure}[t]
  \centering
  \begin{minipage}[b]{0.48\linewidth}
        \includegraphics[width=0.5685\linewidth]{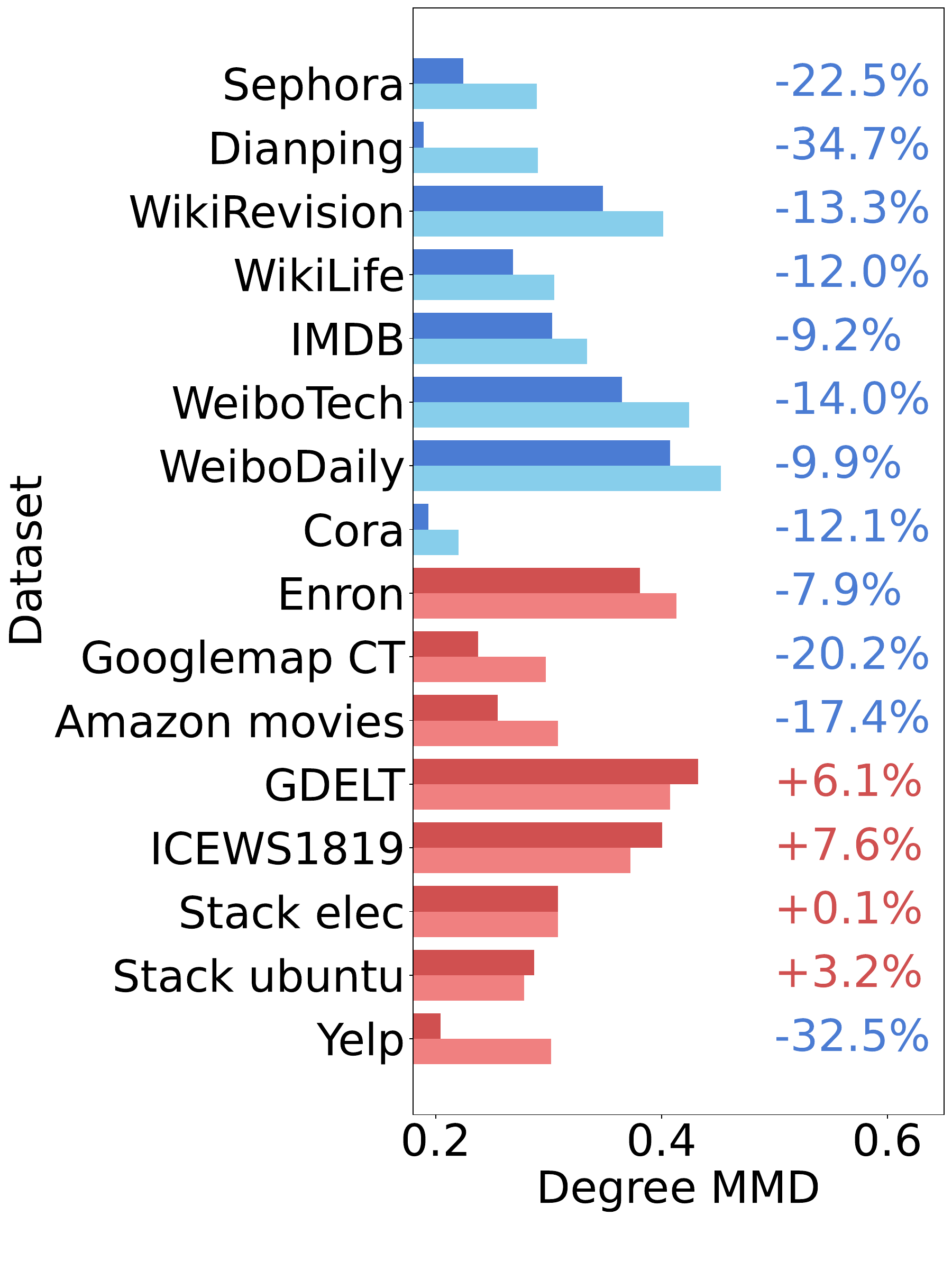}
        \includegraphics[width=0.3695\linewidth]{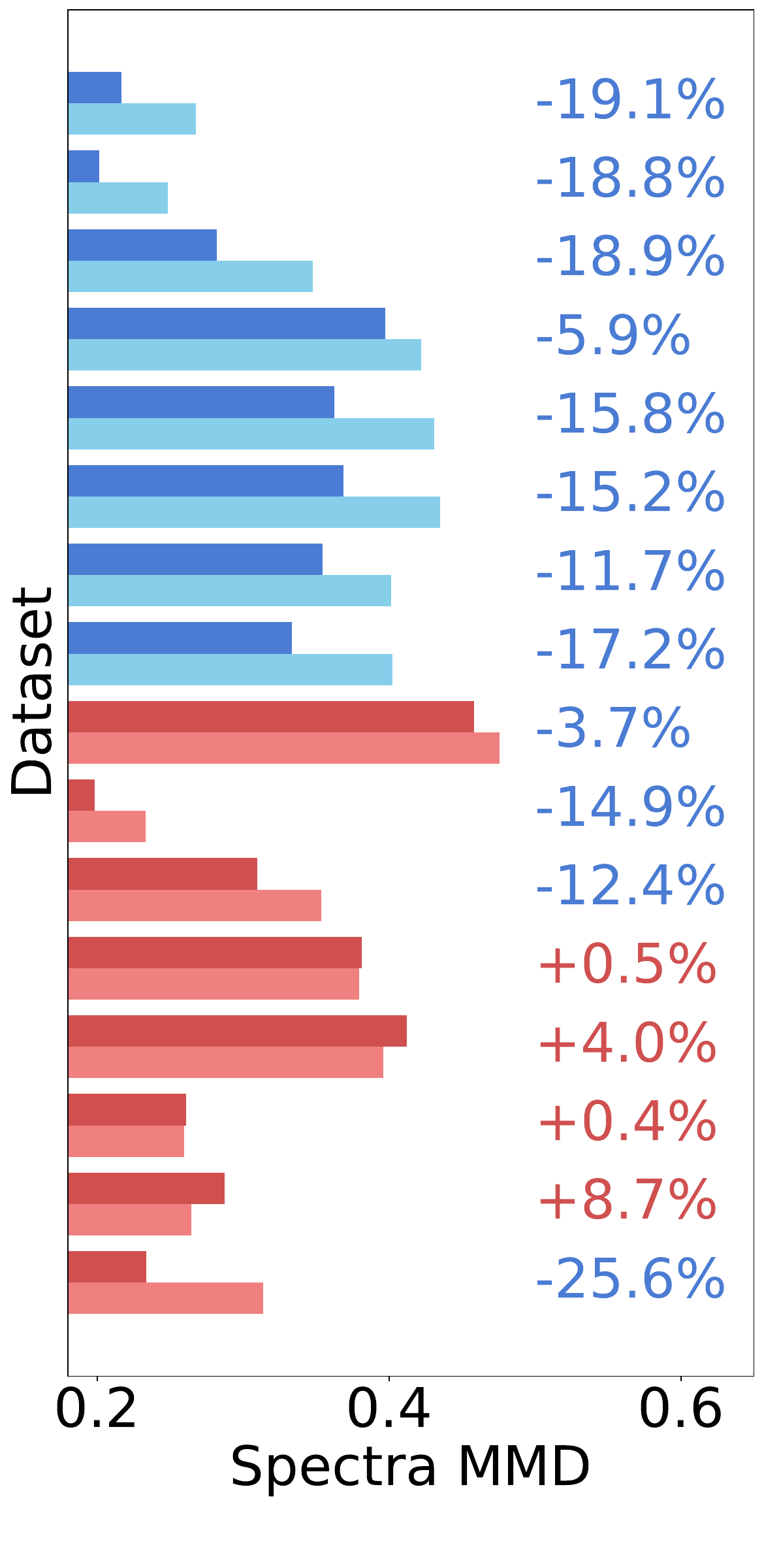}
        \caption{\textbf{Left}: Comparison of Degree MMD w/ and w/o text on GDGB datasets and DTGB datasets in DG-Gen. \textbf{Right}: Comparison of Spectra MMD w/ and w/o text on GDGB datasets and DTGB datasets in DG-Gen.}
        \label{fig:dggen_degree_spectra_mmd}
    \end{minipage}  
    \hfill
  \begin{minipage}[b]{0.48\linewidth}
        \includegraphics[width=0.3693\linewidth]{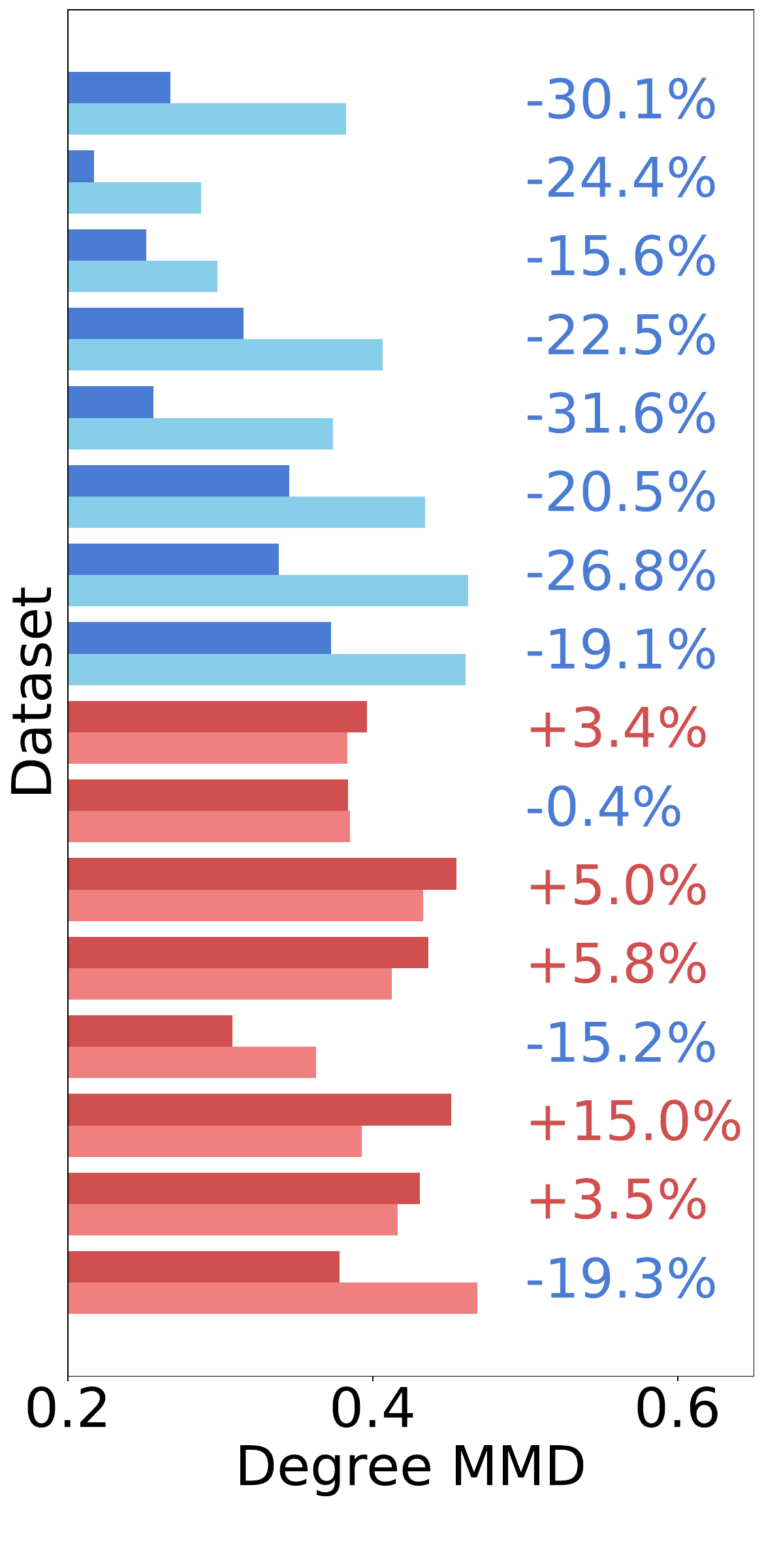}
        \includegraphics[width=0.62\linewidth]{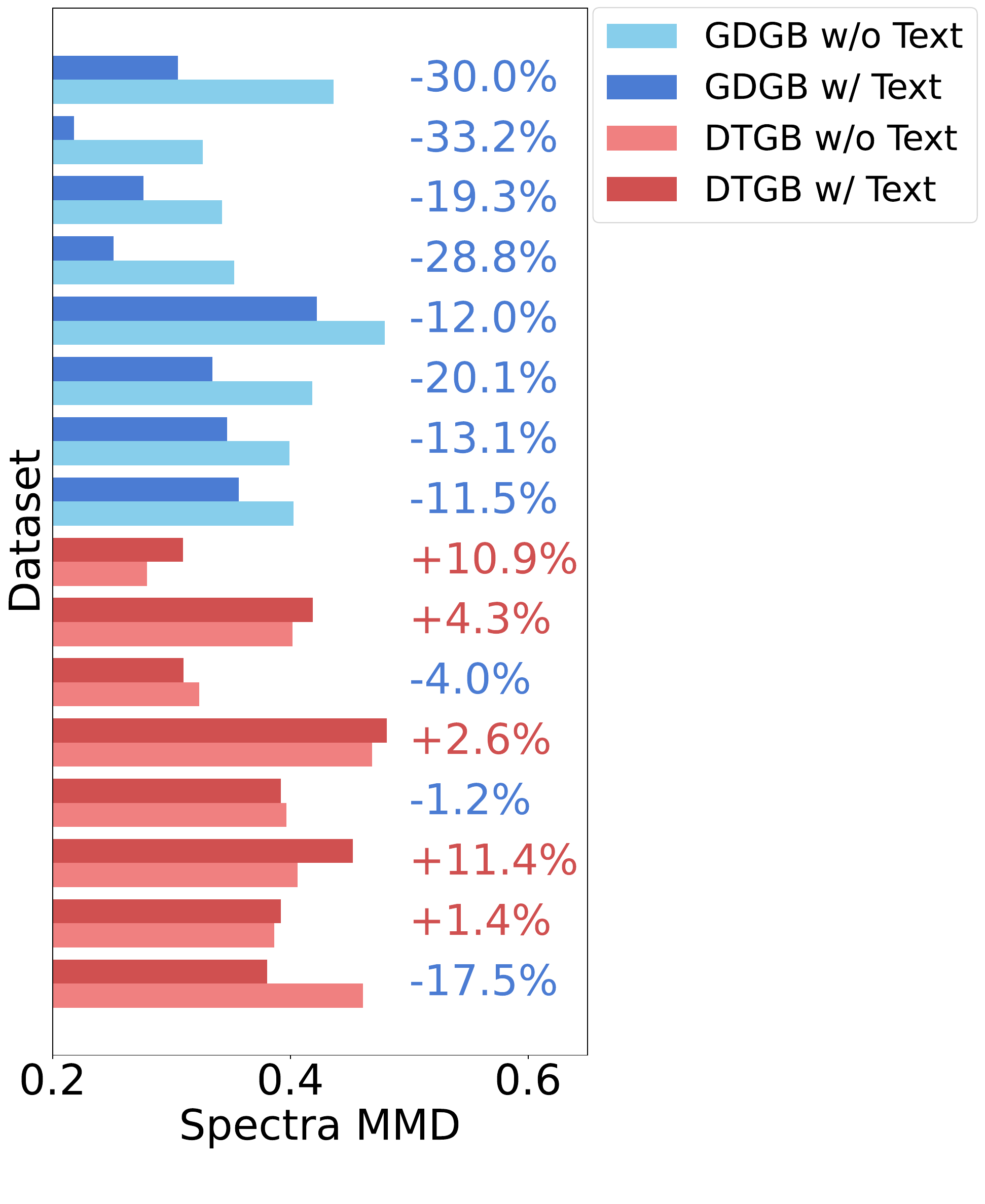}
        \caption{\textbf{Left}: Comparison of Degree MMD w/ and w/o text on GDGB datasets and DTGB datasets in VRDAG. \textbf{Right}: Comparison of Spectra MMD w/ and w/o text on GDGB datasets and DTGB datasets in VRDAG. }
        \label{fig:vrdag_degree_spectra_mmd}
    \end{minipage}  
\end{figure}

\paragraph{Textual Information Richness and Functional Utility.}
As illustrated in \cref{fig:node_edge_text_len}, we compare the average text length of node and edge attributes across datasets in GDGB and DTGB, demonstrating that GDGB datasets exhibit significantly richer textual content. 
However, text length alone does not guarantee utility in generative modeling. 
To assess the functional value of textual features, we evaluate two feature-aware dynamic graph generation models—VRDAG~\citep{VRDAG} and DG-Gen~\citep{DGGEN}—on both benchmarks, with and without the incorporation of BERT-based textual embeddings. 
As shown in \cref{fig:dggen_degree_spectra_mmd,fig:vrdag_degree_spectra_mmd}, textual features consistently \textbf{improve} generation quality on GDGB, as evidenced by reduced Degree and Spectra MMD scores. 
In contrast, on DTGB, the inclusion of textual features often \textbf{degrades} performance—particularly for VRDAG, which is highly dependent on node feature inputs. 
This suggests that the textual attributes in DTGB are semantically impoverished or noisy, and thus detrimental to generative tasks, whereas GDGB's textual content is both meaningful and effectively leveraged by learning models.

\paragraph{Intrinsic Text Quality Evaluation.}
To directly evaluate the linguistic quality of node and edge texts, we conduct intrinsic analysis using two complementary metrics:
(i) \textbf{Perplexity (PPL)}, where lower values indicate higher fluency and coherence with natural language distributions; and
(ii) \textbf{LLM-based human-like scoring} (on a 1--5 scale) across five dimensions: \textit{Contextual Fidelity}, \textit{Personality Depth}, \textit{Dynamic Adaptability}, \textit{Immersive Quality}, and \textit{Content Richness}.
The aggregated results are summarized in \cref{tab:intrinsic_text}. 
GDGB exhibits significantly lower perplexity (average value less than 90), indicating that its textual content is more coherent and aligned with natural language patterns. 
Furthermore, LLM evaluations confirm the superior semantic quality of GDGB, with average scores exceeding 4.0 across both node and edge texts. 
In contrast, DTGB achieves notably lower scores for node texts (average: 3.03), particularly in datasets such as Enron, where node attributes often consist of identifiers (e.g., email addresses) lacking meaningful semantics.

\begin{table}[ht]
\centering
\caption{Intrinsic text quality comparison (averaged across datasets).}
\label{tab:intrinsic_text}
\scalebox{0.85}{
\begin{tabular}{l|c|c}
\toprule
\textbf{Metric} & \textbf{DTGB} & \textbf{GDGB} \\
\midrule
PPL of node texts $\downarrow$ & 193.58 & \textbf{75.95} \\
PPL of edge texts $\downarrow$ & 181.82 & \textbf{88.17} \\
LLM rating (node texts) $\uparrow$ & 3.03 & \textbf{4.21} \\
LLM rating (edge texts) $\uparrow$ & 3.55 & \textbf{4.17} \\
\bottomrule
\end{tabular}
}
\end{table}

\paragraph{Generative Performance on GAG-General.}
To further assess the suitability of each benchmark for DyTAG generation, we apply the newly proposed GAG-General framework to {all} datasets from DTGB and {all datasets} from GDGB, using GPT as the LLM backbone under {both TDGG and IDGG tasks}. 
All models are evaluated under identical configurations in \cref{app:Implementation Details} to ensure fair comparison. 
{The results under TDGG are presented in \cref{tab:struct_metrics}, \ref{tab:text_metrics}, and \ref{tab:embed_metrics}.}
{While, the results under IDGG are presented in \cref{tab:struct_metrics_idgg}, \ref{tab:text_metrics_idgg}, and \ref{tab:embed_metrics_idgg}.}
The above results reveal consistent advantages for GDGB across structural, textual, and embedding fidelity metrics. 
For instance, in the graph embedding metric {under TDGG (\cref{tab:embed_metrics})}, DTGB results predominantly fall within the 0.2--0.4 range, while GDGB achieves substantially greater similarity (0.4--0.7), indicating better preservation of global structural and semantic properties. 
{
An even more pronounced trend is observed under IDGG: none of the DyTAGs generated on DTGB datasets satisfy the power-law validity criterion (\cref{tab:struct_metrics_idgg}), and both average textual quality scores (\cref{tab:text_metrics_idgg}) and graph embedding metrics (\cref{tab:embed_metrics_idgg}) are generally lower than those achieved on GDGB datasets. This again highlights the limitations of DTGB’s textual content in effectively supporting generative DyTAG tasks.}
Although the integration of memory and reflection mechanisms in GAG-General yields improvements on DTGB, these gains are constrained by the underlying data limitations, underscoring the dataset's inherent constraints for generative tasks.

\begin{table}[ht]
\centering
\caption{The results of graph structural quality under TDGG with GPT as the LLM backbone.}
\label{tab:struct_metrics}
\scalebox{0.85}{
\begin{tabular}{l|l|c|c|c|c|c}
\toprule
\textbf{Benchmark} & \textbf{Dataset} & \textbf{Degree MMD} $\downarrow$ & \textbf{Spectra MMD} $\downarrow$ & \textbf{$D_k$} & \textbf{$\alpha$} & \textbf{Power-law Validity} \\
\midrule
\multirow{8}{*}{\textbf{DTGB}} 
& Enron        & 0.331 & 0.266 & 0.125 & 1.982 & $\times$ \\
& GDELT        & 0.252 & 0.326 & 0.174 & 1.829 & $\times$ \\
& {ICEWS1819} & {0.302}& {0.341}& {0.132}& {1.794}& {$\times$ }\\
& Stack elec   & 0.228 & 0.345 & 0.133 & 1.822 & $\times$ \\
& {Stack ubuntu} & {0.235}& {0.359}& {0.130}& {1.901}& {$\times$} \\
& Googlemap CT & 0.255 & 0.375 & 0.080 & 2.278 & \checkmark \\
& {Amazon movies} & {0.404}& {0.350}& {0.182}& {1.636}& {$\times$} \\
& {Yelp} & {0.213}& {0.224}& {0.121}& {2.033}& {\checkmark}\\
\midrule
\multirow{8}{*}{\textbf{GDGB}} 
& Sephora      & 0.023 & 0.011 & 0.143 & 2.993 & \checkmark \\
& {Dianping} & {0.055} & {0.328} & {0.041} & {2.234} & {$\checkmark$} \\
& {WikiRevision} & {0.108} & {0.156} & {0.056} & {2.041} & {$\checkmark$} \\
& WikiLife     & 0.181 & 0.223 & 0.099 & 2.204 & \checkmark \\
& {IMDB} & {0.278} & {0.316} & {0.135} & {1.720} & {$\times$} \\
& WeiboTech    & 0.243 & 0.297 & 0.030 & 2.011 & \checkmark \\
& {WeiboDaily} & {0.247} & {0.493} & {0.048} & {1.845} & {$\times$} \\
& Cora         & 0.128 & 0.156 & 0.049 & 2.378 & \checkmark \\
\bottomrule
\end{tabular}
}
\end{table}

\begin{table}[ht]
\centering
\caption{The results on average textual quality scores under TDGG with GPT as the LLM backbone. M. and R. denote node memory and reflection mechanism, respectively.}
\label{tab:text_metrics}
\scalebox{0.85}{
\begin{tabular}{l|l|c|c|c}
\toprule
\textbf{Benchmark} & \textbf{Dataset} & \textbf{w/o M.} & \textbf{w/ M.} & \textbf{w/ M. \& R.} \\
\midrule
\multirow{8}{*}{\textbf{DTGB}} 
& Enron        & 3.65 & \underline{3.72} & \textbf{3.83} \\
& GDELT        & 3.70 & \underline{3.98} & \textbf{4.23} \\
& {ICEWS1819}    
  & {3.47} 
  & {\underline{3.87}} 
  & {\textbf{4.14}} \\
& Stack elec   & 3.68 & \underline{3.70} & \textbf{3.96} \\
& {Stack ubuntu} 
  & {3.38} 
  & {\underline{3.56}} 
  & {\textbf{3.78}} \\
& Googlemap CT & 3.10 & \underline{3.30} & \textbf{4.06} \\
& {Amazon movies} 
  & {3.78} 
  & {\underline{3.79}} 
  & {\textbf{4.11}} \\
& {Yelp} 
  & {3.80} 
  & {\underline{4.23}} 
  & {\textbf{4.33}} \\
\midrule
\multirow{8}{*}{\textbf{GDGB}} 
& Sephora      & 4.69 & \underline{4.69} & \textbf{4.77} \\
& {Dianping} 
  & {4.32} 
  & {\underline{4.34}} 
  & {\textbf{4.71}} \\
& {WikiRevision} 
  & {4.74} 
  & {\underline{4.79}} 
  & {\textbf{4.96}} \\
& WikiLife     & 4.44 & \underline{4.46} & \textbf{4.59} \\
& {IMDB} 
  & {3.91} 
  & {\underline{4.02}} 
  & {\textbf{4.44}} \\
& WeiboTech    & 4.84 & \underline{4.88} & \textbf{4.97} \\
& {WeiboDaily} 
  & {4.80} 
  & {\underline{4.92}} 
  & {\textbf{4.99}} \\
& Cora         & 3.98 & \underline{4.10} & \textbf{4.52} \\
\bottomrule
\end{tabular}
}
\end{table}

\begin{table}[ht]
\centering
\caption{The results on the graph embedding metric under TDGG with GPT as the LLM backbone.}
\label{tab:embed_metrics}
\scalebox{0.85}{
\begin{tabular}{l|l|c|c|c}
\toprule
\textbf{Benchmark} & \textbf{Dataset} & \textbf{w/o M.} & \textbf{w/ M.} & \textbf{w/ M. \& R.} \\
\midrule
\multirow{8}{*}{\textbf{DTGB}} 
& Enron        & 0.201 & \underline{0.259} & \textbf{0.400} \\
& GDELT        & 0.255 & \underline{0.348} & \textbf{0.393} \\
& {ICEWS1819}    
  & {0.198} 
  & {\underline{0.289}} 
  & {\textbf{0.383}} \\
& Stack elec   & 0.235 & \textbf{0.319} & \underline{0.291} \\
& {Stack ubuntu} 
  & {0.223} 
  & {\underline{0.327}} 
  & {\textbf{0.330}} \\
& Googlemap CT & 0.271 & \underline{0.302} & \textbf{0.462} \\
& {Amazon movies} 
  & {0.303} 
  & {\underline{0.359}} 
  & {\textbf{0.427}} \\
& {Yelp} 
  & {0.279} 
  & {\underline{0.396}} 
  & {\textbf{0.411}} \\
\midrule
\multirow{8}{*}{\textbf{GDGB}} 
& Sephora      & 0.588 & \textbf{0.671} & \underline{0.654} \\
& {Dianping} 
  & {0.351} 
  & {\underline{0.368}} 
  & {\textbf{0.369}} \\
& {WikiRevision} 
  & {\underline{0.438}} 
  & {0.422} 
  & {\textbf{0.454}} \\
& WikiLife     & \underline{0.459} & 0.453 & \textbf{0.463} \\
& {IMDB} 
  & {\underline{0.432}} 
  & {0.420} 
  & {\textbf{0.435}} \\
& WeiboTech    & 0.327 & \underline{0.488} & \textbf{0.698} \\
& {WeiboDaily} 
  & {0.681} 
  & {\underline{0.689}} 
  & {\textbf{0.694}} \\
& Cora         & \textbf{0.511} & 0.446 & \underline{0.465} \\
\bottomrule
\end{tabular}
}
\end{table}

\begin{table}[ht]
\centering
\caption{{The results of graph structural quality under IDGG with GPT as the LLM backbone.}}
\label{tab:struct_metrics_idgg}
\scalebox{0.85}{
\begin{tabular}{l|l|c|c|c|c|c}
\toprule
\textbf{Benchmark} & \textbf{Dataset} & \textbf{Degree MMD} $\downarrow$ & \textbf{Spectra MMD} $\downarrow$ & \textbf{$D_k$} & \textbf{$\alpha$} & \textbf{Power-law Validity} \\
\midrule
\multirow{8}{*}{\textbf{DTGB}} 
& Enron        & 0.412 & 0.378 & 0.132 & 1.742 & $\times$ \\
& GDELT        & 0.489 & 0.456 & 0.167 & 1.891 & $\times$ \\
& {ICEWS1819} 
  & {0.503} 
  & {0.412} 
  & {0.184} 
  & {1.623} 
  & {$\times$} \\
& Stack elec   & 0.537 & 0.459 & 0.181 & 1.654 & $\times$ \\
& {Stack ubuntu} 
  & {0.521} 
  & {0.488} 
  & {0.195} 
  & {1.577} 
  & {$\times$} \\
& Googlemap CT & 0.398 & 0.342 & 0.112 & 1.803 & $\times$ \\
& {Amazon movies} 
  & {0.476} 
  & {0.401} 
  & {0.143} 
  & {1.668} 
  & {$\times$} \\
& {Yelp} 
  & {0.355} 
  & {0.317} 
  & {0.078} 
  & {1.921} 
  & {$\times$} \\
\midrule
\multirow{8}{*}{\textbf{GDGB}} 
& Sephora      & 0.454 & 0.189 & 0.147 & 2.057 & \checkmark \\
& {Dianping}     
  & {0.300} 
  & {0.453} 
  & {0.069} 
  & {2.430} 
  & {\checkmark} \\
& {WikiRevision} 
  & {0.267} 
  & {0.229} 
  & {0.045} 
  & {2.050} 
  & {\checkmark} \\
& WikiLife     & 0.083 & 0.208 & 0.099 & 2.204 & \checkmark \\
& {IMDB} 
  & {0.321} 
  & {0.423} 
  & {0.252} 
  & {1.746} 
  & {$\times$} \\
& WeiboTech    & 0.268 & 0.201 & 0.064 & 1.867 & $\times$ \\
& {WeiboDaily} 
  & {0.268} 
  & {0.439} 
  & {0.156} 
  & {1.732} 
  & {$\times$} \\
& Cora         & 0.136 & 0.244 & 0.091 & 2.250 & \checkmark \\
\bottomrule
\end{tabular}
}
\end{table}

\begin{table}[ht]
\centering
\caption{{The results on average textual quality scores under IDGG with GPT as the LLM backbone. M. and R. denote node memory and reflection mechanism, respectively.}}
\label{tab:text_metrics_idgg}
\scalebox{0.85}{
\begin{tabular}{l|l|c|c|c}
\toprule
\textbf{Benchmark} & \textbf{Dataset} & \textbf{w/o M.} & \textbf{w/ M.} & \textbf{w/ M. \& R.} \\
\midrule
\multirow{8}{*}{\textbf{DTGB}} 
& Enron        & 3.51 & \underline{3.89} & \textbf{3.93} \\
& GDELT        & 3.62 & \underline{3.82} & \textbf{4.02} \\
& {ICEWS1819}    
  & {3.59} 
  & {\underline{3.90}} 
  & {\textbf{4.25}} \\
& Stack elec   & 3.23 & \underline{3.46} & \textbf{3.64} \\
& {Stack ubuntu} 
  & {3.19} 
  & {\underline{3.47}} 
  & {\textbf{3.73}} \\
& Googlemap CT & 3.22 & \underline{3.42} & \textbf{3.87} \\
& {Amazon movies} 
  & {3.63} 
  & {\underline{3.75}} 
  & {\textbf{4.22}} \\
& {Yelp} 
  & {3.53} 
  & {\underline{4.03}} 
  & {\textbf{4.16}} \\
\midrule
\multirow{8}{*}{\textbf{GDGB}} 
& Sephora      & 4.58 & \underline{4.77} & \textbf{4.87} \\
& {Dianping} 
  & {4.56} 
  & {\underline{4.71}} 
  & {\textbf{4.86}} \\
& {WikiRevision} 
  & {4.39} 
  & {\underline{4.54}} 
  & {\textbf{4.71}} \\
& WikiLife     & 4.28 & \underline{4.39} & \textbf{4.47} \\
& {IMDB} 
  & {4.19} 
  & {\underline{4.29}} 
  & {\textbf{4.49}} \\
& WeiboTech    & 4.60 & \underline{4.71} & \textbf{4.93} \\
& {WeiboDaily} 
  & {4.74} 
  & {\underline{4.83}} 
  & {\textbf{4.99}} \\
& Cora         & 4.27 & \underline{4.36} & \textbf{4.57} \\
\bottomrule
\end{tabular}
}
\end{table}

\begin{table}[ht]
\centering
\caption{{The results on the graph embedding metric under IDGG with GPT as the LLM backbone.}}
\label{tab:embed_metrics_idgg}
\scalebox{0.85}{
\begin{tabular}{l|l|c|c|c}
\toprule
\textbf{Benchmark} & \textbf{Dataset} & \textbf{w/o M.} & \textbf{w/ M.} & \textbf{w/ M. \& R.} \\
\midrule
\multirow{8}{*}{\textbf{DTGB}} 
& Enron        & 0.221 & \underline{0.278} & \textbf{0.387} \\
& GDELT        & 0.242 & \underline{0.359} & \textbf{0.388} \\
& {ICEWS1819}    
  & {0.234} 
  & {\underline{0.297}} 
  & {\textbf{0.361}} \\
& Stack elec   & 0.243 & \underline{0.289} & \textbf{0.321} \\
& {Stack ubuntu} 
  & {0.239} 
  & {\textbf{0.319}} 
  & {\underline{0.309}} \\
& Googlemap CT & 0.289 & \textbf{0.333} & \underline{0.330} \\
& {Amazon movies} 
  & {0.294} 
  & {\underline{0.338}} 
  & {\textbf{0.396}} \\
& {Yelp} 
  & {0.281} 
  & {\underline{0.369}} 
  & {\textbf{0.417}} \\
\midrule
\multirow{8}{*}{\textbf{GDGB}} 
& Sephora      & 0.601 & \underline{0.603} & \textbf{0.628} \\
& {Dianping} 
  & {\underline{0.521}} 
  & {\textbf{0.527}} 
  & {0.505} \\
& {WikiRevision} 
  & {0.589} 
  & {\underline{0.614}} 
  & {\textbf{0.629}} \\
& WikiLife     & 0.510 & \textbf{0.534} & \underline{0.531} \\
& {IMDB} 
  & {0.521} 
  & {\underline{0.535}} 
  & {\textbf{0.563}} \\
& WeiboTech    & 0.501 & \underline{0.514} & \textbf{0.536} \\
& {WeiboDaily} 
  & {0.459} 
  & {\underline{0.481}} 
  & {\textbf{0.502}} \\
& Cora         & 0.541 & \underline{0.552} & \textbf{0.572} \\
\bottomrule
\end{tabular}
}
\end{table}

\paragraph{Summary.}
Collectively, the experimental results demonstrate that GDGB significantly outperforms DTGB across multiple dimensions of generative evaluation. The superior performance on structural metrics, textual quality, and embedding similarity highlights GDGB's enhanced semantic richness and compatibility with generative modeling frameworks. 
In contrast, the limited and often noisy textual content in DTGB hinders its effectiveness in supporting high-fidelity DyTAG generation. 
Therefore, GDGB is not merely an extension of DTGB with longer text, but a \textbf{purpose-built benchmark for generative DyTAG tasks}, where high-quality textual attributes are a \textbf{core design principle}. These findings establish GDGB as a more suitable foundation for advancing research in generative modeling of dynamic text-attributed graphs.

\section{Details of Tasks and Metrics}

\subsection{Details of GAG-General}\label{app:gag-general}
GAG-General is a generalized agent-based framework designed for both TDGG and IDGG within the DyTAG paradigm. It extends the existing GAG framework \citep{GAG} through key architectural and procedural generalizations, enabling broader applicability and domain adaptability. The full procedure is specified in \cref{alg:tdgg,alg:idgg}; here we detail its design and core distinctions from GAG.

The framework proceeds in three stages. First, generalized node formulation initializes node agents, each optionally equipped with Node Memory and a Reflection Mechanism to maintain and summarize interaction history. Unlike GAG, this stage supports both bipartite and non-bipartite graph structures, allowing modeling of diverse network types.
Second, during abstracted interaction simulation, agents generate interactions (or new nodes in IDGG) over sequential rounds. Guided by an LLM backbone and informed by memory and reflection, agents produce textual content such as relations and attributes. Crucially, this workflow is abstracted from domain-specific rules, enabling seamless transfer across datasets without customization—a significant departure from GAG’s rigid, task-specific designs.
Finally, in the graph projection and evaluation phase, all generated elements are compiled into a dynamic graph and evaluated using our multi-dimensional metrics.

Hence, the key generalizations over GAG lie in three aspects. 1) GAG-General removes the bipartite constraint, supporting arbitrary graph topologies. 2) It replaces GAG’s domain-specific simulation logic with a unified, abstracted process, enhancing reusability across domains. 3) While GAG focuses on a single generation mode, GAG-General explicitly supports both TDGG and IDGG, capturing distinct dynamics of temporal evolution and structural growth.

In sum, GAG-General serves as a demonstration framework that validates the feasibility of generative DyTAG modeling. Its generalizations extend beyond the original GAG’s limitations, offering a more flexible foundation for future work. While our primary contribution is the GDGB benchmark—encompassing datasets, tasks, and metrics—GAG-General illustrates how such benchmarks can drive the development of more generalizable generative frameworks.

\subsection{Details of TDGG and IDGG on GAG-General}
\label{app:tdgg_idgg_details}
{\textbf{Transductive Dynamic Graph Generation.}}  
TDGG operates on a fixed node set $\mathcal{N}$, evolving from an initial seed graph $\mathcal{G}_0 = (\mathcal{N}, \mathcal{E}^0)$ to a final graph $\mathcal{G}_K = (\mathcal{N}, \mathcal{E}^K )$ after $K$ rounds. For instance, during the $k$-th round ($1 \leq k \leq K$), the DyTAG generation process includes the following three stages:
\begin{enumerate}
    \item \textbf{Node Update}: Based on the given source node set $\mathcal{N}_{\text{src}}^{k-1}$ and destination node set $\mathcal{N}_{\text{dst}}^{k-1}$ from the last round, we first update the node sets as $\mathcal{N}_{\text{src}}^k = \mathcal{N}_{\text{src}}^{k-1} \cup \widetilde{\mathcal{N}_{\text{src}}^k}$ and $\mathcal{N}_{\text{dst}}^k = \mathcal{N}_{\text{dst}}^{k-1} \cup \widetilde{\mathcal{N}_{\text{dst}}^k}$, where $\widetilde{\mathcal{N}_{\text{src}}^k}$ and $\widetilde{\mathcal{N}_{\text{dst}}^k}$ are obtained based on the original DyTAG with a pre-defined size.
    We then initialize active source nodes as $\mathcal{N}_{\text{src}}^{k-\text{active}} = \widetilde{\mathcal{N}_{\text{src}}^k}$ for the following transducitve generation.
    \item \textbf{Node Selection}: According to $\mathcal{N}_{\text{src}}^{k-\text{active}}$, we activate source node agents to perform pairwise interactions. 
    LLM-based agents recall and select final destination nodes $\mathcal{N}_{\text{dst}}^{k-\text{select}}$ from $\mathcal{N}_{\text{dst}}^k$ by analyzing the source node's textual profile and its memories. The memory reflection mechanism is optionally used to summarize the memories.
    \item \textbf{Interaction Generation}: After the destination node selection, the active source node agents in the $k$-th round generate a new edge set $\widetilde{\mathcal{E}^k}$ (between $\mathcal{N}_{\text{src}}^{k-\text{active}}$ and $\mathcal{N}_{\text{dst}}^{k-\text{select}}$), including edge labels $\mathcal{L}_{u,v}$ and edge texts $\mathcal{E}_{u,v}^{\text{text}}$. We then update the edge set as $\mathcal{E}^k = \mathcal{E}^{k-1} \cup \widetilde{\mathcal{E}^k}$.
\end{enumerate}

The destination node selection and edge generation stages in TDGG naturally integrate traditional discriminative tasks (e.g., node retrieval, edge classification) with LLM-driven text understanding and generation, forming a DyTAG-specific generative paradigm. 
The pseudocode of TDGG on GAG-General is shown in \cref{alg:tdgg}.

\begin{algorithm}[H]
\caption{Transductive Dynamic Graph Generation}
\label{alg:tdgg}
\begin{algorithmic}[1]
\REQUIRE Initial seed DyTAG $\mathcal{G}_0 = ( \mathcal{N}, \mathcal{E}^0 )$, rounds $K$, edge generation size $S$
\ENSURE Final generated DyTAG $\mathcal{G}_K = ( \mathcal{N}, \mathcal{E}^K )$
\STATE Initialize $\mathcal{N}_{\text{src}}^0 = \mathcal{N}_{\text{src}}^{\text{init}}, \mathcal{N}_{\text{dst}}^0 = \mathcal{N}_{\text{dst}}^{\text{init}}, \mathcal{E}^0 = \mathcal{E}^0_{\text{init}}$
\FOR{$k = 1$ to $K$}
    \STATE \textbf{Node Update}: 
    \STATE $\widetilde{\mathcal{N}_{\text{src}}^k} = \text{Sample}(\mathcal{N}, S), \widetilde{\mathcal{N}_{\text{dst}}^k} = \text{Sample}(\mathcal{N}, S)$
    \STATE $\mathcal{N}_{\text{src}}^k = \mathcal{N}_{\text{src}}^{k-1} \cup \widetilde{\mathcal{N}_{\text{src}}^k}, \mathcal{N}_{\text{dst}}^k = \mathcal{N}_{\text{dst}}^{k-1} \cup \widetilde{\mathcal{N}_{\text{dst}}^k}$
    \STATE $\mathcal{N}_{\text{src}}^{k-\text{active}} = \widetilde{\mathcal{N}_{\text{src}}^k}$
    \STATE \textbf{Node Selection}: 
    \STATE $\mathcal{N}_{\text{dst}}^{k-\text{select}} = \text{Recall, Select}(\mathcal{N}_{\text{src}}^{k-\text{active}}, \mathcal{N}_{\text{dst}}^k)$
    \STATE \textbf{Interaction Generation}: 
    \STATE $\widetilde{\mathcal{E}^k} = \text{GenerateEdges}(\mathcal{N}_{\text{src}}^{k-\text{active}},\mathcal{N}_{\text{dst}}^{k-\text{select}})$
    \STATE $\mathcal{E}^k = \mathcal{E}^{k-1} \cup \widetilde{\mathcal{E}^k}$
\ENDFOR
\RETURN $\mathcal{G}_K = ( \mathcal{N}, \mathcal{E}^K )$
\end{algorithmic}
\end{algorithm}

{\textbf{Inductive Dynamic Graph Generation.}}  
IDGG extends TDGG by simultaneously generating new nodes and edges. Starting from seed graph $\mathcal{G}_0 = ( \mathcal{N}^0, \mathcal{E}^0 )$, it evolves to $\mathcal{G}_K = ( \mathcal{N}^K, \mathcal{E}^K )$ through $K$ rounds.
Key steps for the $k$-th round ($1 \leq k \leq K$) include:  
\begin{enumerate}
    \item \textbf{Node Generation}: We apply node generator agents to generate $\widetilde{\mathcal{N}_{\text{src}}^{k}}$ (size of $R_{\text{src}}$) and $\widetilde{\mathcal{N}_{\text{dst}}^{k}}$ (size of $R_{\text{dst}}$) along with their textual features, according to the recent active nodes. $R_{\text{src}}$ and $R_{\text{dst}}$ are predefined by the seed graph based on the average numbers of new source nodes and new destination nodes per round.
    \item \textbf{Node Update}: After node generation, we update the node sets as $\mathcal{N}_{\text{src}}^k = \mathcal{N}_{\text{src}}^{k-1} \cup \widetilde{\mathcal{N}_{\text{src}}^{k}}$ and $\mathcal{N}_{\text{dst}}^k = \mathcal{N}_{\text{dst}}^{k-1} \cup \widetilde{\mathcal{N}_{\text{dst}}^{k}}$.
    We then initialize active source nodes as $\mathcal{N}_{\text{src}}^{k-\text{active}} = \text{RndSample}(\mathcal{N}_{\text{src}}^k)$ for the following inductive generation, where $\text{RndSample}()$ is a random sampling function.
    \item \textbf{Node Selection \& Interaction Generation}: Similar to TDGG, active source node agents recall and select destination nodes $\mathcal{N}_{\text{dst}}^{k-\text{select}}$ from $\mathcal{N}_{\text{dst}}^k$, and generate $\widetilde{\mathcal{E}^k}$ (between $\mathcal{N}_{\text{src}}^{k-\text{active}}$ and $\mathcal{N}_{\text{dst}}^{k-\text{select}}$), including edge labels $\mathcal{L}_{u,v}$ and edge texts $\mathcal{E}_{u,v}^{\text{text}}$. We then update the edge set as $\mathcal{E}^k = \mathcal{E}^{k-1} \cup \widetilde{\mathcal{E}^k}$.
\end{enumerate}

The IDGG above faithfully replicates the evolutionary dynamics of real-world DyTAG through new node and edge generation over time.
Although inherently challenging, this paradigm establishes a foundational direction for advancing dynamic graph generation research by simulating open-ended growth processes.
The pseudocode of IDGG on GAG-General is shown in \cref{alg:idgg}.
The source/destination node generation counts $R_{\text{src}}/R_{\text{dst}}$ for IDGG are shown in \cref{tab:src_dst_node_generaion_counts}). 

\begin{algorithm}[H]
\caption{Inductive Dynamic Graph Generation}
\label{alg:idgg}
\begin{algorithmic}[1]
\REQUIRE Initial seed DyTAG $\mathcal{G}_0 = ( \mathcal{N}^0, \mathcal{E}^0 )$, rounds $K$, edge generation size $S$, source/destination node generation numbers $R_{\text{src}}, R_{\text{dst}}$
\ENSURE Final generated DyTAG $\mathcal{G}_K = ( \mathcal{N}^K, \mathcal{E}^K )$
\STATE $\mathcal{N}_{\text{src}}^0 = \mathcal{N}_{\text{src}}^{\text{init}}, \mathcal{N}_{\text{dst}}^0 = \mathcal{N}_{\text{dst}}^{\text{init}}, \mathcal{E}^0 = \mathcal{E}^0_{\text{init}}$
\FOR{$k = 1$ to $K$}
    \STATE \textbf{Node Generation}: 
    \STATE $\widetilde{\mathcal{N}_{\text{src}}^{k}} = \text{GenerateNodes}(R_{\text{src}}), \widetilde{\mathcal{N}_{\text{dst}}^{k}} = \text{GenerateNodes}(R_{\text{dst}})$
    \STATE \textbf{Node Update}: 
    \STATE $\mathcal{N}_{\text{src}}^k = \mathcal{N}_{\text{src}}^{k-1} \cup \widetilde{\mathcal{N}_{\text{src}}^{k}}, \mathcal{N}_{\text{dst}}^k = \mathcal{N}_{\text{dst}}^{k-1} \cup \widetilde{\mathcal{N}_{\text{dst}}^{k}}$
    \STATE $\mathcal{N}_{\text{src}}^{k-\text{active}} = \text{RndSample}(\mathcal{N}_{\text{src}}^k, S)$
    \STATE \textbf{Node Selection}: 
    \STATE $\mathcal{N}_{\text{dst}}^{k-\text{select}} = \text{Recall, Select}(\mathcal{N}_{\text{src}}^{k-\text{active}}, \mathcal{N}_{\text{dst}}^k)$
    \STATE \textbf{Interaction Generation}: 
    \STATE $\widetilde{\mathcal{E}^k} = \text{GenerateEdge}(\mathcal{N}_{\text{src}}^{k-\text{active}},\mathcal{N}_{\text{dst}}^{k-\text{select}})$
    \STATE $\mathcal{E}^k = \mathcal{E}^{k-1} \cup \widetilde{\mathcal{E}^k}$
\ENDFOR
\RETURN $\mathcal{G}_K = ( \mathcal{N}^K, \mathcal{E}^K )$
\end{algorithmic}
\end{algorithm}

\begin{table}[h]
    \centering
    \caption{The source node generation counts $R_{src}$ and the destination node generation counts $R_{dst}$, which are determined by the seed graph.}
    \resizebox{1.0\linewidth}{!}{
    \begin{tabular}{c|cccccccc}
    \toprule
         Dataset&  Sephora&  Dianping&  WikiRevision&  WikiLife&  IMDB&  WeiboTech&  WeiboDaily& Cora\\
         \midrule
         $R_{src}$&  27&  33&  6&  13&  17&  3&  6& 32\\
         $R_{dst}$&  4&  44&  37&  12&  17&  29&  19& 18\\
         \bottomrule
    \end{tabular}}
    \label{tab:src_dst_node_generaion_counts}
\end{table}

\subsection{Details of Metrics}
\label{app:metrics}
We provide more detailed mathematical formulations and implementation specifics of our evaluation metrics in \cref{sec:metric} as follows.

\textbf{Graph Structural Metrics.} 
We evaluate the structural quality of generated DyTAGs using two classical approaches:  
\begin{itemize}
    \item \textbf{Degree/Spectra MMD.} 
    This metric quantifies the discrepancy between the distribution of graph descriptors (e.g., degree/spectral properties) in the generated and ground-truth graphs \citep{MMD,graphrnn,EDGE}. 
    We compute the distribution distance, using the maximum mean discrepancy (MMD) with a radial basis function (RBF) kernel. 
    Thus, lower MMD values indicate higher structural fidelity. 
    Specifically, we employ a positive definite RBF kernel with a smoothing parameter $v$, defined as:
    $$
    k\left(x_i, x_j\right)=\exp \left(-\frac{\left\|x_i-x_j\right\|^2}{2 v^2}\right),
    $$
    where $ x_i$ and $x_j $ are feature vectors representing graph descriptors (e.g., degrees/spectral properties) for nodes $ i $ and $ j $.
    $ v $ denotes smoothing parameter controlling the width of the RBF kernel. Larger $ v $ values reduce sensitivity to local variations. 
    $ k(\cdot, \cdot) $ denotes the RBF kernel function measuring similarity between graph descriptors.
    
    We treat nodes as samples, where $n$ and $m$ denote the number of nodes in the generated and ground-truth graphs, respectively.
    The MMD is computed as: 
    $$
    \operatorname{MMD}^2(X, Y):=\frac{1}{n^2} \sum_{i, j=1}^n \mathrm{k}\left(x_i, x_j\right)+\frac{1}{m^2} \sum_{i, j=1}^m \mathrm{k}\left(y_i, y_j\right)-\frac{2}{n m} \sum_{i=1}^n \sum_{j=1}^m \mathrm{k}\left(x_i, y_j\right)
    $$
    where $ X = \{x_1, x_2, \dots, x_n\} $is the set of feature vectors from the generated DyTAG.
    $ Y = \{y_1, y_2, \dots, y_m\} $is the set of feature vectors from the ground-truth DyTAG.  
    
    \item \textbf{Power-law Analysis} Real-world networks often exhibit power-law degree distributions. 
    We assess power-law behavior in degree distributions using the Kolmogorov-Smirnov (KS) distance:
    $$
    D_k=\max _{k \geq k_{\min }}|F(k)-H(k)|,
    $$
    where $F(k)$ is the empirical cumulative distribution function (CDF) of the degree data, and $H(k)$ is the CDF of the best-fitting power-law model.
    $ k_{\min} $ denotes the minimum degree value used for fitting, set to $ k_{\min} = 2 $ to ensure robustness. 
    The power-law exponent $\alpha$ indicates adherence to a power-law distribution, typically valid if $\alpha \in[2,3]$ \citep{powerlaw}. 
    A DyTAG is power-law valid if $ D_k < 0.15 $ and the power-law exponent $ \alpha \in [2,3] $.  
\end{itemize}

\textbf{Textual Quality Metrics.}
Inspired by recent role-playing agent research \citep{RoleBench,SocialBench,charbox}, we employ the {LLM-as-Evaluator} framework to assess the text quality in the generated DyTAGs. 
Considering the targets and requirements of DyTAG  generation tasks, we define five scoring criteria: 
\begin{enumerate}
    \item \textbf{Contextual Fidelity}: Consistency of node/edge texts with historical interactions. 
    \item \textbf{Personality Depth}: Richness of semantic and stylistic diversity in node profiles.
    \item \textbf{Dynamic Adaptability}: Temporal coherence of evolving textual content.  
    \item \textbf{Immersive Quality}: Engagement and realism of generated narratives.  
    \item \textbf{Content Richness}: Information density and relevance of edge texts.  
\end{enumerate}
Each criterion is scored on a 1–5 scale by LLM evaluators, ensuring a multidimensional assessment of textual quality. 
{For the LLM-as-Evaluator component, we consistently employ GPT as the underlying LLM backbone for all baselines, with hyperparameters (e.g., temperature, top‑p, and repetition penalty) kept identical to those specified in \cref{app:Implementation Details}. The prompts used in the LLM-as-Evaluator framework are provided in \cref{app:prompt_llm_eval}.}
This approach outperforms embedding-based metrics like BERTScore \citep{bertscore} by avoiding information loss during feature compression.

\textbf{Graph Embedding Metric.} 
To jointly evaluate structural, temporal, and textual fidelity, we extend the JL-Metric \citep{jlmetric} to a better adaptation on DyTAGs with textual node feature integration to a graph embedding-based indicator. 
Specifically, the metric is calculated through three stages:  
\begin{itemize}
    \item \textbf{Node Embedding Construction.} For node $ j $, we concatenate textual attributes and temporal information of historical interactions into a contextualized node embedding:  
   $$
   \mathbf{v}_j = \left[ \tilde{c}(\mathcal{T}_{j,1}) \| \tilde{c}(\mathcal{T}_{j,2}) \| \dots \| \tilde{c}(\mathcal{T}_{j,m_j}) \| \mathcal{N}_j^{\text{text}} \right],
   $$  
   where $ \tilde{c}(\mathcal{T}_{j,i}) = (\mathcal{T}_{j,i}, \mathcal{E}_{j,i}^{\text{text}}(t_i), \mathcal{N}_i^{\text{text}}) $ encodes the $ i $-th  historical interaction event (including timestamp $\mathcal{T}_{j,i}$, textual edge feature $\mathcal{E}_{j,i}^{\text{text}}(t_i)$, and textual interacted node feature $\mathcal{N}_i^{\text{text}}$), and $ m_j $ is the total  historical interaction count for node $ j $. 
   Additionally, we append the textual attribute of node $ j $ as the final component of the embedding vector, denoted by $\mathcal{N}_j^{\text{text}}$.

   \item \textbf{Random Projection Mapping.} Leveraging the Johnson-Lindenstrauss theorem for transforming varying dimensionality with bounded error guarantees \citep{jlmetric}, we project high-dimensional embeddings into a consistent space with random matrices, which enables cross-graph alignment despite different node counts and interaction histories. 
   The projected graph embedding is:  
   $$
   \hat{\mathcal{G}} = \left\{ \hat{\mathbf{v}}_1, \hat{\mathbf{v}}_2, \dots, \hat{\mathbf{v}}_j , \dots\right\}, 
   $$  
   where $ \hat{\mathbf{v}}_j $ denotes the $ j $-th projected node embedding.  
   \item \textbf{Similarity Measurement.} We compute the cosine distance between the generated graph $ \hat{\mathcal{G}}_g $ and ground-truth graph $ \hat{\mathcal{G}}_o $:  
   $$
   \rho(\hat{\mathcal{G}}_g, \hat{\mathcal{G}}_o) = 1 - \frac{\langle \hat{\mathcal{G}}_g, \hat{\mathcal{G}}_o \rangle_F}{\|\hat{\mathcal{G}}_g\|_F \cdot \|\hat{\mathcal{G}}_o\|_F},
   $$  
   where $ \| \cdot \|_F $ is the Frobenius norm and $ \langle \cdot, \cdot \rangle_F $ is the Frobenius inner product. Lower $ \rho $ values indicate higher global fidelity.  
\end{itemize}

\section{Supplementary Experimental Details}
\label{app:supplementary_implement_details}
\subsection{Large Language Model Baselines}
LLMs possess strong capabilities in language understanding and generation, demonstrating impressive performance across a wide range of natural language processing tasks. To comprehensively evaluate the performance of different models, we selected several representative LLMs for comparison in our experiments. These include open-source models: DeepSeek-R1-Distill-Qwen-32B \citep{deepseek}, Llama-3-70B-Instruct \citep{llama3}, and Qwen2.5-72B-Instruct \citep{qwen2.5}. In addition, we include a closed-source model, GPT-4o-Mini \citep{gpt4omini}, as a reference baseline.

\subsection{Dynamic Graph Neural Network Baselines}

\textbf{JODIE~\citep{JODIE}.} This model is a representation learning framework for nodes in temporal networks that captures dynamic embedding trajectories based on sequences of interactions. JODIE uses two coupled recurrent neural networks to update the states of interacting entities and employs a projection operation to forecast future embedding trajectories.

\textbf{TGN~\citep{TGN}.} This method models sequences of time-stamped events. By combining memory modules with graph-based operators, TGN effectively captures temporal and structural dependencies in evolving graphs. This innovative architecture enables TGNs to maintain historical context while updating node representations in real time. 

\textbf{CAWN~\citep{CAWN}.} This approach inductively represents temporal networks through an anonymization strategy based on sampled walks. These walks explore the causal structure of network dynamics and generate inductive node representations. A neural network encodes and aggregates the sampled walks to produce the final node embeddings.

\textbf{GraphMixer~\citep{Graphmixer}.} This model is a simple yet effective architecture for temporal link prediction, composed of three key components: a multi-layer perceptron (MLP)-based link encoder to capture information from temporal links, a node encoder using neighbor mean-pooling, and an MLP-based link classifier for prediction.

\textbf{DyGFormer~\citep{DyGLib}.} This method is a Transformer-based model for dynamic graph learning that leverages nodes’ historical first-hop interactions to learn meaningful representations. It introduces a neighbor co-occurrence encoding scheme to capture correlations between source and destination nodes based on their interaction histories. To handle longer sequences efficiently, DyGFormer uses a patching technique that divides each historical sequence into smaller segments, allowing the Transformer to process long-term dependencies effectively.

\subsection{Dynamic Graph Generation Model Baselines}
\textbf{DG-Gen~\citep{DGGEN}.} This model is a generative framework for CTDG that enables assumption-free and inductive graph generation. Built as an encoder-decoder model, it learns conditional probabilities of temporal interactions to model topological evolution effectively. By relying on temporal embeddings instead of node IDs, DG-Gen supports inductive learning, allowing it to generate graphs with unseen nodes, timestamps, and edge features.

\textbf{VRDAG~\citep{VRDAG}.} This model is a novel framework for simultaneously generating dynamic graph topology and node attributes in a data-driven manner. It uses a bi-flow graph encoder to preserve directional message flow and attribute information in node embeddings. A GRU-based module captures temporal dependencies, updating hidden node states across time. VRDAG also parameterizes a flexible prior distribution to sample latent variables at future timesteps, enabling it to model complex dependencies within and across topology and attribute evolution. This variational approach effectively captures both structural and attribute dynamics in evolving graphs.

{
\textbf{TIGGER~\cite{TIGGER}.} This model is a scalable generative model for continuous-time dynamic graphs that overcomes the transductive limitations and node identity leakage of prior methods. Its inductive variant, TIGGER-I, learns a distribution over node embeddings rather than node IDs, enabling generation of graphs with unseen nodes. Notably, TIGGER-I employs a GAN-based module to jointly generate realistic features and graph structure, supporting flexible up- or down-sampling of graph size without requiring one-to-one node mapping—making it suitable for privacy-sensitive and feature-aware dynamic graph synthesis.}

\subsection{Implementation Details}\label{app:Implementation Details}
\textbf{Implementation Details of TDGG.}
For {TDGG}, we set the size of the target expanded DyTAG as 10,000 edges. 
To ensure fairness in discriminative task comparisons, we adopt a dataset split of 1,000/1,000/8,000 (train/val/test) for DGNNs based on the implementation of DyGLib \citep{DyGLib, DTGB}, and compare with the last 8,000 edges in the generated DyTAG from GAG-General. 
For edge classification, we evaluate the models using weighted Precision, Recall, and F1 scores \citep{DTGB}.
For the discriminative task comparisons in TDGG, following DTGB \citep{DTGB}, node/edge texts are encoded using the BERT-base-uncased model \citep{BERT} as initialization for each DGNN. 
In terms of discriminative task-specific metrics, we use Hit@1 and Hit@10 for node retrieval (link prediction) tasks. 
Specifically, for the node retrieval task, we set the number of negative samples to 100 for each positive sample. 
Notably, the Hit@1 metric can be considered a more challenging form of link prediction, as it involves a positive-to-negative sample ratio of 1:100, compared to the easier 1:1 ratio. 

\textbf{Implementation Details of IDGG.}
For {IDGG}, we set the size of the target expanded DyTAG as 2,000 edges, due to the additional stage of node generation. 
To ensure fair comparisons with dynamic graph generation baselines, we use the seed DyTAG with 1,000 edges as the input data for training DG-Gen and VRDAG and adjust the generation ratio in DG-Gen to generate 2,000 edges using its publicly available code \citep{DGGEN}, whereas VRDAG does not offer explicit generation size control \citep{VRDAG}. 
Therefore, evaluations are conducted based on the graphs actually generated by VRDAG. 
As the current dynamic graph generation models' lack of support for DyTAG's textual features, our focus is on comparing the quality of generated graph structures and graph embeddings, rather than the texts.
Similar to TDGG, following DTGB \citep{DTGB}, node/edge texts are encoded using the BERT-base-uncased model \citep{BERT} as initialization for each dynamic graph generation baseline.

\textbf{Implementation Details of GAG-General on TDGG and IDGG.}
Following GAG \citep{GAG}, we maintain a memory module for each node in DyTAG to record the information of its historical interacted neighbors, effectively incorporating structural and temporal dynamics from the DyTAG. 
In our proposed GAG-General, the node memory module is implemented via random walks \citep{sauerwald2019random}.
We set the number of random walks for each node as 10, with a random walk length of 10.
The node memory is capped at a maximum memory context length of 1,000.  
Additionally, our generative framework includes an optional memory reflection mechanism, which leverages the LLMs to distill node memories into valuable summaries, akin to the message aggregation process in GNNs \citep{kipf2016semi,peng2024beyond}.
During the destination node selection process for the active source node agent, for both TDGG and IDGG, we set the number of recalled candidate destination nodes for each source node to 10. 
For the inference parameter configuration of benchmarked LLMs, we set the temperature as 0.8, the top-p (nucleus sampling) as 0.9, the repetition penalty as 1.1, and a maximum token limit as 2,000 to balance randomness, diversity, and redundancy suppression.  
With regarding to the LLM prompt design for TDGG and IDGG, we incorporates detailed node descriptions, edge descriptions, memory reflection instructions, interaction instructions, and node generation instructions in our proposed GAG-General. 
The example full templates on Sephroa (bipartite) and WeiboTech (non-bipartite) are provided in \cref{app:prompt}. 
As for the LLM-as-Evaluator framework of the textual quality metric, based on the implementation of CharacterBox \citep{charbox}, we use GPT as the LLM backbone for the evaluation framework.
{
The hyperparameters (temperature, top‑p, repetition penalty, etc.) of the LLM used in the textual quality metric are kept identical to those used in generation. 
Notably, although multiple LLMs serve as backbones in GAG-General for generation, all generated outputs are evaluated by the same GPT-based evaluator with fixed hyperparameters, ensuring robustness and reproducibility of the assessment.
The prompts used in the LLM-as-Evaluator framework are provided in \cref{app:prompt_llm_eval}.}
Experiments are conducted on NVIDIA A800 with 80 GB memory.



\section{Supplementary Experimental Results}

\subsection{Scalability Analysis}  
\label{app:scalability}
Following GAG \citep{GAG}, we adopt parallel processing techniques \citep{agentscope} in implementing GAG-General to mitigate idle inference time in LLMs during DyTAG generation. 
Agents are grouped into distinct clusters: each active source node agent collaborates with its corresponding destination node agent, and these groups execute in parallel across CPU cores with multiple ports, enabling an efficient DyTAG generation pipeline.

\begin{figure}[h]
  \centering
  \vspace{-1mm}
  \includegraphics[width=0.75\linewidth]{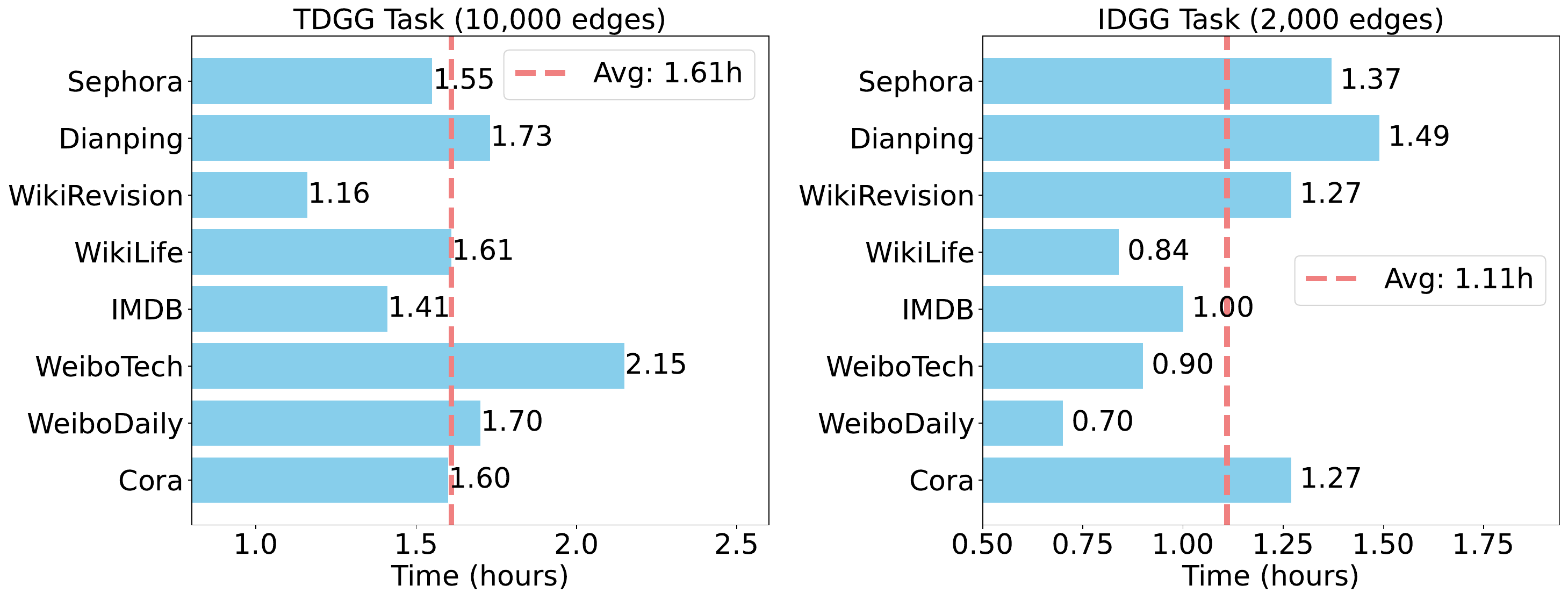}
  \caption{The time required to generate DyTAGs under TDGG (10,000 edges) and IDGG task (2,000 edges) from a 1,000-edge seed graph across GDGB datasets.}
  \vspace{-1mm}
  \label{fig:scalability}
\end{figure}

\begin{figure}[h]
  \centering
  \vspace{-1mm}
  \includegraphics[width=0.75\linewidth]{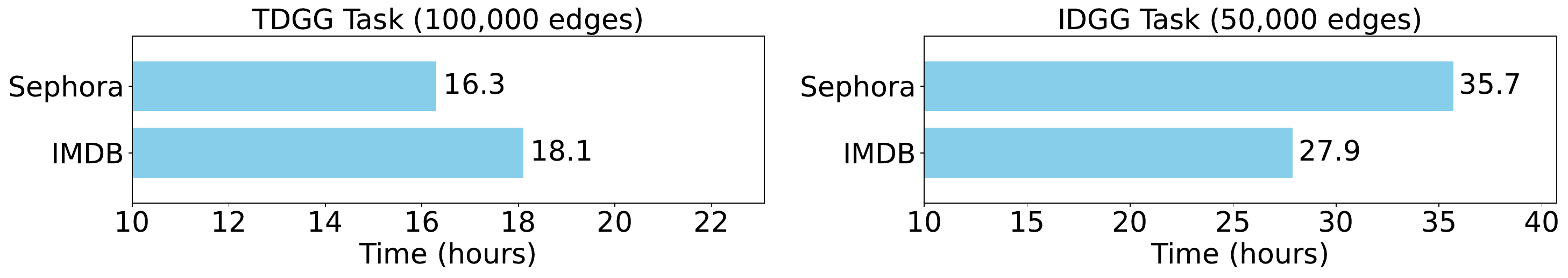}
  \caption{The time required to generate DyTAGs under TDGG (100,000 edges) and IDGG task (50,000 edges) from a 1,000-edge seed graph on Sephora and IMDB.}
  \vspace{-1mm}
  \label{fig:scalability_large}
\end{figure}

As shown in \cref{fig:scalability_large}, we evaluate the time costs of generating a DyTAG with 10,000 edges in TDGG and a DyTAG with 2,000 edges in IDGG, using a 1,000-edge seed graph across diverse datasets with GAG-General applying GPT as the LLM backbone.
Under the parallel architecture, the TDGG task achieves an average generation time of 1.61 hours, demonstrating satisfying scalability for DyTAG generation. 
For IDGG, the additional computational overhead of generating new nodes increases total time, which is jointly determined by both node and edge counts. 
Across eight datasets, the IDGG task achieves an average generation time of 1.11 hours, indicating moderate scalability with room for further optimization.
Furthermore, to rigorously assess the scalability of GAG-General, we conduct experiments on generating large-scale DyTAGs in both TDGG and IDGG tasks. As illustrated in \cref{fig:scalability}, for instance, generating a DyTAG with 100,000 edges in TDGG on Sephora takes 16.3 hours, while generating a DyTAG with 50,000 edges in IDGG requires 35.7 hours. 
Although the current time costs for large-scale DyTAG generation remain relatively high, our GDGB datasets—comprising multiple million-edge-scale DyTAGs—provide a robust foundation for future research on scalability challenges in DyTAG generation.

\subsection{Results of TDGG on Graph Structural Metrics}
Comprehensive results on Degree MMD, Spectra MMD, $D_k$, $\alpha$, and power-law validity under TDGG using three additional LLM backbones (Deepseek, Llama, and Qwen) are available in \cref{tab:tdgg_structural_deepseek}, \ref{tab:tdgg_structural_llama}, and \ref{tab:tdgg_structural_qwen}.
The results demonstrate that the generated DyTAGs on most datasets exhibit low Degree/Spectra MMD values and satisfy the power-law validity criterion, confirming high-quality generation in TDGG with GAG-General.

\begin{table}[H]
\vspace{-5mm}
\centering \caption{The results on Degree MMD, Spectra MMD, $D_k$, $\alpha$, and power-law validity under TDGG with Deepseek as the LLM backbone.} 
\label{tab:tdgg_structural_deepseek}
\resizebox{0.9\linewidth}{!}{
    \begin{tabular}{l|cccccccc}
    \toprule
        Dataset & \textbf{Sephora} & \textbf{Dianping} & \textbf{WikiRevision} & \textbf{WikiLife} & \textbf{IMDB} & \textbf{WeiboTech} & \textbf{WeiboDaily} & \textbf{Cora} \\
        \midrule
        Degree MMD$\downarrow$        & 0.050 & 0.033 & 0.209 & 0.112 & 0.295 & 0.110 & 0.123 & 0.219 \\
        Spectra MMD$\downarrow$       & 0.142 & 0.188 & 0.149 & 0.257 & 0.331 & 0.177 & 0.273 & 0.220 \\
        $D_k$            & 0.161 & 0.057 & 0.040 & 0.099 & 0.099 & 0.076 & 0.018 & 0.052 \\
        $\alpha$         & 2.382 & 2.808 & 2.246 & 2.204 & 2.065 & 2.255 & 2.028 & 2.562 \\
        Power-law Validity& \ding{55}& \ding{51}& \ding{51}& \ding{51}& \ding{51}& \ding{51}& \ding{51}& \ding{51}\\
    \bottomrule
\end{tabular}}

\end{table}

\begin{table}[H]
\centering \caption{The results on Degree MMD, Spectra MMD, $D_k$, $\alpha$, and power-law validity under TDGG with Llama as the LLM backbone.} 
\label{tab:tdgg_structural_llama}
\resizebox{0.9\linewidth}{!}{
    \begin{tabular}{l|cccccccc}
    \toprule
        Dataset & \textbf{Sephora} & \textbf{Dianping} & \textbf{WikiRevision} & \textbf{WikiLife} & \textbf{IMDB} & \textbf{WeiboTech} & \textbf{WeiboDaily} & \textbf{Cora} \\
        \midrule
        Degree MMD$\downarrow$          & 0.043 & 0.062 & 0.018 & 0.164 & 0.285 & 0.188 & 0.259 & 0.109 \\
        Spectra MMD$\downarrow$         & 0.010 & 0.399 & 0.048 & 0.217 & 0.316 & 0.257 & 0.421 & 0.126 \\
        $D_k$              & 0.135 & 0.044 & 0.108 & 0.099 & 0.125 & 0.051 & 0.039 & 0.047 \\
        $\alpha$           & 2.976 & 2.331 & 2.328 & 2.204 & 1.750 & 2.145 & 1.979 & 2.328 \\
        Power-law Validity & \ding{51}& \ding{51}& \ding{51}& \ding{51}& \ding{55}& \ding{51}& \ding{55}& \ding{51}\\
    \bottomrule
\end{tabular}}
\end{table}

\begin{table}[H]
\centering \caption{The results on Degree MMD, Spectra MMD, $D_k$, $\alpha$, and power-law validity under TDGG with Qwen as the LLM backbone. } 
\label{tab:tdgg_structural_qwen}
\resizebox{0.9\linewidth}{!}{
    \begin{tabular}{l|cccccccc}
    \toprule
        Dataset & \textbf{Sephora} & \textbf{Dianping} & \textbf{WikiRevision} & \textbf{WikiLife} & \textbf{IMDB} & \textbf{WeiboTech} & \textbf{WeiboDaily} & \textbf{Cora} \\
        \midrule
        Degree MMD$\downarrow$          & 0.034 & 0.041 & 0.098 & 0.181 & 0.248 & 0.222 & 0.293 & 0.085 \\
        Spectra MMD$\downarrow$         & 0.013 & 0.216 & 0.141 & 0.225 & 0.307 & 0.307 & 0.486 & 0.127 \\
        $D_k$              & 0.139 & 0.026 & 0.055 & 0.099 & 0.149 & 0.020 & 0.054 & 0.045 \\
        $\alpha$           & 2.978 & 2.177 & 2.065 & 2.204 & 1.723 & 2.020 & 1.844 & 2.367 \\
        Power-law Validity & \ding{51}& \ding{51}& \ding{51}& \ding{51}& \ding{55}& \ding{51}& \ding{55}& \ding{51}\\
    \bottomrule
\end{tabular}}
\end{table}

\subsection{Results of TDGG on Textual Quality Metrics}
Full results on average textual quality scores under TDGG are shown in \cref{tab:tdgg_textual_average}. 
Detailed results for each scoring criterion—\emph{Contextual Fidelity, Personality Depth, Dynamic Adaptability, Immersive Quality, and Content Richness}—rated on a 1–5 scale, are available in \cref{tab:tdgg_textual_sephora}–\ref{tab:tdgg_textual_cora}. 
Full descriptions of these criteria are provided in \cref{app:metrics}. 
The experimental results demonstrate that node memory or reflection mechanisms consistently improve textual generation quality compared to their absence, highlighting the necessity of these components and the value of integrating structural and textual information in DyTAG generation.

\begin{table}[H]
\centering \caption{The full results on \textbf{average} textual quality scores under TDGG. M. and R. denote node memory and reflection mechanism, respectively. The best and the runner-up scores are highlighted in bold and underlined fonts.} 
\label{tab:tdgg_textual_average}
\resizebox{1.0\linewidth}{!}{
\begin{tabular}{l|ccc|ccc|ccc|ccc}\toprule
 & \multicolumn{3}{c|}{\textbf{DeepSeek}} & \multicolumn{3}{c|}{\textbf{Llama}} & \multicolumn{3}{c|}{\textbf{Qwen}} & \multicolumn{3}{c}{\textbf{GPT}}\\
 & \multicolumn{1}{c}{w/o M.} & \multicolumn{1}{c}{w/ M.} & \multicolumn{1}{c|}{w/ M.R.} & \multicolumn{1}{c}{w/o M.} & \multicolumn{1}{c}{w/ M.} & \multicolumn{1}{c|}{w/ M.R.} & \multicolumn{1}{c}{w/o M.} & \multicolumn{1}{c}{w/ M.} & \multicolumn{1}{c|}{w/ M.R.} & \multicolumn{1}{c}{w/o M.} & \multicolumn{1}{c}{w/ M.} & \multicolumn{1}{c}{w/ M.R.} \\ 
 \midrule
Sephora& 4.09 & \underline{4.10} & \textbf{4.37} & 4.57 & \underline{4.58} &\textbf{4.66} & 4.61 & \underline{4.64} & \textbf{4.70} & 4.69 & \underline{4.69} & \textbf{4.77} \\
Dianping& 4.29 & \underline{4.34} & \textbf{4.41} & 4.13 & \underline{4.18} & \textbf{4.46} & \underline{4.14} & 4.13 & \textbf{4.43} & 4.32 & \underline{4.34} & \textbf{4.71} \\
WikiRevision& 4.19 & \underline{4.19} & \textbf{4.36} & 4.21 & \underline{4.31} & \textbf{4.48} & 4.66 & \underline{4.73} & \textbf{4.93} & 4.74 & \underline{4.79} & \textbf{4.96} \\
WikiLife& \underline{4.30} & 4.23 & \textbf{4.34} & \underline{4.31} & 4.25 & \textbf{4.55} & 4.31 & \underline{4.33} & \textbf{4.40} & 4.44 & \underline{4.46} & \textbf{4.59} \\
IMDB& 3.65 & \underline{3.82} & \textbf{3.99} & 3.97 & \underline{4.02} & \textbf{4.32} & 4.10 & \underline{4.18} & \textbf{4.33} & 3.91 & \underline{4.02} & \textbf{4.44} \\
WeiboTech& 3.88 & \underline{3.89} & \textbf{3.92} & 4.49 & \underline{4.56} & \textbf{4.86} & \underline{4.93} & 4.91 & \textbf{4.96} & 4.84 & \underline{4.88} & \textbf{4.97} \\
WeiboDaily& 3.95 & \underline{4.22} & \textbf{4.25} & 4.51 & \underline{4.58} & \textbf{4.90} & 4.88 & \underline{4.88} & \textbf{4.98} & 4.80 & \underline{4.92} & \textbf{4.99} \\
Cora& 4.31 & \underline{4.40} & \textbf{4.60} & 4.12 & \underline{4.12} & \textbf{4.43} & 4.21 & \underline{4.24} & \textbf{4.42} & 3.98 & \underline{4.10} & \textbf{4.52} \\
\bottomrule
\end{tabular}}
\end{table}

\begin{table}[H]
\centering 
\caption{The results on each scoring criterion of generated textual quality under TDGG on \textbf{Sephora}.} 
\label{tab:tdgg_textual_sephora}
\resizebox{1.0\linewidth}{!}{
\begin{tabular}{l|ccc|ccc|ccc|ccc}
\toprule
 & \multicolumn{3}{c|}{\textbf{DeepSeek}} & \multicolumn{3}{c|}{\textbf{Llama}} & \multicolumn{3}{c|}{\textbf{Qwen}} & \multicolumn{3}{c}{\textbf{GPT}}\\
 & \multicolumn{1}{c}{w/o M.} & \multicolumn{1}{c}{w/ M.} & \multicolumn{1}{c|}{w/ M.R.} & \multicolumn{1}{c}{w/o M.} & \multicolumn{1}{c}{w/ M.} & \multicolumn{1}{c|}{w/ M.R.} & \multicolumn{1}{c}{w/o M.} & \multicolumn{1}{c}{w/ M.} & \multicolumn{1}{c|}{w/ M.R.} & \multicolumn{1}{c}{w/o M.} & \multicolumn{1}{c}{w/ M.} & \multicolumn{1}{c}{w/ M.R.} \\ 
    \midrule
    Contextual Fidelity 
    & 4.23 & \underline{4.28} & \textbf{4.48} 
    & \underline{4.63} & \underline{4.63} & \textbf{4.71} 
    & 4.69 & \underline{4.72} & \textbf{4.78} 
    & 4.73 & \underline{4.74} & \textbf{4.81} \\
    Personality Depth 
    & 3.86 & \underline{3.89} & \textbf{4.21} 
    & 4.49 & \underline{4.52} & \textbf{4.65} 
    & 4.52 & \underline{4.57} & \textbf{4.67} 
    & 4.61 & \underline{4.62} & \textbf{4.74} \\
    Dynamic Adaptability 
    & \underline{4.07} & 4.02 & \textbf{4.36} 
    & 4.52 & \underline{4.53} & \textbf{4.58} 
    & 4.55 & \underline{4.58} & \textbf{4.62} 
    & \underline{4.65} & 4.64 & \textbf{4.71} \\
    Immersive Quality 
    & 4.11 & \underline{4.12} & \textbf{4.38} 
    & \underline{4.59} & \underline{4.59} & \textbf{4.69} 
    & 4.64 & \underline{4.67} & \textbf{4.74} 
    & 4.71 & \underline{4.72} & \textbf{4.79} \\
    Content Richness 
    & \underline{4.17} & \underline{4.17} & \textbf{4.42} 
    & 4.60 & \underline{4.61} & \textbf{4.67} 
    & 4.66 & \underline{4.67} & \textbf{4.70} 
    & \underline{4.73} & 4.72 & \textbf{4.77} \\
    \midrule
    Average 
    & 4.09 & \underline{4.10} & \textbf{4.37} 
    & 4.57 & \underline{4.58} &\textbf{4.66} 
    & 4.61 & \underline{4.64} & \textbf{4.70} 
    & \underline{4.69} & \underline{4.69} & \textbf{4.77} \\
    \bottomrule
\end{tabular}}
\end{table}

\begin{table}[H]
\centering \caption{The results on each scoring criterion of generated textual quality under TDGG on \textbf{Dianping}.} 
\label{tab:tdgg_textual_dianping}
\resizebox{1.0\linewidth}{!}{
\begin{tabular}{l|ccc|ccc|ccc|ccc}\toprule
 & \multicolumn{3}{c|}{\textbf{DeepSeek}} & \multicolumn{3}{c|}{\textbf{Llama}} & \multicolumn{3}{c|}{\textbf{Qwen}} & \multicolumn{3}{c}{\textbf{GPT}}\\
 & \multicolumn{1}{c}{w/o M.} & \multicolumn{1}{c}{w/ M.} & \multicolumn{1}{c|}{w/ M.R.} & \multicolumn{1}{c}{w/o M.} & \multicolumn{1}{c}{w/ M.} & \multicolumn{1}{c|}{w/ M.R.} & \multicolumn{1}{c}{w/o M.} & \multicolumn{1}{c}{w/ M.} & \multicolumn{1}{c|}{w/ M.R.} & \multicolumn{1}{c}{w/o M.} & \multicolumn{1}{c}{w/ M.} & \multicolumn{1}{c}{w/ M.R.} \\ 
    \midrule
    Contextual Fidelity 
    & 4.42 & \underline{4.44} & \textbf{4.54} 
    & 4.13 & \underline{4.21} & \textbf{4.48} 
    & \underline{4.16} & 4.13 & \textbf{4.46} 
    & \underline{4.33} & \underline{4.33} & \textbf{4.71} \\
    Personality Depth 
    & 4.13 & \underline{4.23} & \textbf{4.28} 
    & 4.02 & \underline{4.06} & \textbf{4.47} 
    & \underline{4.03} & 3.99 & \textbf{4.36} 
    & 4.19 & \underline{4.21} & \textbf{4.66} \\
    Dynamic Adaptability 
    & 4.24 & \underline{4.28} & \textbf{4.34} 
    & 4.09 & \underline{4.16} & \textbf{4.38} 
    & 4.06 & \underline{4.07} & \textbf{4.33} 
    & 4.25 & \underline{4.30} & \textbf{4.68} \\
    Immersive Quality 
    & 4.28 & \underline{4.35} & \textbf{4.42} 
    & 4.15 & \underline{4.19} & \textbf{4.47} 
    & \underline{4.16} & 4.15 & \textbf{4.45} 
    & \underline{4.34} & \underline{4.34} & \textbf{4.72} \\
    Content Richness 
    & 4.40 & \underline{4.43} & \textbf{4.50} 
    & 4.23 & \underline{4.30} & \textbf{4.51} 
    & \underline{4.30} & \underline{4.30} & \textbf{4.54} 
    & 4.50 & \underline{4.51} & \textbf{4.78} \\
    \midrule
    Average 
    & 4.29 & \underline{4.34} & \textbf{4.41} 
    & 4.13 & \underline{4.18} & \textbf{4.46} 
    & \underline{4.14} & 4.13 & \textbf{4.43} 
    & 4.32 & \underline{4.34} & \textbf{4.71} \\
\bottomrule
\end{tabular}}
\end{table}

\begin{table}[H]
\centering \caption{The results on each scoring criterion of generated textual quality under TDGG on \textbf{WikiRevision}.} 
\label{tab:tdgg_textual_wikirevision}
\resizebox{1.0\linewidth}{!}{
\begin{tabular}{l|ccc|ccc|ccc|ccc}\toprule
 & \multicolumn{3}{c|}{\textbf{DeepSeek}} & \multicolumn{3}{c|}{\textbf{Llama}} & \multicolumn{3}{c|}{\textbf{Qwen}} & \multicolumn{3}{c}{\textbf{GPT}}\\
 & \multicolumn{1}{c}{w/o M.} & \multicolumn{1}{c}{w/ M.} & \multicolumn{1}{c|}{w/ M.R.} & \multicolumn{1}{c}{w/o M.} & \multicolumn{1}{c}{w/ M.} & \multicolumn{1}{c|}{w/ M.R.} & \multicolumn{1}{c}{w/o M.} & \multicolumn{1}{c}{w/ M.} & \multicolumn{1}{c|}{w/ M.R.} & \multicolumn{1}{c}{w/o M.} & \multicolumn{1}{c}{w/ M.} & \multicolumn{1}{c}{w/ M.R.} \\ 
     \midrule
    Contextual Fidelity 
    & \underline{4.27} & 4.26 & \textbf{4.42} 
    & 4.64 & \underline{4.66} & \textbf{4.70} 
    & 4.74 & \underline{4.77} & \textbf{4.95} 
    & 4.82 & \underline{4.86} & \textbf{4.97} \\
    Personality Depth 
    & \underline{4.16} & \underline{4.16} & \textbf{4.32} 
    & 4.12 & \underline{4.21} & \textbf{4.33} 
    & 4.59 & \underline{4.62} & \textbf{4.92} 
    & 4.67 & \underline{4.72} & \textbf{4.95} \\
    Dynamic Adaptability 
    & \underline{4.12} & 4.11 & \textbf{4.32} 
    & 4.02 & \underline{4.10} & \textbf{4.50} 
    & 4.55 & \underline{4.60} & \textbf{4.89} 
    & 4.66 & \underline{4.72} & \textbf{4.94} \\
    Immersive Quality 
    & \underline{4.21} & \underline{4.21} & \textbf{4.38}
    & 3.98 & \underline{4.06} & \textbf{4.21} 
    & 4.69 & \underline{4.73} & \textbf{4.94} 
    & 4.77 & \underline{4.82} & \textbf{4.97} \\
    Content Richness 
    & \underline{4.21} & 4.20 & \textbf{4.38} 
    & 4.30 & \underline{4.51} & \textbf{4.66} 
    & 4.72 & \underline{4.73} & \textbf{4.94}
    & 4.79 & \underline{4.81} & \textbf{4.96} \\
    \midrule
    Average 
    & \underline{4.19} & \underline{4.19} & \textbf{4.36} 
    & 4.21 & \underline{4.31} & \textbf{4.48} 
    & 4.66 & \underline{4.73} & \textbf{4.93} 
    & 4.74 & \underline{4.79} & \textbf{4.96} \\
\bottomrule
\end{tabular}}
\end{table}

\begin{table}[H]
\centering \caption{The results on each scoring criterion of generated textual quality under TDGG on \textbf{WikiLife}.} 
\label{tab:tdgg_textual_wikilife}
\resizebox{1.0\linewidth}{!}{
\begin{tabular}{l|ccc|ccc|ccc|ccc}\toprule
 & \multicolumn{3}{c|}{\textbf{DeepSeek}} & \multicolumn{3}{c|}{\textbf{Llama}} & \multicolumn{3}{c|}{\textbf{Qwen}} & \multicolumn{3}{c}{\textbf{GPT}}\\
 & \multicolumn{1}{c}{w/o M.} & \multicolumn{1}{c}{w/ M.} & \multicolumn{1}{c|}{w/ M.R.} & \multicolumn{1}{c}{w/o M.} & \multicolumn{1}{c}{w/ M.} & \multicolumn{1}{c|}{w/ M.R.} & \multicolumn{1}{c}{w/o M.} & \multicolumn{1}{c}{w/ M.} & \multicolumn{1}{c|}{w/ M.R.} & \multicolumn{1}{c}{w/o M.} & \multicolumn{1}{c}{w/ M.} & \multicolumn{1}{c}{w/ M.R.} \\ 
     \midrule
    Contextual Fidelity 
    & \textbf{4.43} & \underline{4.34} & \textbf{4.43} 
    & \underline{4.47} & 4.44 & \textbf{4.65} 
    & 4.40 & \underline{4.44} & \textbf{4.48} 
    & 4.54 & \underline{4.58} & \textbf{4.66} \\
    Personality Depth 
    & \underline{4.29} & 4.19 & \textbf{4.31} 
    & \underline{4.31} & 4.27 & \textbf{4.59}
    & 4.29 & \underline{4.31} & \textbf{4.39} 
    & 4.41 & \underline{4.43} & \textbf{4.58} \\
    Dynamic Adaptability 
    & \underline{4.08} & 4.04 & \textbf{4.19} 
    & \underline{4.10} & \underline{4.10} & \textbf{4.35} 
    & \underline{4.14} & 4.13 & \textbf{4.24}
    & 4.25 & \underline{4.28} & \textbf{4.45} \\
    Immersive Quality 
    & \textbf{4.39} & 4.26 & \underline{4.38} 
    & \underline{4.42} & 4.40 & \textbf{4.64} 
    & 4.39 & \underline{4.41} & \textbf{4.47}
    & \underline{4.54} & \underline{4.54} & \textbf{4.65} \\
    Content Richness 
    & \underline{4.33} & 4.30 & \textbf{4.38} 
    & \underline{4.26} & 4.25 & \textbf{4.54} 
    & 4.31 & \underline{4.34} & \textbf{4.42} 
    & \underline{4.46} & \underline{4.46} & \textbf{4.61} \\
    \midrule
    Average 
    & \underline{4.30} & 4.23 & \textbf{4.34} 
    & \underline{4.31} & 4.25 & \textbf{4.55} 
    & 4.31 & \underline{4.33} & \textbf{4.40} 
    & 4.44 & \underline{4.46} & \textbf{4.59} \\
\bottomrule
\end{tabular}}
\end{table}

\begin{table}[H]
\centering \caption{The results on each scoring criterion of generated textual quality under TDGG on \textbf{IMDB}.} 
\label{tab:tdgg_textual_imdb}
\resizebox{1.0\linewidth}{!}{
\begin{tabular}{l|ccc|ccc|ccc|ccc}\toprule
 & \multicolumn{3}{c|}{\textbf{DeepSeek}} & \multicolumn{3}{c|}{\textbf{Llama}} & \multicolumn{3}{c|}{\textbf{Qwen}} & \multicolumn{3}{c}{\textbf{GPT}}\\
 & \multicolumn{1}{c}{w/o M.} & \multicolumn{1}{c}{w/ M.} & \multicolumn{1}{c|}{w/ M.R.} & \multicolumn{1}{c}{w/o M.} & \multicolumn{1}{c}{w/ M.} & \multicolumn{1}{c|}{w/ M.R.} & \multicolumn{1}{c}{w/o M.} & \multicolumn{1}{c}{w/ M.} & \multicolumn{1}{c|}{w/ M.R.} & \multicolumn{1}{c}{w/o M.} & \multicolumn{1}{c}{w/ M.} & \multicolumn{1}{c}{w/ M.R.} \\ 
     \midrule
    Contextual Fidelity 
    & 3.93 & \underline{4.15} & \textbf{4.30 }
    & 4.22 & \underline{4.31} & \textbf{4.50 }
    & 4.42 & \underline{4.48} & \textbf{4.55 }
    & 4.14 & \underline{4.28} & \textbf{4.61} \\
    Personality Depth 
    & 3.49 & \underline{3.62} & \textbf{3.80} 
    & 3.83 & \underline{3.86} & \textbf{4.38 }
    & 3.95 & \underline{4.03} & \textbf{4.30 }
    & 3.78 & \underline{3.88} & \textbf{4.48} \\
    Dynamic Adaptability 
    & 3.45 & \underline{3.63} & \textbf{3.78} 
    & 3.76 & \underline{3.80} & \textbf{4.04}
    & 3.88 & \underline{3.95} & \textbf{4.06} 
    & 3.69 & \underline{3.82} & \textbf{4.12} \\
    Immersive Quality 
    & 3.72 & \underline{3.88} & \textbf{4.04 }
    & 4.07 & \underline{4.13} & \textbf{4.47 }
    & 4.22 & \underline{4.29} & \textbf{4.46 }
    & 4.01 & \underline{4.15} & \textbf{4.58} \\
    Content Richness 
    & 3.64 & \underline{3.83} & \textbf{4.03} 
    & 3.97 & \underline{4.00} & \textbf{4.24} 
    & 4.06 & \underline{4.14} & \textbf{4.27} 
    & 3.92 & \underline{3.97} & \textbf{4.38} \\
    \midrule
    Average 
    & 3.65 & \underline{3.82} & \textbf{3.99} 
    & 3.97 & \underline{4.02} & \textbf{4.32} 
    & 4.10 & \underline{4.18} & \textbf{4.33} 
    & 3.91 & \underline{4.02} & \textbf{4.44} \\
\bottomrule
\end{tabular}}
\end{table}

\begin{table}[H]
\centering \caption{The results on each scoring criterion of generated textual quality under TDGG on \textbf{WeiboTech}.} 
\label{tab:tdgg_textual_weibotech}
\resizebox{1.0\linewidth}{!}{
\begin{tabular}{l|ccc|ccc|ccc|ccc}\toprule
 & \multicolumn{3}{c|}{\textbf{DeepSeek}} & \multicolumn{3}{c|}{\textbf{Llama}} & \multicolumn{3}{c|}{\textbf{Qwen}} & \multicolumn{3}{c}{\textbf{GPT}}\\
 & \multicolumn{1}{c}{w/o M.} & \multicolumn{1}{c}{w/ M.} & \multicolumn{1}{c|}{w/ M.R.} & \multicolumn{1}{c}{w/o M.} & \multicolumn{1}{c}{w/ M.} & \multicolumn{1}{c|}{w/ M.R.} & \multicolumn{1}{c}{w/o M.} & \multicolumn{1}{c}{w/ M.} & \multicolumn{1}{c|}{w/ M.R.} & \multicolumn{1}{c}{w/o M.} & \multicolumn{1}{c}{w/ M.} & \multicolumn{1}{c}{w/ M.R.} \\ 
     \midrule
    Contextual Fidelity 
    & \underline{4.09} & 4.08 & \textbf{4.12 }
    & 4.59 & \underline{4.66} & \textbf{4.88 }
    & \underline{4.94} & 4.92 & \textbf{4.97 }
    & 4.86 & \underline{4.90} & \textbf{4.99 }\\
    Personality Depth 
    & \underline{3.96} & 3.95 & \textbf{3.97 }
    & 4.50 & \underline{4.54} & \textbf{4.90} 
    & \underline{4.91} & 4.89 & \textbf{4.95} 
    & 4.83 & \underline{4.86} & \textbf{4.97}\\
    Dynamic Adaptability 
    & 3.62 & \underline{3.63} & \textbf{3.70} 
    & 4.40 & \underline{4.50} & \textbf{4.82} 
    & \underline{4.93} & 4.91 & \textbf{4.95} 
    & 4.83 & \underline{4.88} & \textbf{4.97} \\
    Immersive Quality 
    & \underline{3.95} & \underline{3.95} & \textbf{4.00 }
    & 4.55 & \underline{4.62} & \textbf{4.88 }
    & \underline{4.93} & 4.92 & \textbf{4.96 }
    & 4.85 & \underline{4.88} & \textbf{4.98} \\
    Content Richness 
    & 3.79 & \underline{3.83} & \textbf{3.83 }
    & 4.43 & \underline{4.50} & \textbf{4.82} 
    & \underline{4.93} & 4.91 & \textbf{4.96} 
    & 4.82 & \underline{4.88} & \textbf{4.96} \\
    \midrule
    Average 
    & 3.88 & \underline{3.89} & \textbf{3.92} 
    & 4.49 & \underline{4.56} & \textbf{4.86} 
    & \underline{4.93} & 4.91 & \textbf{4.96} 
    & 4.84 & \underline{4.88} & \textbf{4.97} \\
\bottomrule
\end{tabular}}
\end{table}

\begin{table}[H]
\centering \caption{The results on each scoring criterion of generated textual quality under TDGG on \textbf{WeiboDaily}.} 
\label{tab:tdgg_textual_weibodaily}
\resizebox{1.0\linewidth}{!}{
\begin{tabular}{l|ccc|ccc|ccc|ccc}\toprule
 & \multicolumn{3}{c|}{\textbf{DeepSeek}} & \multicolumn{3}{c|}{\textbf{Llama}} & \multicolumn{3}{c|}{\textbf{Qwen}} & \multicolumn{3}{c}{\textbf{GPT}}\\
 & \multicolumn{1}{c}{w/o M.} & \multicolumn{1}{c}{w/ M.} & \multicolumn{1}{c|}{w/ M.R.} & \multicolumn{1}{c}{w/o M.} & \multicolumn{1}{c}{w/ M.} & \multicolumn{1}{c|}{w/ M.R.} & \multicolumn{1}{c}{w/o M.} & \multicolumn{1}{c}{w/ M.} & \multicolumn{1}{c|}{w/ M.R.} & \multicolumn{1}{c}{w/o M.} & \multicolumn{1}{c}{w/ M.} & \multicolumn{1}{c}{w/ M.R.} \\ 
     \midrule
    Contextual Fidelity 
    & 4.16 & \underline{4.38} & \textbf{4.40}
    & 4.61 & \underline{4.66} & \textbf{4.92} 
    & \underline{4.90} & 4.89 & \textbf{4.98} 
    & 4.83 & \underline{4.93} & \textbf{5.00} \\
    Personality Depth 
    & 4.05 & \underline{4.27} & \textbf{4.33} 
    & 4.53 & \underline{4.58} & \textbf{4.93} 
    & \underline{4.87} & 4.86 & \textbf{4.98} 
    & 4.79 & \underline{4.90} & \textbf{4.99} \\
    Dynamic Adaptability 
    & 3.65 & \textbf{3.98} & \underline{3.97} 
    & 4.44 & \underline{4.51} & \textbf{4.87} 
    & \underline{4.88} & 4.87 & \textbf{4.97} 
    & 4.78 & \underline{4.91} & \textbf{4.97} \\
    Immersive Quality 
    & 4.02 & \underline{4.30} & \textbf{4.37} 
    & 4.58 & \underline{4.63} & \textbf{4.92} 
    & \underline{4.90} & 4.89 & \textbf{4.98} 
    & 4.83 & \underline{4.92} & \textbf{5.00} \\
    Content Richness 
    & \underline{3.90} & \textbf{4.18} & \textbf{4.18} 
    & 4.41 & \underline{4.50} & \textbf{4.86} 
    & \underline{4.88} & 4.87 & \textbf{4.97} 
    & 4.77 & \underline{4.91} & \textbf{4.97} \\
    \midrule
    Average 
    & 3.95 & \underline{4.22} & \textbf{4.25} 
    & 4.51 & \underline{4.58} & \textbf{4.90} 
    & \underline{4.88} & \underline{4.88} & \textbf{4.98} 
    & 4.80 & \underline{4.92} & \textbf{4.99} \\
\bottomrule
\end{tabular}}
\end{table}

\begin{table}[H]
\centering \caption{The results on each scoring criterion of generated textual quality under TDGG on \textbf{Cora}. } 
\label{tab:tdgg_textual_cora}
\resizebox{1.0\linewidth}{!}{
\begin{tabular}{l|ccc|ccc|ccc|ccc}\toprule
 & \multicolumn{3}{c|}{\textbf{DeepSeek}} & \multicolumn{3}{c|}{\textbf{Llama}} & \multicolumn{3}{c|}{\textbf{Qwen}} & \multicolumn{3}{c}{\textbf{GPT}}\\
 & \multicolumn{1}{c}{w/o M.} & \multicolumn{1}{c}{w/ M.} & \multicolumn{1}{c|}{w/ M.R.} & \multicolumn{1}{c}{w/o M.} & \multicolumn{1}{c}{w/ M.} & \multicolumn{1}{c|}{w/ M.R.} & \multicolumn{1}{c}{w/o M.} & \multicolumn{1}{c}{w/ M.} & \multicolumn{1}{c|}{w/ M.R.} & \multicolumn{1}{c}{w/o M.} & \multicolumn{1}{c}{w/ M.} & \multicolumn{1}{c}{w/ M.R.} \\ 
     \midrule
    Contextual Fidelity 
    & 4.40 & \underline{4.49} & \textbf{4.69} 
    & \underline{4.21} & 4.19 & \textbf{4.51} 
    & 4.28 & \underline{4.32} & \textbf{4.49} 
    & 4.08 & \underline{4.18} & \textbf{4.59} \\
    Personality Depth 
    & 4.25 & \underline{4.32} & \textbf{4.53}
    & \underline{4.15} & 4.13 & \textbf{4.45} 
    & 4.21 & \underline{4.25} & \textbf{4.43} 
    & 3.99 & \underline{4.11} & \textbf{4.52} \\
    Dynamic Adaptability
    & 4.23 & \underline{4.28} & \textbf{4.47} 
    & 3.88 & \underline{3.93} & \textbf{4.23} 
    & \underline{4.04} & 4.01 & \textbf{4.25} 
    & 3.73 & \underline{3.89} & \textbf{4.33} \\
    Immersive Quality 
    & 4.33 & \underline{4.42} & \textbf{4.64} 
    & \underline{4.20} & 4.18 & \textbf{4.49} 
    & 4.26 & \underline{4.32} & \textbf{4.48} 
    & 4.06 & \underline{4.17} & \textbf{4.58} \\
    Content Richness 
    & 4.35 & \underline{4.49} & \textbf{4.69}
    & \underline{4.18} & 4.17 & \textbf{4.49} 
    & 4.25 & \underline{4.30} & \textbf{4.46} 
    & 4.05 & \underline{4.16} & \textbf{4.59} \\
    \midrule
    Average 
    & 4.31 & \underline{4.40} & \textbf{4.60} 
    & \underline{4.12} & \underline{4.12} & \textbf{4.43} 
    & 4.21 & \underline{4.24} & \textbf{4.42} 
    & 3.98 & \underline{4.10} & \textbf{4.52} \\
\bottomrule
\end{tabular}}
\end{table}

\subsection{Results of TDGG on Graph Embedding Metrics}
Full results on the graph embedding metric under TDGG are shown in \cref{tab:tdgg_graph_embedding}.
Similar to the findings on textual quality, the use of node memory and reflection mechanisms significantly enhances generation quality, underscoring the critical role of integrating structural and textual information in DyTAG generation.

\begin{table}[H]
\centering \caption{The results on the graph embedding metric under TDGG.} 
\label{tab:tdgg_graph_embedding}
\resizebox{1.0\linewidth}{!}{
\begin{tabular}{l|ccc|ccc|ccc|ccc}\toprule
 & \multicolumn{3}{c|}{\textbf{DeepSeek}} & \multicolumn{3}{c|}{\textbf{Llama}} & \multicolumn{3}{c|}{\textbf{Qwen}} & \multicolumn{3}{c}{\textbf{GPT}}\\
 & \multicolumn{1}{c}{w/o M.} & \multicolumn{1}{c}{w/ M.} & \multicolumn{1}{c|}{w/ M.R.} & \multicolumn{1}{c}{w/o M.} & \multicolumn{1}{c}{w/ M.} & \multicolumn{1}{c|}{w/ M.R.} & \multicolumn{1}{c}{w/o M.} & \multicolumn{1}{c}{w/ M.} & \multicolumn{1}{c|}{w/ M.R.} & \multicolumn{1}{c}{w/o M.} & \multicolumn{1}{c}{w/ M.} & \multicolumn{1}{c}{w/ M.R.} \\ 
 \midrule
Sephora        & \underline{0.715} & \textbf{0.758} & 0.679 & \underline{0.675} & \textbf{0.681} & 0.648 & 0.579 & \underline{0.630} & \textbf{0.740} & 0.588 & \textbf{0.671} & \underline{0.654} \\
Dianping       & 0.637 & \underline{0.666} & \textbf{0.722} & 0.368 & \underline{0.386} & \textbf{0.408} & \underline{0.408} & 0.390 & \textbf{0.413} & 0.351 & \underline{0.368} & \textbf{0.369} \\
WikiRevision   & 0.609 & \textbf{0.641} & \underline{0.624} & 0.567 & \underline{0.582} & \textbf{0.636} & \underline{0.458} & 0.429 & \textbf{0.464} & \underline{0.438} & 0.422 & \textbf{0.454} \\
WikiLife       & \textbf{0.599} & 0.554 & \underline{0.588} & 0.429 & \underline{0.462} & \textbf{0.477} & 0.443 & \underline{0.478} & \textbf{0.490} & \underline{0.459} & 0.453 & \textbf{0.463} \\
IMDB           & \underline{0.534} & \textbf{0.586} & 0.532 & 0.396 & \underline{0.482} & \textbf{0.533} & 0.487 & \underline{0.493} & \textbf{0.519} & \underline{0.432} & 0.420 & \textbf{0.435} \\
WeiboTech      & 0.415 & \textbf{0.577} & \underline{0.501} & 0.335 & \underline{0.417} & \textbf{0.420} & 0.296 & \underline{0.353} & \textbf{0.364} & 0.327 & \underline{0.488} & \textbf{0.698} \\
WeiboDaily     & \underline{0.471} & 0.424 & \textbf{0.494} & 0.403 & \textbf{0.524} & \underline{0.494} & \textbf{0.437} & 0.377 & \underline{0.389} & 0.681 & \underline{0.689} & \textbf{0.694} \\
Cora           & \textbf{0.649} & 0.556 & \underline{0.581} & \textbf{0.689} & \underline{0.499} & 0.483 & \underline{0.485} & 0.450 & \textbf{0.536} & \textbf{0.511} & 0.446 & \underline{0.465} \\

\bottomrule
\end{tabular}}
\end{table}

\subsection{Results of TDGG on Direct Usability in Downstream Tasks}\label{TDGG-downstream}
To evaluate the direct usability of DyTAGs generated under the TDGG task, we assess their effectiveness as substitutes for original graphs in standard transductive downstream tasks (edge classification and link prediction), following the conventional experimental settings in dynamic graph learning \citep{DyGLib,DTGB,peng2025tidformer}. We employ four established dynamic graph neural network (DGNN) models—TGN~\citep{TGN}, CAWN~\citep{CAWN}, GraphMixer~\citep{Graphmixer}, and DyGFormer~\citep{DyGLib}—to perform edge classification and link prediction on both the generated (Gen.) and original (Ori.) graphs across four datasets: Sephora, WikiLife, Cora, and WeiboTech.

For {edge classification}, the results in \cref{tab:tdgg_downstream_ec} show consistently small performance gaps between the generated and original graphs. The absolute differences in Precision ($|\Delta P|$), Recall ($|\Delta R|$), and F1-score ($|\Delta F|$) are uniformly below 0.04 across all models and datasets, with most values under 0.02. This indicates that the generated graphs preserve sufficient structural and textual fidelity to support accurate classification. Similarly, in {link prediction} (transductive setting in dynamic graph learning), the Average Precision (AP) differences ($|\Delta AP|$) between using generated and original graphs are minimal, as summarized in \cref{tab:tdgg_downstream_lp}. The maximum observed $|\Delta AP|$ is 0.025, further confirming that the generated DyTAGs accurately capture the temporal evolution patterns of the original graphs.These results demonstrate that DyTAGs generated by our framework in the TDGG task are of high quality and can be directly deployed in downstream applications with negligible performance degradation.
\begin{table}[ht]
\centering
\caption{The result of edge classification performance on generated graphs from TDGG and original graphs.}
\label{tab:tdgg_downstream_ec}
\resizebox{0.9\textwidth}{!}{
\begin{tabular}{l|l|l|lll|lll}
\toprule
\textbf{Model} & \textbf{Dataset} & \textbf{Type} & \textbf{Precision} & \textbf{Recall} & \textbf{F1} & $|\Delta P|\downarrow$ & $|\Delta R|\downarrow$ & $|\Delta F|\downarrow$ \\
\midrule
\multirow{8}{*}{\textbf{TGN}} 
& \multirow{2}{*}{Sephora} & Gen. & 0.556 $\pm$ 0.009 & 0.490 $\pm$ 0.012 & 0.523 $\pm$ 0.004 & \multirow{2}{*}{0.022} & \multirow{2}{*}{0.004} & \multirow{2}{*}{0.011} \\
& & Ori. & 0.534 $\pm$ 0.009 & 0.494 $\pm$ 0.036 & 0.512 $\pm$ 0.026 & & & \\
\cmidrule{2-9}
& \multirow{2}{*}{WikiLife} & Gen. & 0.176 $\pm$ 0.024 & 0.299 $\pm$ 0.017 & 0.191 $\pm$ 0.033 & \multirow{2}{*}{0.033} & \multirow{2}{*}{0.017} & \multirow{2}{*}{0.030} \\
& & Ori. & 0.209 $\pm$ 0.016 & 0.282 $\pm$ 0.035 & 0.221 $\pm$ 0.010 & & & \\
\cmidrule{2-9}
& \multirow{2}{*}{Cora} & Gen. & 0.219 $\pm$ 0.016 & 0.241 $\pm$ 0.004 & 0.206 $\pm$ 0.027 & \multirow{2}{*}{0.014} & \multirow{2}{*}{0.009} & \multirow{2}{*}{0.016} \\
& & Ori. & 0.233 $\pm$ 0.021 & 0.232 $\pm$ 0.017 & 0.190 $\pm$ 0.033 & & & \\
\cmidrule{2-9}
& \multirow{2}{*}{WeiboTech} & Gen. & 0.781 $\pm$ 0.008 & 0.790 $\pm$ 0.019 & 0.782 $\pm$ 0.012 & \multirow{2}{*}{0.004} & \multirow{2}{*}{0.019} & \multirow{2}{*}{0.026} \\
& & Ori. & 0.777 $\pm$ 0.009 & 0.771 $\pm$ 0.024 & 0.756 $\pm$ 0.012 & & & \\
\midrule
\multirow{8}{*}{\textbf{CAWN}} 
& \multirow{2}{*}{Sephora} & Gen. & 0.526 $\pm$ 0.017 & 0.690 $\pm$ 0.016 & 0.563 $\pm$ 0.014 & \multirow{2}{*}{0.005} & \multirow{2}{*}{0.006} & \multirow{2}{*}{0.007} \\
& & Ori. & 0.531 $\pm$ 0.029 & 0.696 $\pm$ 0.003 & 0.556 $\pm$ 0.023 & & & \\
\cmidrule{2-9}
& \multirow{2}{*}{WikiLife} & Gen. & 0.190 $\pm$ 0.015 & 0.292 $\pm$ 0.016 & 0.206 $\pm$ 0.023 & \multirow{2}{*}{0.009} & \multirow{2}{*}{0.008} & \multirow{2}{*}{0.022} \\
& & Ori. & 0.199 $\pm$ 0.003 & 0.284 $\pm$ 0.033 & 0.228 $\pm$ 0.026 & & & \\
\cmidrule{2-9}
& \multirow{2}{*}{Cora} & Gen. & 0.212 $\pm$ 0.014 & 0.278 $\pm$ 0.038 & 0.188 $\pm$ 0.032 & \multirow{2}{*}{0.010} & \multirow{2}{*}{0.009} & \multirow{2}{*}{0.016} \\
& & Ori. & 0.222 $\pm$ 0.020 & 0.269 $\pm$ 0.031 & 0.204 $\pm$ 0.013 & & & \\
\cmidrule{2-9}
& \multirow{2}{*}{WeiboTech} & Gen. & 0.783 $\pm$ 0.018 & 0.787 $\pm$ 0.017 & 0.784 $\pm$ 0.019 & \multirow{2}{*}{0.013} & \multirow{2}{*}{0.002} & \multirow{2}{*}{0.017} \\
& & Ori. & 0.770 $\pm$ 0.020 & 0.789 $\pm$ 0.020 & 0.767 $\pm$ 0.031 & & & \\
\midrule
\multirow{8}{*}{\textbf{GraphMixer}} 
& \multirow{2}{*}{Sephora} & Gen. & 0.576 $\pm$ 0.024 & 0.690 $\pm$ 0.003 & 0.563 $\pm$ 0.018 & \multirow{2}{*}{0.023} & \multirow{2}{*}{0.005} & \multirow{2}{*}{0.012} \\
& & Ori. & 0.553 $\pm$ 0.020 & 0.685 $\pm$ 0.017 & 0.575 $\pm$ 0.033 & & & \\
\cmidrule{2-9}
& \multirow{2}{*}{WikiLife} & Gen. & 0.227 $\pm$ 0.015 & 0.315 $\pm$ 0.005 & 0.218 $\pm$ 0.021 & \multirow{2}{*}{0.014} & \multirow{2}{*}{0.014} & \multirow{2}{*}{0.009} \\
& & Ori. & 0.213 $\pm$ 0.015 & 0.301 $\pm$ 0.013 & 0.209 $\pm$ 0.008 & & & \\
\cmidrule{2-9}
& \multirow{2}{*}{Cora} & Gen. & 0.222 $\pm$ 0.014 & 0.268 $\pm$ 0.015 & 0.222 $\pm$ 0.021 & \multirow{2}{*}{0.013} & \multirow{2}{*}{0.008} & \multirow{2}{*}{0.002} \\
& & Ori. & 0.235 $\pm$ 0.009 & 0.260 $\pm$ 0.033 & 0.224 $\pm$ 0.012 & & & \\
\cmidrule{2-9}
& \multirow{2}{*}{WeiboTech} & Gen. & 0.776 $\pm$ 0.018 & 0.787 $\pm$ 0.036 & 0.775 $\pm$ 0.026 & \multirow{2}{*}{0.005} & \multirow{2}{*}{0.007} & \multirow{2}{*}{0.002} \\
& & Ori. & 0.771 $\pm$ 0.009 & 0.780 $\pm$ 0.007 & 0.773 $\pm$ 0.006 & & & \\
\midrule
\multirow{8}{*}{\textbf{DyGFormer}} 
& \multirow{2}{*}{Sephora} & Gen. & 0.476 $\pm$ 0.011 & 0.690 $\pm$ 0.004 & 0.563 $\pm$ 0.029 & \multirow{2}{*}{0.007} & \multirow{2}{*}{0.008} & \multirow{2}{*}{0.003} \\
& & Ori. & 0.469 $\pm$ 0.025 & 0.698 $\pm$ 0.007 & 0.560 $\pm$ 0.017 & & & \\
\cmidrule{2-9}
& \multirow{2}{*}{WikiLife} & Gen. & 0.242 $\pm$ 0.024 & 0.311 $\pm$ 0.014 & 0.194 $\pm$ 0.029 & \multirow{2}{*}{0.003} & \multirow{2}{*}{0.000} & \multirow{2}{*}{0.002} \\
& & Ori. & 0.239 $\pm$ 0.029 & 0.311 $\pm$ 0.020 & 0.196 $\pm$ 0.017 & & & \\
\cmidrule{2-9}
& \multirow{2}{*}{Cora} & Gen. & 0.209 $\pm$ 0.018 & 0.274 $\pm$ 0.039 & 0.206 $\pm$ 0.023 & \multirow{2}{*}{0.010} & \multirow{2}{*}{0.024} & \multirow{2}{*}{0.007} \\
& & Ori. & 0.219 $\pm$ 0.021 & 0.250 $\pm$ 0.024 & 0.213 $\pm$ 0.030 & & & \\
\cmidrule{2-9}
& \multirow{2}{*}{WeiboTech} & Gen. & 0.806 $\pm$ 0.010 & 0.803 $\pm$ 0.024 & 0.780 $\pm$ 0.014 & \multirow{2}{*}{0.020} & \multirow{2}{*}{0.013} & \multirow{2}{*}{0.013} \\
& & Ori. & 0.786 $\pm$ 0.007 & 0.790 $\pm$ 0.006 & 0.767 $\pm$ 0.006 & & & \\
\bottomrule
\end{tabular}
}
\end{table}

\begin{table}[ht]
\centering
\caption{The results of link prediction performance (transductive) on generated graphs from TDGG and original graphs.}
\label{tab:tdgg_downstream_lp}
\resizebox{0.55\textwidth}{!}{
\begin{tabular}{l|l|l|c|c}
\toprule
\textbf{Model} & \textbf{Dataset} & \textbf{Type} & \textbf{Average Precision} & $|\Delta AP|\downarrow$ \\
\midrule
\multirow{8}{*}{\textbf{TGN}} 
& \multirow{2}{*}{Sephora} & Gen. & 0.663 $\pm$ 0.025 & \multirow{2}{*}{0.010} \\
& & Ori. & 0.673 $\pm$ 0.016 & \\
\cmidrule{2-5}
& \multirow{2}{*}{WikiLife} & Gen. & 0.807 $\pm$ 0.017 & \multirow{2}{*}{0.017} \\
& & Ori. & 0.790 $\pm$ 0.023 & \\
\cmidrule{2-5}
& \multirow{2}{*}{Cora} & Gen. & 0.621 $\pm$ 0.022 & \multirow{2}{*}{0.025} \\
& & Ori. & 0.596 $\pm$ 0.012 & \\
\cmidrule{2-5}
& \multirow{2}{*}{WeiboTech} & Gen. & 0.859 $\pm$ 0.009 & \multirow{2}{*}{0.021} \\
& & Ori. & 0.838 $\pm$ 0.027 & \\
\midrule
\multirow{8}{*}{\textbf{CAWN}} 
& \multirow{2}{*}{Sephora} & Gen. & 0.731 $\pm$ 0.017 & \multirow{2}{*}{0.001} \\
& & Ori. & 0.730 $\pm$ 0.015 & \\
\cmidrule{2-5}
& \multirow{2}{*}{WikiLife} & Gen. & 0.831 $\pm$ 0.003 & \multirow{2}{*}{0.019} \\
& & Ori. & 0.850 $\pm$ 0.023 & \\
\cmidrule{2-5}
& \multirow{2}{*}{Cora} & Gen. & 0.616 $\pm$ 0.011 & \multirow{2}{*}{0.024} \\
& & Ori. & 0.592 $\pm$ 0.017 & \\
\cmidrule{2-5}
& \multirow{2}{*}{WeiboTech} & Gen. & 0.862 $\pm$ 0.020 & \multirow{2}{*}{0.013} \\
& & Ori. & 0.849 $\pm$ 0.030 & \\
\midrule
\multirow{8}{*}{\textbf{GraphMixer}} 
& \multirow{2}{*}{Sephora} & Gen. & 0.729 $\pm$ 0.019 & \multirow{2}{*}{0.018} \\
& & Ori. & 0.711 $\pm$ 0.010 & \\
\cmidrule{2-5}
& \multirow{2}{*}{WikiLife} & Gen. & 0.813 $\pm$ 0.002 & \multirow{2}{*}{0.018} \\
& & Ori. & 0.795 $\pm$ 0.030 & \\
\cmidrule{2-5}
& \multirow{2}{*}{Cora} & Gen. & 0.626 $\pm$ 0.030 & \multirow{2}{*}{0.002} \\
& & Ori. & 0.624 $\pm$ 0.021 & \\
\cmidrule{2-5}
& \multirow{2}{*}{WeiboTech} & Gen. & 0.868 $\pm$ 0.028 & \multirow{2}{*}{0.010} \\
& & Ori. & 0.858 $\pm$ 0.011 & \\
\midrule
\multirow{8}{*}{\textbf{DyGFormer}} 
& \multirow{2}{*}{Sephora} & Gen. & 0.631 $\pm$ 0.007 & \multirow{2}{*}{0.013} \\
& & Ori. & 0.644 $\pm$ 0.027 & \\
\cmidrule{2-5}
& \multirow{2}{*}{WikiLife} & Gen. & 0.800 $\pm$ 0.022 & \multirow{2}{*}{0.004} \\
& & Ori. & 0.796 $\pm$ 0.012 & \\
\cmidrule{2-5}
& \multirow{2}{*}{Cora} & Gen. & 0.824 $\pm$ 0.020 & \multirow{2}{*}{0.004} \\
& & Ori. & 0.828 $\pm$ 0.027 & \\
\cmidrule{2-5}
& \multirow{2}{*}{WeiboTech} & Gen. & 0.800 $\pm$ 0.020 & \multirow{2}{*}{0.001} \\
& & Ori. & 0.799 $\pm$ 0.009 & \\
\bottomrule
\end{tabular}
}
\end{table}

\subsection{Results of Node Retrieval}
Comprehensive results on the node retrieval task, including Hit@1 and Hit@10 metrics, are presented in \cref{tab:node_retrieval_exp}. 
Detailed results for all LLM backbones and DGNN baselines are provided in \cref{tab:node_retrieval_all_llm_@1}–\ref{tab:node_retrieval_all_dgnn}. 
The experimental findings reveal that DGNNs maintain a significant advantage in node retrieval performance. 
However, with reduced training data availability, the GAG-General model achieves superior results than DGNNs on the Sephora and IMDB datasets under a generative paradigm. 
This highlights the strong dependency of DGNNs on large-scale training data and their limited generalization capability in low-data scenarios.

\begin{table}[H]
\centering \caption{The results on Hit@1 and Hit@10 under the node retrieval task. Ours correspond to the best results of our proposed GAG-General among four LLM backbones. } 
\label{tab:node_retrieval_exp}
\resizebox{0.95\linewidth}{!}{
\begin{tabular}{l|ccccc|ccccc}\toprule
 & \multicolumn{5}{c|}{\textbf{Hit@1}} & \multicolumn{5}{c}{\textbf{Hit@10}} \\
 & \multicolumn{1}{c}{Ours} & \multicolumn{1}{c}{TGN} & \multicolumn{1}{c}{CAWN} & \multicolumn{1}{c}{GraphMixer} & \multicolumn{1}{c|}{DyGFormer} & \multicolumn{1}{c}{Ours} & \multicolumn{1}{c}{TGN} & \multicolumn{1}{c}{CAWN} & \multicolumn{1}{c}{GraphMixer} & \multicolumn{1}{c}{DyGFormer} \\ 
 \midrule
Sephora&      \textbf{0.285} & 0.045 & 0.023 & \underline{0.065} & 0.028 & \textbf{0.386} & 0.259 & 0.206 & \underline{0.309} &0.184 \\
Dianping&     \underline{0.176} & 0.040 & 0.033 & 0.036 & \textbf{0.534} & {0.178} & 0.188 & 0.193 & \underline{0.207} & \textbf{0.585} \\
WikiRevision& \underline{0.105} & 0.031 & 0.067 & 0.099 & \textbf{0.439} & {0.230} & 0.164 & \underline{0.287} & 0.272 & \textbf{0.507} \\
WikiLife&     {0.139} & 0.119 & 0.116 & \textbf{0.175} & \underline{0.160} & {0.315} & \underline{0.388} & 0.379 & \textbf{0.511} & 0.264 \\
IMDB&         \textbf{0.142} & {0.015} & \underline{0.009} & 0.005 & 0.008 & \textbf{0.159} & \underline{0.123} & 0.110 & 0.105 & 0.111 \\
WeiboTech&    {0.051} & \underline{0.183} & 0.128 & 0.169 & \textbf{ 0.672} & {0.056} & 0.371 & \underline{0.448} & 0.378 & \textbf{0.678} \\
WeiboDaily&   {0.091} & 0.138 & \underline{0.267} & \textbf{0.445} & 0.154 & {0.100} & 0.295 & \underline{0.615} & \textbf{0.663} & 0.438 \\
Cora&         \underline{}{0.122} & 0.025 & 0.016 & 0.064 & \textbf{0.465} & {0.237} & 0.193 & 0.175 & \underline{0.299} & \textbf{0.321} \\
\bottomrule
\end{tabular}}
\end{table}

\begin{table}[H]
\centering \caption{The results on Hit@1 under the node retrieval task with our framework.} 
\label{tab:node_retrieval_all_llm_@1}
\resizebox{1.0\linewidth}{!}{
\begin{tabular}{l|ccc|ccc|ccc|ccc}\toprule
 & \multicolumn{3}{c|}{\textbf{DeepSeek}} & \multicolumn{3}{c|}{\textbf{Llama}} & \multicolumn{3}{c|}{\textbf{Qwen}} & \multicolumn{3}{c}{\textbf{GPT}}\\
 & \multicolumn{1}{c}{w/o M.} & \multicolumn{1}{c}{w/ M.} & \multicolumn{1}{c|}{w/ M.R.} & \multicolumn{1}{c}{w/o M.} & \multicolumn{1}{c}{w/ M.} & \multicolumn{1}{c|}{w/ M.R.} & \multicolumn{1}{c}{w/o M.} & \multicolumn{1}{c}{w/ M.} & \multicolumn{1}{c|}{w/ M.R.} & \multicolumn{1}{c}{w/o M.} & \multicolumn{1}{c}{w/ M.} & \multicolumn{1}{c}{w/ M.R.} \\ 
     \midrule
    Sephora        
    & 0.002 & \textbf{0.285} & \underline{0.048} 
    & 0.000 & \underline{0.039} & \textbf{0.073} 
    & 0.019 & \underline{0.046} & \textbf{0.064} 
    & 0.016 & \underline{0.066} & \textbf{0.077} \\
    Dianping       
    & 0.000 & \textbf{0.176} & \underline{0.094} 
    & 0.001 & \underline{0.002} & \textbf{0.003} 
    & 0.001 & \underline{0.008} & \textbf{0.041} 
    & 0.001 & \underline{0.002} & \textbf{0.013} \\
    WikiRevision   
    & 0.003 & \textbf{0.013} & \underline{0.005} 
    & 0.084 & \underline{0.092} & \textbf{0.105}
    & \textbf{0.013} & \underline{0.012} & 0.009 
    & \underline{0.013} & \underline{0.013} & \textbf{0.015} \\
    WikiLife       
    & 0.029 & \textbf{0.135} & \underline{0.034} 
    & 0.086 & \underline{0.094} & \textbf{0.097} 
    & 0.067 & \underline{0.073} & \textbf{0.075} 
    & 0.107 & \underline{0.126} & \textbf{0.139} \\
    IMDB           
    & 0.000 & \textbf{0.142} & \underline{0.020} 
    & 0.001 & \textbf{0.022} & \underline{0.005} 
    & 0.043 & \underline{0.043} & \textbf{0.053} 
    & 0.001 & \underline{0.017} & \textbf{0.019} \\
    WeiboTech      
    & 0.002 & \textbf{0.051} & \underline{0.021} 
    & \underline{0.001} & \textbf{0.002} & \underline{0.001} 
    & \underline{0.004} & \underline{0.004} & \textbf{0.005} 
    & 0.000 & \underline{0.007} & \textbf{0.008} \\
    WeiboDaily     
    & 0.009 & \textbf{0.091} & \underline{0.040} 
    & \underline{0.003} & \textbf{0.006} & \underline{0.003} 
    & 0.009 & \underline{0.012} & \textbf{0.013} 
    & 0.002 & \textbf{0.014} & \underline{0.004} \\
    Cora           
    & 0.002 & \underline{0.069} & \textbf{0.082} 
    & 0.000 & \underline{0.024} & \textbf{0.033} 
    & \underline{0.110} & 0.108 & \textbf{0.122} 
    & 0.000 & \underline{0.045} & \textbf{0.115} \\

\bottomrule
\end{tabular}}
\end{table}

\begin{table}[H]
\centering \caption{The results on Hit@10 under the node retrieval task with our framework.} 
\label{tab:node_retrieval_all_llm_@10}
\resizebox{1.0\linewidth}{!}{
\begin{tabular}{l|ccc|ccc|ccc|ccc}\toprule
 & \multicolumn{3}{c|}{\textbf{DeepSeek}} & \multicolumn{3}{c|}{\textbf{Llama}} & \multicolumn{3}{c|}{\textbf{Qwen}} & \multicolumn{3}{c}{\textbf{GPT}}\\
 & \multicolumn{1}{c}{w/o M.} & \multicolumn{1}{c}{w/ M.} & \multicolumn{1}{c|}{w/ M.R.} & \multicolumn{1}{c}{w/o M.} & \multicolumn{1}{c}{w/ M.} & \multicolumn{1}{c|}{w/ M.R.} & \multicolumn{1}{c}{w/o M.} & \multicolumn{1}{c}{w/ M.} & \multicolumn{1}{c|}{w/ M.R.} & \multicolumn{1}{c}{w/o M.} & \multicolumn{1}{c}{w/ M.} & \multicolumn{1}{c}{w/ M.R.} \\ 
     \midrule
    Sephora        
    & 0.064 & \textbf{0.386} & \underline{0.169} 
    & 0.090 & \underline{0.164} & \textbf{0.314} 
    & 0.085 & \underline{0.191} & \textbf{0.278} 
    & 0.071 & \underline{0.291} & \textbf{0.369} \\
    Dianping       
    & 0.004 & \textbf{0.178} & \underline{0.108}
    & 0.003 & \underline{0.004} & \textbf{0.014} 
    & 0.002 & \underline{0.015} & \textbf{0.061} 
    & 0.004 & \underline{0.009} & \textbf{0.024} \\
    WikiRevision   
    & \underline{0.031} & \textbf{0.041} & \underline{0.031} 
    & 0.143 & \underline{0.177} & \textbf{0.230} 
    & \textbf{0.067} & \underline{0.061} & 0.047 
    & \underline{0.074} & 0.065 & \textbf{0.077} \\
    WikiLife       
    & \underline{0.096} & \textbf{0.205} & 0.093 
    & 0.216 & \underline{0.255} & \textbf{0.260} 
    & 0.145 & \underline{0.156} & \textbf{0.178} 
    & 0.238 & \underline{0.275} & \textbf{0.315} \\
    IMDB           
    & 0.035 & \textbf{0.159} & \underline{0.063} 
    & 0.015 & \textbf{0.033} & \underline{0.021} 
    & \underline{0.055} & 0.054 & \textbf{0.064} 
    & 0.012 & \underline{0.025} & \textbf{0.029} \\
    WeiboTech      
    & 0.006 & \textbf{0.056} & \underline{0.028}
    & \textbf{0.006} & \underline{0.005} & \underline{0.005}
    & 0.007 & \underline{0.008} & \textbf{0.008} 
    & 0.005 & \underline{0.011} & \textbf{0.026} \\
    WeiboDaily     
    & 0.021 & \textbf{0.100} & \underline{0.049} 
    & \underline{0.015} & \textbf{0.024} & 0.012 
    & \textbf{0.022} & \underline{0.021} & 0.019 
    & \textbf{0.036} & \underline{0.022} & 0.016 \\
    Cora           
    & 0.064 & \underline{0.128} & \textbf{0.160} 
    & 0.111 & \underline{0.135} & \textbf{0.174} 
    & 0.205 & \underline{0.209} & \textbf{0.226} 
    & 0.128 & \underline{0.172} & \textbf{0.237} \\

\bottomrule
\end{tabular}}
\end{table}

\begin{table}[H]
\centering \caption{The results on Hit@1 and Hit@10 under the node retrieval task with DGNN baselines.} 
\label{tab:node_retrieval_all_dgnn}
\resizebox{1.0\linewidth}{!}{
\begin{tabular}{l|ccccc|ccccc}\toprule
 & \multicolumn{5}{c|}{\textbf{Hit@1}} & \multicolumn{5}{c}{\textbf{Hit@10}} \\
 & \multicolumn{1}{c}{JODIE} & \multicolumn{1}{c}{TGN} & \multicolumn{1}{c}{CAWN} & \multicolumn{1}{c}{GraphMixer} & \multicolumn{1}{c|}{DyGFormer} & \multicolumn{1}{c}{JODIE} & \multicolumn{1}{c}{TGN} & \multicolumn{1}{c}{CAWN} & \multicolumn{1}{c}{GraphMixer} & \multicolumn{1}{c}{DyGFormer} \\ 
 \midrule
Sephora       
& 0.018 & \underline{0.045} & 0.023 & \textbf{0.065} & 0.028 
& 0.120 & \underline{0.259} & 0.206 & \textbf{0.309} & 0.184 \\
Dianping      
& \underline{0.419} & 0.040 & 0.033 & 0.036 & \textbf{0.534} 
& \underline{0.427} & 0.188 & 0.193 & 0.207 & \textbf{0.585} \\
WikiRevision  
& \underline{0.284} & 0.031 & 0.067 & 0.099 & \textbf{0.439} 
& \underline{0.375} & 0.164 & 0.287 & 0.272 & \textbf{0.507} \\
WikiLife      
& 0.049 & 0.119 & 0.116 & \textbf{0.175} & \underline{0.160} 
& \underline{0.410} & 0.388 & 0.379 & \textbf{0.511} & 0.264 \\
IMDB          
& \underline{0.013} & \textbf{0.015} & 0.009 & 0.005 & 0.008 
& \underline{0.111} & \textbf{0.123} & 0.110 & 0.105 & \underline{0.111} \\
WeiboTech     
& \underline{0.632} & 0.183 & 0.128 & 0.169 &\textbf{0.672} 
&\underline{0.650} & 0.371 & 0.448 & 0.378 & \textbf{0.678} \\
WeiboDaily    
& 0.027 & 0.138 & \underline{0.267} & \textbf{0.445} & 0.154 
& 0.440 & 0.295 & \underline{0.615} & \textbf{0.663} & 0.438 \\
Cora          
& 0.017 & 0.025 & 0.016 & \underline{0.064} & \textbf{0.465} 
& 0.129 & 0.193 & 0.175 & \underline{0.299} & \textbf{0.321} \\

\bottomrule
\end{tabular}}
\end{table}

\subsection{Results of Edge Classification}
Full results on Precision, Recall, and F1 metric under the edge classification task are available in \cref{tab:edge_classification_exp}. 
Detailed results on all LLM backbones and DGNN baselines are available in \cref{tab:edge_classification_llm} and \ref{tab:edge_classification_dgnn}.
The experimental findings demonstrate that our GAG-General model achieves excellent performance on this discriminative task under the generative paradigm. 
This success stems from the strong correlation between edge labels and the semantic information encoded in both node texts and generated edge texts. 
For instance, in e-commerce recommendation scenarios, if a negative review is generated as the edge text, the corresponding edge label (e.g., low rating) aligns closely with this sentiment. 
In contrast, existing DGNNs struggle to effectively capture and utilize the semantic richness of node/edge texts in DyTAG datasets, resulting in significant performance gaps on edge classification tasks.

\begin{table}[H]
\caption{The results on Precision, Recall, and F1 metric under the edge classification task. Ours correspond to the best results of our proposed GAG-General among four LLM backbones.} \centering
\label{tab:edge_classification_exp}
\resizebox{0.70\textwidth}{!}{
\begin{tabular}{c|c|cccccc}
\toprule
\textbf{Datasets}
&\textbf{Models}
&\multicolumn{1}{c}{\textbf{Ours}}
 &\multicolumn{1}{c}{\textbf{JODIE}}
&\multicolumn{1}{c}{\textbf{TGN}}
&\multicolumn{1}{c}{\textbf{CAWN}}
&\multicolumn{1}{c}{\textbf{GraphMixer}}
&\multicolumn{1}{c}{\textbf{DyGFormer}}\\
\midrule
&Precision &\textbf{0.798} &0.463 &0.452 &0.442 &0.443 &\underline{0.471} \\
 Sephora        &Recall &\textbf{0.790} &0.661 &0.672 &0.661 &0.684 &\underline{0.687} \\
 &F1 &\textbf{0.790} &0.555 &0.529 &0.532 &0.546 &\underline{0.559} \\

\midrule

 &Precision &\textbf{0.572} &0.359 &0.345 &\underline{0.409} &0.325 &0.217 \\
 Dianping       &Recall &\textbf{0.533} &0.437 &0.436 &0.459 &0.466 &\underline{0.466} \\
 &F1 &\textbf{0.522} &\underline{0.341} &0.279 &0.330 &0.298 &0.296 \\

\midrule

 &Precision &\textbf{0.804} &0.291 &0.474 &0.739 &\underline{0.756} &0.401 \\
 WikiRevision   &Recall &\textbf{0.829} &0.378 &0.387 &\underline{0.444} &0.382 &0.431 \\
 &F1 &\textbf{0.782} &0.209 &0.232 &\underline{0.341} &0.217 &0.303 \\

\midrule

 &Precision &\textbf{0.562} &0.151 &0.171 &\underline{0.199} &0.194 &0.093 \\
 WikiLife       &Recall &\textbf{0.451} &0.255 &0.288 &0.290 &\underline{0.291} &0.284 \\
 &F1 &\textbf{0.484} &0.163 &0.188 &\underline{0.209} &0.172 &0.128 \\

 \midrule

 &Precision &\textbf{0.698} &0.423 &0.442 &0.470 &\underline{0.473} &0.463 \\
 IMDB           &Recall &\underline{0.666} &0.680 &0.672 &0.677 &0.650 &\textbf{0.682} \\
 &F1 &\textbf{0.672} &0.531 &0.542 &\underline{0.551} &0.546 &0.533 \\

 \midrule

 &Precision &\textbf{0.726} &\underline{0.615} &0.479 &0.479 &0.503 &0.533 \\
WeiboTech      &Recall &\textbf{0.779} &0.687 &0.692 &0.690 &0.692 &\underline{0.692} \\
 &F1 &\textbf{0.740}&0.587 &0.566 &0.566 &0.566 &\underline{0.568} \\

 \midrule

 &Precision &\textbf{0.980}&0.853 &0.853 &0.867 &0.853 &\underline{0.898} \\
WeiboDaily     &Recall &\textbf{0.990}&0.924 &0.924 &0.922 &0.924 &\underline{0.925} \\
 &F1 &\textbf{0.985} &0.887 &0.887 &\underline{0.891} &0.887 &0.890 \\

\midrule

 &Precision &\textbf{0.566} &\underline{0.186} &0.149 &0.149 &0.153 &0.099 \\
Cora           &Recall &\textbf{0.550}&0.273 &0.280 &0.279 &0.279 &\underline{0.280} \\
 &F1 &\textbf{0.539} &\underline{0.152} &0.148 &0.134 &0.131 &0.126 \\

\bottomrule
\end{tabular}}
\end{table}

\begin{table}[H]
\caption{The results on Precision, Recall, and F1 metric under the edge classification task on all LLM backbones. } \centering
\label{tab:edge_classification_llm}
\resizebox{1.0\textwidth}{!}{
\begin{tabular}{l|c|ccc|ccc|ccc|ccc}\toprule
 & & \multicolumn{3}{c|}{\textbf{DeepSeek}} & \multicolumn{3}{c|}{\textbf{Llama}} & \multicolumn{3}{c|}{\textbf{Qwen}} & \multicolumn{3}{c}{\textbf{GPT}}\\
 & & \multicolumn{1}{c}{w/o M.} & \multicolumn{1}{c}{w/ M.} & \multicolumn{1}{c|}{w/ M.R.} & \multicolumn{1}{c}{w/o M.} & \multicolumn{1}{c}{w/ M.} & \multicolumn{1}{c|}{w/ M.R.} & \multicolumn{1}{c}{w/o M.} & \multicolumn{1}{c}{w/ M.} & \multicolumn{1}{c|}{w/ M.R.} & \multicolumn{1}{c}{w/o M.} & \multicolumn{1}{c}{w/ M.} & \multicolumn{1}{c}{w/ M.R.} \\ 

\midrule
&Precision 
&0.714 &\underline{0.744} &\textbf{0.798} 
&0.631 &\textbf{0.638} &\underline{0.634} 
&\underline{0.550} &\textbf{0.571} &0.546 
&\underline{0.542} &\textbf{0.572} &\textbf{0.572} \\
Sephora
&Recall 
&0.682 &\underline{0.742} &\textbf{0.790} 
&\underline{0.628} &\underline{0.628} &\textbf{0.637} 
&0.386 &\underline{0.397} &\textbf{0.400}
&\textbf{0.593} &0.586 &\underline{0.591} \\
&F1 
&0.686 &\underline{0.734} &\textbf{0.790} 
&\underline{0.620} &\underline{0.620} &\textbf{0.625} 
&0.405 &\underline{0.416} &\textbf{0.420}
&\underline{0.566} &0.564 &\textbf{0.570}\\
\midrule

&Precision 
&\underline{0.541} &\textbf{0.572} &0.531 
&0.427 &\textbf{0.462} &\underline{0.455} 
&0.410 &\underline{0.422} &\textbf{0.425} 
&\textbf{0.409} &\underline{0.396} &0.393 \\
Dianping
&Recall 
&\underline{0.529} &0.330 &\textbf{0.533} 
&0.346 &\underline{0.352} &\textbf{0.380}
&0.408 &\textbf{0.432} &\underline{0.422} 
&\textbf{0.374} &0.343 &\underline{0.354} \\
&F1 
&\underline{0.521} &0.411 &\textbf{0.522} 
&0.260 &\underline{0.271} &\textbf{0.291} 
&0.384 &\textbf{0.411} &\underline{0.397} 
&\textbf{0.311} &0.286 &\underline{0.288} \\
\midrule

&Precision 
&\textbf{0.750} &0.715 &\underline{0.736} 
&0.712 &\underline{0.717} &\textbf{0.804} 
&0.587 &\underline{0.605} &\textbf{0.625} 
&\underline{0.681} &0.678 &\textbf{0.689} \\
WikiRevision
&Recall 
&\textbf{0.741} &0.709 &\underline{0.727} 
&\underline{0.739} &0.731 &\textbf{0.829} 
&0.489 &\underline{0.496} &\textbf{0.505} 
&\underline{0.497} &0.493 &\textbf{0.503} \\
&F1 
&\textbf{0.727} &0.681 &\underline{0.712} 
&\underline{0.695} &0.686 &\textbf{0.782} 
&0.429 &\underline{0.437} &\textbf{0.444} 
&\underline{0.410} &0.404 &\textbf{0.420}\\
\midrule

&Precision 
&0.411 &\underline{0.414} &\textbf{0.438} 
&\underline{0.537} &\textbf{0.561} &\underline{0.537} 
&0.393 &\textbf{0.416} &\underline{0.414}
&0.544 &\underline{0.549} &\textbf{0.562} \\
WikiLife
&Recall 
&\underline{0.387} &0.386 &\textbf{0.415} 
&0.240 &\textbf{0.254} &\underline{0.243} 
&0.246 &\underline{0.249} &\textbf{0.253} 
&\underline{0.442} &0.441 &\textbf{0.451} \\
&F1 
&\underline{0.381} &0.380 &\textbf{0.409} 
&0.292 &\textbf{0.306} &\underline{0.297} 
&0.210 &\textbf{0.219} &\underline{0.215} 
&0.472 &\underline{0.475} &\textbf{0.484} \\
\midrule

&Precision 
&0.640 &\underline{0.666} &\textbf{0.698} 
&0.634 &\textbf{0.686} &\underline{0.647} 
&\textbf{0.683} &\underline{0.681} &0.647 
&0.550 &\underline{0.617} &\textbf{0.606} \\
IMDB
&Recall 
&0.573 &\underline{0.615} &\textbf{0.657} 
&0.431 &\underline{0.529} &\textbf{0.566} 
&\textbf{0.666} &\underline{0.664} &0.665 
&0.427 &\underline{0.526} &\textbf{0.576} \\
&F1 
&0.598 &\underline{0.633} &\textbf{0.672} 
&0.497 &\underline{0.572} &\textbf{0.595} 
&\textbf{0.626} &0.620 &\underline{0.621} 
&0.466 &\underline{0.547} &\textbf{0.577} \\
\midrule

&Precision 
&\underline{0.676} &0.675 &\textbf{0.688} 
&0.629 &\textbf{0.701} &\underline{0.631} 
&\underline{0.644} &\textbf{0.693} &0.642 
&0.626 &\underline{0.636} &\textbf{0.726} \\
WeiboTech
&Recall 
&\textbf{0.735} &0.728 &\underline{0.733} 
&\underline{0.697} &\textbf{0.745} &0.694 
&\textbf{0.704} &\textbf{0.704} &\underline{0.703} 
&0.703 &\underline{0.704} &\textbf{0.779} \\
&F1 
&\underline{0.685} &0.679 &\textbf{0.693} 
&\underline{0.589} &\textbf{0.646} &0.574 
&\underline{0.583} &\underline{0.583} &\textbf{0.595} 
&\underline{0.590} &0.584 &\textbf{0.740}\\
\midrule

&Precision 
&\underline{0.918} &0.913 &\textbf{0.923} 
&0.902 &\underline{0.925} &\textbf{0.927} 
&0.888 &\textbf{0.935} &\underline{0.906}
&\underline{0.980} &0.902 &\textbf{0.980}\\
WeiboDaily
&Recall 
&\underline{0.918} &0.916 &\textbf{0.931} 
&\underline{0.924} &\textbf{0.948} &0.921 
&0.930 &\underline{0.931} &\textbf{0.931} 
&\underline{0.990} &0.931 &\textbf{0.990}\\
&F1 
&\underline{0.918} &0.914 &\textbf{0.927} 
&\underline{0.891} &\textbf{0.923} &0.883 
&0.896 &\underline{0.897} &\textbf{0.898} 
&\underline{0.985} &0.899 &\textbf{0.985} \\
\midrule

&Precision 
&\textbf{0.566} &0.493 &\underline{0.542} 
&\textbf{0.528} &\underline{0.495} &0.460
&0.427 &\underline{0.445} &\textbf{0.456} 
&\textbf{0.480} &0.377 &\underline{0.410}\\
Cora
&Recall 
&\textbf{0.550} &0.474 &\underline{0.520}
&\textbf{0.473} &\underline{0.431} &0.414 
&0.414 &\underline{0.427} &\textbf{0.436} 
&\textbf{0.421} &0.377 &\underline{0.386} \\
&F1 
&\textbf{0.539} &0.454 &\underline{0.508} 
&\textbf{0.436} &\underline{0.385} &0.367 
&0.382 &\underline{0.398} &\textbf{0.403} 
&\textbf{0.370} &\underline{0.335} &0.334 \\

\bottomrule
\end{tabular}}
\end{table}

\begin{table}[H]
\caption{The results on Precision, Recall, and F1 metric under the edge classification task on all DGNN baselines.} \centering
\label{tab:edge_classification_dgnn}
\resizebox{0.65\textwidth}{!}{
\begin{tabular}{l|c|ccccc}
\toprule
\textbf{Datasets}
&\textbf{Models}

 &\multicolumn{1}{c}{\textbf{JODIE}}
&\multicolumn{1}{c}{\textbf{TGN}}
&\multicolumn{1}{c}{\textbf{CAWN}}
&\multicolumn{1}{c}{\textbf{GraphMixer}}
&\multicolumn{1}{c}{\textbf{DyGFormer}}\\
\midrule
&Precision & \underline{0.463} & 0.452 & 0.442 & 0.443 & \textbf{0.471} \\
Sephora        
&Recall & 0.661 & 0.672 & 0.661 & \underline{0.684} & \textbf{0.687} \\
 &F1 & \underline{0.555} & 0.529 & 0.532 & 0.546 & \textbf{0.559} \\
\midrule

&Precision & \underline{0.359} & 0.345 & \textbf{0.409} & 0.325 & 0.217 \\
Dianping      
&Recall & 0.437 & 0.436 & 0.459 & \underline{0.466} & \textbf{0.466} \\
 &F1 & \textbf{0.341} & 0.279 & \underline{0.330} & 0.298 & 0.296 \\
\midrule

&Precision & 0.291 & 0.474 & \underline{0.739} & \textbf{0.756} & 0.401 \\
WikiRevision   
&Recall & 0.378 & 0.387 & \textbf{0.444} & 0.382 & \underline{0.431} \\
 &F1 & 0.209 & 0.232 & \textbf{0.341} & 0.217 & \underline{0.303} \\
\midrule

&Precision & 0.151 & 0.171 & \textbf{0.199} & \underline{0.194} & 0.093 \\
WikiLife       
&Recall & 0.255 & 0.288 & \underline{0.290} & \textbf{0.291} & 0.284 \\
 &F1 & 0.163 & \underline{0.188} & \textbf{0.209} & 0.172 & 0.128 \\
\midrule

&Precision & 0.423 & 0.442 & \underline{0.470} & \textbf{0.473} & 0.463 \\
IMDB           
&Recall & \underline{0.680} & 0.672 & 0.677 & 0.650 & \textbf{0.682} \\
 &F1 & 0.531 & 0.542 & \textbf{0.551} & \underline{0.546} & 0.533 \\
\midrule

&Precision & \textbf{0.615} & 0.479 & 0.479 & 0.503 & \underline{0.533} \\
WeiboTech      
&Recall & 0.687 & \underline{0.692} & 0.690 & \underline{0.692} & \textbf{0.692} \\
 &F1 & \underline{0.587} & 0.566 & 0.566 & 0.566 & \textbf{0.568} \\
\midrule

&Precision & 0.853 & 0.853 & \underline{0.867} & 0.853 & \textbf{0.898} \\
WeiboDaily     
&Recall & \underline{0.924} & \underline{0.924} & 0.922 & \underline{0.924} & \textbf{0.925} \\
 &F1 & 0.887 & 0.887 & \textbf{0.891} & 0.887 & \underline{0.890} \\
\midrule

&Precision & \textbf{0.186} & 0.149 & 0.149 & \underline{0.153} & 0.099 \\
Cora           
&Recall & \underline{0.273} & \textbf{0.280} & 0.279 & 0.279 & \textbf{0.280} \\
 &F1 & \textbf{0.152} & \underline{0.148} & 0.134 & 0.131 & 0.126 \\

\bottomrule
\end{tabular}}
\end{table}

\subsection{Results of IDGG on Graph Structural Metrics}
Full results on Degree MMD, Spectra MMD, $D_k$, $\alpha$, and power-law validity under IDGG using other three LLM backbones are available in \cref{tab:idgg_structural_deepseek}, \ref{tab:idgg_structural_llama}, and \ref{tab:idgg_structural_qwen}.
Compared to the TDGG task, DyTAGs generated under IDGG exhibit slightly higher Degree/Spectra MMD values, primarily due to the introduction of new node generation, which increases the complexity of the modeling process. 
Nevertheless, over half of the generated DyTAG graphs still satisfy the power-law distribution criterion, demonstrating the robustness of the GAG-General framework in maintaining structural quality even under more challenging generation conditions.

\begin{table}[H]
\centering \caption{The results on Degree MMD, Spectra MMD, $D_k$, $\alpha$, and power-law validity under IDGG with Deepseek as the LLM backbone.} 
\label{tab:idgg_structural_deepseek}
\resizebox{0.9\linewidth}{!}{
    \begin{tabular}{l|cccccccc}
    \toprule
    Dataset & \textbf{Sephora} & \textbf{Dianping} & \textbf{WikiRevision} & \textbf{WikiLife} & \textbf{IMDB} & \textbf{WeiboTech} & \textbf{WeiboDaily} & \textbf{Cora} \\
    \midrule
    Degree MMD          & 0.370 & 0.150 & 0.191 & 0.087 & 0.329 & 0.212 & 0.219 & 0.073 \\
    Spectra MMD         & 0.250 & 0.351 & 0.158 & 0.206 & 0.440 & 0.172 & 0.397 & 0.181 \\
    $D_k$              & 0.099 & 0.056 & 0.029 & 0.099 & 0.238 & 0.059 & 0.131 & 0.084 \\
    $\alpha$           & 2.135 & 3.010 & 2.185 & 2.204 & 1.805 & 1.953 & 1.839 & 2.337 \\
    Power-law Validity & \ding{51}  & \ding{55}    & \ding{51}  & \ding{51}  & \ding{55}    & \ding{55}    & \ding{55}    & \ding{51} \\
    \bottomrule
\end{tabular}}
\end{table}

\begin{table}[H]
\centering \caption{The results on Degree MMD, Spectra MMD, $D_k$, $\alpha$, and power-law validity under IDGG with Llama as the LLM backbone.} 
\label{tab:idgg_structural_llama}
\resizebox{0.9\linewidth}{!}{
    \begin{tabular}{l|cccccccc}
    \toprule
    Dataset & \textbf{Sephora} & \textbf{Dianping} & \textbf{WikiRevision} & \textbf{WikiLife} & \textbf{IMDB} & \textbf{WeiboTech} & \textbf{WeiboDaily} & \textbf{Cora} \\
    \midrule
    Degree MMD          & 0.411 & 0.236 & 0.102& 0.078 & 0.145 & 0.231 & 0.246 & 0.123 \\
    Spectra MMD         & 0.243 & 0.413 & 0.105& 0.223 & 0.379 & 0.215 & 0.385 & 0.232 \\
    $D_k$              & 0.098 & 0.067 & 0.103 & 0.099 & 0.226 & 0.029 & 0.088 & 0.130 \\
    $\alpha$           & 2.077 & 2.480 & 2.300 & 2.204 & 1.972 & 2.099 & 1.968 & 2.174 \\
    Power-law Validity & \ding{51}  & \ding{51}  & \ding{51}  & \ding{51}  & \ding{55}    & \ding{51}  & \ding{55}    & \ding{51}\\
    \bottomrule
\end{tabular}}
\end{table}

\begin{table}[H]
\centering \caption{The results on Degree MMD, Spectra MMD, $D_k$, $\alpha$, and power-law validity under IDGG with Qwen as the LLM backbone. } 
\label{tab:idgg_structural_qwen}
\resizebox{0.9\linewidth}{!}{
    \begin{tabular}{l|cccccccc}
    \toprule
    Dataset & \textbf{Sephora} & \textbf{Dianping} & \textbf{WikiRevision} & \textbf{WikiLife} & \textbf{IMDB} & \textbf{WeiboTech} & \textbf{WeiboDaily} & \textbf{Cora} \\
    \midrule
    Degree MMD          & 0.512 & 0.321 & 0.286 & 0.083 & 0.303 & 0.237 & 0.295 & 0.095 \\
    Spectra MMD         & 0.234 & 0.436 & 0.193 & 0.218 & 0.415 & 0.189 & 0.461 & 0.192 \\
    $D_k$              & 0.174 & 0.076 & 0.039 & 0.099 & 0.207 & 0.050 & 0.152 & 0.093 \\
    $\alpha$           & 2.015 & 2.497 & 2.114 & 2.204 & 1.783 & 1.900 & 1.757 & 2.247 \\
    Power-law Validity & \ding{55}    & \ding{51}  & \ding{51}  & \ding{51}  & \ding{55}    & \ding{55}    & \ding{55}    & \ding{51}\\
    \bottomrule
\end{tabular}}
\end{table}

\subsection{Results of IDGG on Textual Quality Metrics}
Full results on average textual quality scores under IDGG are shown in \cref{tab:idgg_textual}. 
Detailed results on each scoring criterion are available in \cref{tab:idgg_textual_sephora}-\ref{tab:idgg_textual_cora}.
Similar to the findings under the TDGG task, the experimental results demonstrate that incorporating node memory or reflection mechanisms significantly enhances the textual quality of generated DyTAGs, highlighting the importance of integrating structural and textual information in DyTAG generation.

\begin{table}[H]
\centering \caption{The results on \textbf{average} textual quality scores under IDGG. M. and R. denote node memory and memory reflection mechanism, respectively.} 
\label{tab:idgg_textual}
\resizebox{1.0\linewidth}{!}{
\begin{tabular}{l|ccc|ccc|ccc|ccc}\toprule
 & \multicolumn{3}{c|}{\textbf{DeepSeek}} & \multicolumn{3}{c|}{\textbf{Llama}} & \multicolumn{3}{c|}{\textbf{Qwen}} & \multicolumn{3}{c}{\textbf{GPT}}\\
 & \multicolumn{1}{c}{w/o M.} & \multicolumn{1}{c}{w/ M.} & \multicolumn{1}{c|}{w/ M.R.} & \multicolumn{1}{c}{w/o M.} & \multicolumn{1}{c}{w/ M.} & \multicolumn{1}{c|}{w/ M.R.} & \multicolumn{1}{c}{w/o M.} & \multicolumn{1}{c}{w/ M.} & \multicolumn{1}{c|}{w/ M.R.} & \multicolumn{1}{c}{w/o M.} & \multicolumn{1}{c}{w/ M.} & \multicolumn{1}{c}{w/ M.R.} \\ 
 \midrule
Sephora& 4.63 & \underline{4.73} & \textbf{4.77} & 4.58 & \underline{4.65} &\textbf{4.74} & 4.58 & \underline{4.68} & \textbf{4.78} & 4.58 & \underline{4.77} & \textbf{4.87} \\
Dianping& 4.29 & \underline{4.44} & \textbf{4.65} & 4.29 & \underline{4.51} & \textbf{4.87} & 4.04 & \underline{4.50} & \textbf{4.68} & 4.56 & \underline{4.71} & \textbf{4.86} \\
WikiRevision& 4.21 & \underline{4.46} & \textbf{4.63} & 3.98 & \underline{4.21} & \textbf{4.30} & 4.37 & \underline{4.53} & \textbf{4.71} & 4.39 & \underline{4.54} & \textbf{4.71} \\
WikiLife& 4.15 & \underline{4.30} & \textbf{4.44} & 4.12 & \underline{4.26} & \textbf{4.38} & 4.11 & \underline{4.26} & \textbf{4.29} & 4.28 & \underline{4.39} & \textbf{4.47} \\
IMDB& 4.13 & \underline{4.28} & \textbf{4.39} & 4.22 & \underline{4.31} & \textbf{4.43} & 4.23 & \underline{4.36} & \textbf{4.51} & 4.19 & \underline{4.29} & \textbf{4.49} \\
WeiboTech& 4.60 & \underline{4.75} & \textbf{4.85} & 3.94 & \underline{4.00} & \textbf{4.75} & 4.56 & \underline{4.74} &\textbf{4.93} & 4.60 & \underline{4.71} & \textbf{4.93} \\
WeiboDaily& 4.68 & \underline{4.81} & \textbf{4.92} & 4.32 & \underline{4.43} & \textbf{4.82} & 4.75 & \underline{4.87} & \textbf{4.98} & 4.74 & \underline{4.83} & \textbf{4.99} \\
Cora& 4.27 & \underline{4.44} & \textbf{4.56} & 4.26 & \underline{4.36} & \textbf{4.54} & 4.29 & \underline{4.44} & \textbf{4.53} & 4.27 & \underline{4.36} & \textbf{4.57} \\
\bottomrule
\end{tabular}}
\end{table}

\begin{table}[H]
\centering 
\caption{The results on each scoring criterion of generated textual quality under IDGG on \textbf{Sephora}. } 
\label{tab:idgg_textual_sephora}
\resizebox{1.0\linewidth}{!}{
\begin{tabular}{l|ccc|ccc|ccc|ccc}\toprule
 & \multicolumn{3}{c|}{\textbf{DeepSeek}} & \multicolumn{3}{c|}{\textbf{Llama}} & \multicolumn{3}{c|}{\textbf{Qwen}} & \multicolumn{3}{c}{\textbf{GPT}}\\
 & \multicolumn{1}{c}{w/o M.} & \multicolumn{1}{c}{w/ M.} & \multicolumn{1}{c|}{w/ M.R.} & \multicolumn{1}{c}{w/o M.} & \multicolumn{1}{c}{w/ M.} & \multicolumn{1}{c|}{w/ M.R.} & \multicolumn{1}{c}{w/o M.} & \multicolumn{1}{c}{w/ M.} & \multicolumn{1}{c|}{w/ M.R.} & \multicolumn{1}{c}{w/o M.} & \multicolumn{1}{c}{w/ M.} & \multicolumn{1}{c}{w/ M.R.} \\ 
     \midrule
    Contextual Fidelity 
    & 4.63& \underline{4.80} & \textbf{4.87} 
    & 4.64 & \underline{4.73} & \textbf{4.83} 
    & 4.51& \underline{4.78} & \textbf{4.92} 
    & 4.77& \underline{4.84} & \textbf{4.93} \\
    Personality Depth 
    & 4.53& \underline{4.56} & \textbf{4.61} 
    & 4.55& \underline{4.52} & \textbf{4.63} 
    & 4.42& \underline{4.46} & \textbf{4.60} 
    & 4.20& \underline{4.61} & \textbf{4.78} \\
    Dynamic Adaptability 
    & 4.60& \underline{4.67} & \textbf{4.70} 
    & 4.41& \underline{4.59} & \textbf{4.66} 
    & 4.55& \underline{4.61} & \textbf{4.69} 
    & 4.54& \underline{4.71} & \textbf{4.82} \\
    Immersive Quality 
    & 4.71& \underline{4.83} & \textbf{4.87} 
    & 4.71& \underline{4.73} & \textbf{4.84} 
    & 4.77& \underline{4.81} & \textbf{4.91} 
    & 4.69& \underline{4.86} & \textbf{4.95} \\
    Content Richness 
    & 4.66& \underline{4.79} & \textbf{4.81} 
    & 4.59& \underline{4.67} & \textbf{4.74} 
    & 4.66& \underline{4.74} & \textbf{4.79} 
    & 4.72& \underline{4.80} & \textbf{4.86} \\
    \midrule
    Average 
    & 4.63 & \underline{4.73} & \textbf{4.77} 
    & 4.58 & \underline{4.65} & \textbf{4.74} 
    & 4.58 & \underline{4.68} & \textbf{4.78} 
    & 4.58 & \underline{4.77} & \textbf{4.87} \\
\bottomrule
\end{tabular}}
\end{table}

\begin{table}[H]
\centering \caption{The results on each scoring criterion of generated textual quality under IDGG on \textbf{Dianping}.} 
\label{tab:idgg_textual_dianping}
\resizebox{1.0\linewidth}{!}{
\begin{tabular}{l|ccc|ccc|ccc|ccc}\toprule
 & \multicolumn{3}{c|}{\textbf{DeepSeek}} & \multicolumn{3}{c|}{\textbf{Llama}} & \multicolumn{3}{c|}{\textbf{Qwen}} & \multicolumn{3}{c}{\textbf{GPT}}\\
 & \multicolumn{1}{c}{w/o M.} & \multicolumn{1}{c}{w/ M.} & \multicolumn{1}{c|}{w/ M.R.} & \multicolumn{1}{c}{w/o M.} & \multicolumn{1}{c}{w/ M.} & \multicolumn{1}{c|}{w/ M.R.} & \multicolumn{1}{c}{w/o M.} & \multicolumn{1}{c}{w/ M.} & \multicolumn{1}{c|}{w/ M.R.} & \multicolumn{1}{c}{w/o M.} & \multicolumn{1}{c}{w/ M.} & \multicolumn{1}{c}{w/ M.R.} \\ 
     \midrule
    Contextual Fidelity 
    & 4.40& \underline{4.50} & \textbf{4.85} 
    & 4.52& \underline{4.64} & \textbf{4.95} 
    & 3.89& \underline{4.57} & \textbf{4.77} 
    & 4.56& \underline{4.77} & \textbf{4.93} \\
    Personality Depth 
    & 4.03& \underline{4.21} & \textbf{4.32} 
    & 4.33& \underline{4.21} & \textbf{4.72} 
    & 4.01& \underline{4.21} & \textbf{4.45} 
    & 4.33& \underline{4.42} & \textbf{4.69} \\
    Dynamic Adaptability 
    & 4.21& \underline{4.37} & \textbf{4.50} 
    & 4.28& \underline{4.39} & \textbf{4.82} 
    & 3.98& \underline{4.43} & \textbf{4.61} 
    & 4.50& \underline{4.69} & \textbf{4.80} \\
    Immersive Quality 
    & 4.47& \underline{4.57} & \textbf{4.81} 
    & 4.11& \underline{4.67} & \textbf{4.95} 
    & 4.22& \underline{4.60} & \textbf{4.79} 
    & 4.70& \underline{4.82} & \textbf{4.94} \\
    Content Richness 
    & 4.33& \underline{4.57} & \textbf{4.79} 
    & 4.20& \underline{4.65} & \textbf{4.93} 
    & 4.10& \underline{4.68} & \textbf{4.80} 
    & 4.69& \underline{4.83} & \textbf{4.92} \\
    \midrule
    Average 
    & 4.29 & \underline{4.44} & \textbf{4.65} 
    & 4.29 & \underline{4.51} & \textbf{4.87} 
    & 4.04 & \underline{4.50} & \textbf{4.68} 
    & 4.56 & \underline{4.71} & \textbf{4.86} \\
\bottomrule
\end{tabular}}
\end{table}

\begin{table}[H]
\centering \caption{The results on each scoring criterion of generated textual quality under IDGG on \textbf{WikiRevision}.} 
\label{tab:idgg_textual_wikirevision}
\resizebox{1.0\linewidth}{!}{
\begin{tabular}{l|ccc|ccc|ccc|ccc}\toprule
 & \multicolumn{3}{c|}{\textbf{DeepSeek}} & \multicolumn{3}{c|}{\textbf{Llama}} & \multicolumn{3}{c|}{\textbf{Qwen}} & \multicolumn{3}{c}{\textbf{GPT}}\\
 & \multicolumn{1}{c}{w/o M.} & \multicolumn{1}{c}{w/ M.} & \multicolumn{1}{c|}{w/ M.R.} & \multicolumn{1}{c}{w/o M.} & \multicolumn{1}{c}{w/ M.} & \multicolumn{1}{c|}{w/ M.R.} & \multicolumn{1}{c}{w/o M.} & \multicolumn{1}{c}{w/ M.} & \multicolumn{1}{c|}{w/ M.R.} & \multicolumn{1}{c}{w/o M.} & \multicolumn{1}{c}{w/ M.} & \multicolumn{1}{c}{w/ M.R.} \\ 
     \midrule
    Contextual Fidelity 
    & \underline{4.88}& 4.86 & \textbf{4.94} 
    & 4.22& \underline{4.76} & \textbf{4.82} 
    & 4.82& \underline{4.90} & \textbf{4.98} 
    & 4.82& \underline{4.92} & \textbf{4.98} \\
    Personality Depth 
    & \underline{4.01}& 3.81 & \textbf{4.09} 
    & \underline{3.68}& 3.65 & \textbf{3.75} 
    & 3.81& \underline{3.90} & \textbf{4.17} 
    & 3.89& \underline{3.91} & \textbf{4.20} \\
    Dynamic Adaptability 
    & 3.92& \underline{4.28} & \textbf{4.44} 
    & 3.76& \underline{3.91} & \textbf{4.01} 
    & 4.21& \underline{4.32} & \textbf{4.56} 
    & 4.10& \underline{4.33} & \textbf{4.55} \\
    Immersive Quality 
    & 4.23& \underline{4.81} & \textbf{4.91} 
    & 4.20& \underline{4.60} & \textbf{4.69} 
    & 4.62& \underline{4.86} & \textbf{4.96} 
    & 4.83& \underline{4.89} & \textbf{4.96} \\
    Content Richness 
    & 4.01& \underline{4.53} & \textbf{4.76} 
    & 4.05& \underline{4.14} & \textbf{4.21} 
    & 4.40& \underline{4.66} & \textbf{4.86} 
    & 4.29& \underline{4.64} & \textbf{4.86} \\
    \midrule
    Average 
    & 4.21 & \underline{4.46} & \textbf{4.63} 
    & 3.98 & \underline{4.21} & \textbf{4.30} 
    & 4.37 & \underline{4.53} & \textbf{4.71} 
    & 4.39 & \underline{4.54} & \textbf{4.71} \\
\bottomrule
\end{tabular}}
\end{table}

\begin{table}[H]
\centering \caption{The results on each scoring criterion of generated textual quality under IDGG on \textbf{WikiLife}.} 
\label{tab:idgg_textual_wikilife}
\resizebox{1.0\linewidth}{!}{
\begin{tabular}{l|ccc|ccc|ccc|ccc}\toprule
 & \multicolumn{3}{c|}{\textbf{DeepSeek}} & \multicolumn{3}{c|}{\textbf{Llama}} & \multicolumn{3}{c|}{\textbf{Qwen}} & \multicolumn{3}{c}{\textbf{GPT}}\\
 & \multicolumn{1}{c}{w/o M.} & \multicolumn{1}{c}{w/ M.} & \multicolumn{1}{c|}{w/ M.R.} & \multicolumn{1}{c}{w/o M.} & \multicolumn{1}{c}{w/ M.} & \multicolumn{1}{c|}{w/ M.R.} & \multicolumn{1}{c}{w/o M.} & \multicolumn{1}{c}{w/ M.} & \multicolumn{1}{c|}{w/ M.R.} & \multicolumn{1}{c}{w/o M.} & \multicolumn{1}{c}{w/ M.} & \multicolumn{1}{c}{w/ M.R.} \\ 
     \midrule
    Contextual Fidelity 
    & 4.52& \underline{4.67} & \textbf{4.83} 
    & \underline{4.53}& \textbf{4.63} & \textbf{4.63} 
    & 4.30& \textbf{4.58} & \underline{4.57} 
    & 4.62& \underline{4.74} & \textbf{4.79} \\
    Personality Depth 
    & 3.63& \underline{3.77} & \textbf{3.90} 
    & 3.62& \underline{3.75} & \textbf{4.07} 
    & 3.69& \underline{3.73} & \textbf{3.81} 
    & 3.73& \underline{3.85} & \textbf{4.00} \\
    Dynamic Adaptability 
    & \underline{4.04}& 4.03 & \textbf{4.11} 
    & 3.92& \underline{4.02} & \textbf{4.11}
    & 3.92& \underline{4.02} & \textbf{4.06} 
    & 4.02& \underline{4.10} & \textbf{4.17} \\
    Immersive Quality 
    & 4.34& \underline{4.64} & \textbf{4.81} 
    & 4.42& \underline{4.60} & \textbf{4.62}
    & 4.32& \underline{4.56} & \textbf{4.57} 
    & 4.63& \underline{4.73} & \textbf{4.75} \\
    Content Richness 
    & 4.24& \underline{4.38} & \textbf{4.57} 
    & 4.13& \underline{4.28} & \textbf{4.46} 
    & 4.33& \underline{4.39} & \textbf{4.43} 
    & 4.42& \underline{4.55} & \textbf{4.66} \\
    \midrule
    Average 
    & 4.15 & \underline{4.30} & \textbf{4.44} 
    & 4.12 & \underline{4.26} & \textbf{4.38} 
    & 4.11 & \underline{4.26} & \textbf{4.29} 
    & 4.28 & \underline{4.39} & \textbf{4.47} \\
\bottomrule
\end{tabular}}
\end{table}

\begin{table}[H]
\centering \caption{The results on each scoring criterion of generated textual quality under IDGG on \textbf{IMDB}.} 
\label{tab:idgg_textual_imdb}
\resizebox{1.0\linewidth}{!}{
\begin{tabular}{l|ccc|ccc|ccc|ccc}\toprule
 & \multicolumn{3}{c|}{\textbf{DeepSeek}} & \multicolumn{3}{c|}{\textbf{Llama}} & \multicolumn{3}{c|}{\textbf{Qwen}} & \multicolumn{3}{c}{\textbf{GPT}}\\
 & \multicolumn{1}{c}{w/o M.} & \multicolumn{1}{c}{w/ M.} & \multicolumn{1}{c|}{w/ M.R.} & \multicolumn{1}{c}{w/o M.} & \multicolumn{1}{c}{w/ M.} & \multicolumn{1}{c|}{w/ M.R.} & \multicolumn{1}{c}{w/o M.} & \multicolumn{1}{c}{w/ M.} & \multicolumn{1}{c|}{w/ M.R.} & \multicolumn{1}{c}{w/o M.} & \multicolumn{1}{c}{w/ M.} & \multicolumn{1}{c}{w/ M.R.} \\ 
     \midrule
    Contextual Fidelity 
    & 4.53& \underline{4.71} & \textbf{4.88} 
    & 4.76& \underline{4.83} & \textbf{4.84} 
    & \underline{4.82} & 4.80 & \textbf{4.88} 
    & 4.65 & \underline{4.70} & \textbf{4.78} \\
    Personality Depth 
    & 3.72& \underline{3.81} & \textbf{3.84} 
    & 3.67& \underline{3.76} & \textbf{4.03} 
    & 3.78& \underline{3.87} & \textbf{4.17} 
    & 3.72& \underline{3.82} & \textbf{4.19} \\
    Dynamic Adaptability 
    & 3.89& \underline{4.06} & \textbf{4.10} 
    & 4.01& \underline{4.04} & \textbf{4.16} 
    & 4.02& \underline{4.17} & \textbf{4.25} 
    & 3.90& \underline{4.08} & \textbf{4.23} \\
    Immersive Quality 
    & 4.30& \underline{4.65} & \textbf{4.82} 
    & 4.58& \underline{4.72} & \textbf{4.80} 
    & 4.33& \underline{4.71} & \textbf{4.85} 
    & 4.56& \underline{4.65} & \textbf{4.77} \\
    Content Richness 
    & \underline{4.20} & 4.18 & \textbf{4.29} 
    & 4.10& \underline{4.20} & \textbf{4.30} 
    & 4.19& \underline{4.26} & \textbf{4.40} 
    & 4.11& \underline{4.18} & \textbf{4.48} \\
    \midrule
    Average 
    & 4.13 & \underline{4.28} & \textbf{4.39} 
    & 4.22 & \underline{4.31} & \textbf{4.43} 
    & 4.23 & \underline{4.36} & \textbf{4.51} 
    & 4.19 & \underline{4.29} & \textbf{4.49} \\
\bottomrule
\end{tabular}}
\end{table}

\begin{table}[H]
\centering \caption{The results on each scoring criterion of generated textual quality under IDGG on \textbf{WeiboTech}.} 
\label{tab:idgg_textual_weibotech}
\resizebox{1.0\linewidth}{!}{
\begin{tabular}{l|ccc|ccc|ccc|ccc}\toprule
 & \multicolumn{3}{c|}{\textbf{DeepSeek}} & \multicolumn{3}{c|}{\textbf{Llama}} & \multicolumn{3}{c|}{\textbf{Qwen}} & \multicolumn{3}{c}{\textbf{GPT}}\\
 & \multicolumn{1}{c}{w/o M.} & \multicolumn{1}{c}{w/ M.} & \multicolumn{1}{c|}{w/ M.R.} & \multicolumn{1}{c}{w/o M.} & \multicolumn{1}{c}{w/ M.} & \multicolumn{1}{c|}{w/ M.R.} & \multicolumn{1}{c}{w/o M.} & \multicolumn{1}{c}{w/ M.} & \multicolumn{1}{c|}{w/ M.R.} & \multicolumn{1}{c}{w/o M.} & \multicolumn{1}{c}{w/ M.} & \multicolumn{1}{c}{w/ M.R.} \\ 
     \midrule
    Contextual Fidelity 
    & 4.77& \underline{4.89} & \textbf{4.94} 
    & 4.22& \underline{4.23} & \textbf{4.83} 
    & 4.89& \underline{4.90} & \textbf{4.99} 
    & 4.82& \underline{4.91} & \textbf{4.99} \\
    Personality Depth 
    & 4.30& \underline{4.50} & \textbf{4.72} 
    & 3.65& \underline{3.76} & \textbf{4.68} 
    & 4.22& \underline{4.49} & \textbf{4.83}
    & 4.33& \underline{4.43} & \textbf{4.84} \\
    Dynamic Adaptability 
    & 4.42& \underline{4.75} & \textbf{4.84} 
    & 3.95& \underline{3.96} & \textbf{4.70} 
    & 4.35& \underline{4.77} & \textbf{4.94} 
    & 4.65& \underline{4.73} & \textbf{4.92} \\
    Immersive Quality 
    & 4.83& \underline{4.87} & \textbf{4.93} 
    & 4.02& \underline{4.16} & \textbf{4.82} 
    & 4.76& \underline{4.89} & \textbf{4.98} 
    & 4.67& \underline{4.88} & \textbf{4.99} \\
    Content Richness 
    & 4.69& \underline{4.72} & \textbf{4.83} 
    & 3.86& \underline{3.87} & \textbf{4.70} 
    & 4.57& \underline{4.66} & \textbf{4.91} 
    & 4.52& \underline{4.60} & \textbf{4.89} \\
    \midrule
    Average 
    & 4.60 & \underline{4.75} & \textbf{4.85} 
    & 3.94 & \underline{4.00} & \textbf{4.75} 
    & 4.56 & \underline{4.74} & \textbf{4.93} 
    & 4.60 & \underline{4.71} & \textbf{4.93} \\
\bottomrule
\end{tabular}}
\end{table}

\begin{table}[H]
\centering \caption{The results on each scoring criterion of generated textual quality under IDGG on \textbf{WeiboDaily}.} 
\label{tab:idgg_textual_weibodaily}
\resizebox{1.0\linewidth}{!}{
\begin{tabular}{l|ccc|ccc|ccc|ccc}\toprule
 & \multicolumn{3}{c|}{\textbf{DeepSeek}} & \multicolumn{3}{c|}{\textbf{Llama}} & \multicolumn{3}{c|}{\textbf{Qwen}} & \multicolumn{3}{c}{\textbf{GPT}}\\
 & \multicolumn{1}{c}{w/o M.} & \multicolumn{1}{c}{w/ M.} & \multicolumn{1}{c|}{w/ M.R.} & \multicolumn{1}{c}{w/o M.} & \multicolumn{1}{c}{w/ M.} & \multicolumn{1}{c|}{w/ M.R.} & \multicolumn{1}{c}{w/o M.} & \multicolumn{1}{c}{w/ M.} & \multicolumn{1}{c|}{w/ M.R.} & \multicolumn{1}{c}{w/o M.} & \multicolumn{1}{c}{w/ M.} & \multicolumn{1}{c}{w/ M.R.} \\ 
     \midrule
    Contextual Fidelity 
    & 4.82& \underline{4.90} & \textbf{4.96} 
    & 4.34& \underline{4.57} & \textbf{4.86} 
    & 4.82& \underline{4.93} & \textbf{4.99} 
    & 4.82& \underline{4.90} & \textbf{4.99} \\
    Personality Depth 
    & 4.52& \underline{4.69} & \textbf{4.87} 
    & 4.19& \underline{4.29} & \textbf{4.79} 
    & 4.72& \underline{4.79} & \textbf{4.97} 
    & 4.69& \underline{4.72} & \textbf{4.98} \\
    Dynamic Adaptability 
    & 4.64& \underline{4.80} & \textbf{4.91} 
    & \underline{4.42} & 4.38 & \textbf{4.79} 
    & 4.82& \underline{4.86} & \textbf{4.97} 
    & 4.71& \underline{4.85} & \textbf{4.99} \\
    Immersive Quality 
    & 4.82& \underline{4.90} & \textbf{4.96} 
    & 4.32& \underline{4.56} & \textbf{4.86} 
    & 4.72& \underline{4.94} & \textbf{4.99} 
    & 4.82& \underline{4.90} & \textbf{4.99} \\
    Content Richness 
    & 4.62& \underline{4.77} & \textbf{4.91} 
    & 4.32& \underline{4.34} & \textbf{4.79} 
    & 4.68& \underline{4.85} & \textbf{4.97} 
    & 4.67& \underline{4.80} & \textbf{4.98} \\
    \midrule
    Average 
    & 4.68 & \underline{4.81} & \textbf{4.92} 
    & 4.32 & \underline{4.43} & \textbf{4.82} 
    & 4.75 & \underline{4.87} & \textbf{4.98} 
    & 4.74 & \underline{4.83} & \textbf{4.99} \\
\bottomrule
\end{tabular}}
\end{table}

\begin{table}[H]
\centering \caption{The results on each scoring criterion of generated textual quality under IDGG on \textbf{Cora}. } 
\label{tab:idgg_textual_cora}
\resizebox{1.0\linewidth}{!}{
\begin{tabular}{l|ccc|ccc|ccc|ccc}\toprule
 & \multicolumn{3}{c|}{\textbf{DeepSeek}} & \multicolumn{3}{c|}{\textbf{Llama}} & \multicolumn{3}{c|}{\textbf{Qwen}} & \multicolumn{3}{c}{\textbf{GPT}}\\
 & \multicolumn{1}{c}{w/o M.} & \multicolumn{1}{c}{w/ M.} & \multicolumn{1}{c|}{w/ M.R.} & \multicolumn{1}{c}{w/o M.} & \multicolumn{1}{c}{w/ M.} & \multicolumn{1}{c|}{w/ M.R.} & \multicolumn{1}{c}{w/o M.} & \multicolumn{1}{c}{w/ M.} & \multicolumn{1}{c|}{w/ M.R.} & \multicolumn{1}{c}{w/o M.} & \multicolumn{1}{c}{w/ M.} & \multicolumn{1}{c}{w/ M.R.} \\ 
     \midrule
    Contextual Fidelity 
    & 4.64& \underline{4.84} & \textbf{4.94} 
    & 4.61& \underline{4.77} & \textbf{4.92} 
    & 4.63& \underline{4.82} & \textbf{4.91} 
    & 4.72& \underline{4.78} & \textbf{4.93} \\
    Personality Depth 
    & 3.69& \underline{3.71} & \textbf{3.80} 
    & 3.57& \underline{3.60} & \textbf{3.74} 
    & 3.51& \underline{3.69} & \textbf{3.76} 
    & \underline{3.68} & 3.58 & \textbf{3.80} \\
    Dynamic Adaptability 
    & 4.25& \underline{4.26} & \textbf{4.36} 
    & 4.20& \underline{4.21} & \textbf{4.44} 
    & 4.24& \underline{4.28} & \textbf{4.40} 
    & 4.11& \underline{4.19} & \textbf{4.43} \\
    Immersive Quality 
    & 4.26& \underline{4.81} & \textbf{4.92} 
    & 4.65& \underline{4.73} & \textbf{4.89} 
    & 4.76& \underline{4.80} & \textbf{4.88} 
    & 4.36& \underline{4.72} & \textbf{4.89} \\
    Content Richness 
    & 4.53& \underline{4.59} & \textbf{4.80} 
    & 4.25& \underline{4.50} & \textbf{4.71} 
    & 4.31& \underline{4.59} & \textbf{4.70} 
    & 4.49& \underline{4.51} & \textbf{4.78} \\
    \midrule
    Average 
    & 4.27 & \underline{4.44} & \textbf{4.56} 
    & 4.26 & \underline{4.36} & \textbf{4.54} 
    & 4.29 & \underline{4.44} & \textbf{4.53} 
    & 4.27 & \underline{4.36} & \textbf{4.57} \\
\bottomrule
\end{tabular}}
\end{table}

\subsection{Results of IDGG on Graph Embedding Metrics}
Full results on the graph embedding metric under IDGG are provided in \cref{tab:idgg_graph_embedding}, exhibiting the importance of incorporating structural and textual information in DyTAG generation.

\begin{table}[H]
\centering \caption{The results on the graph embedding metric under IDGG.} 
\label{tab:idgg_graph_embedding}
\resizebox{1.0\linewidth}{!}{
\begin{tabular}{l|ccc|ccc|ccc|ccc}\toprule
 & \multicolumn{3}{c|}{\textbf{DeepSeek}} & \multicolumn{3}{c|}{\textbf{Llama}} & \multicolumn{3}{c|}{\textbf{Qwen}} & \multicolumn{3}{c}{\textbf{GPT}}\\
 & \multicolumn{1}{c}{w/o M.} & \multicolumn{1}{c}{w/ M.} & \multicolumn{1}{c|}{w/ M.R.} & \multicolumn{1}{c}{w/o M.} & \multicolumn{1}{c}{w/ M.} & \multicolumn{1}{c|}{w/ M.R.} & \multicolumn{1}{c}{w/o M.} & \multicolumn{1}{c}{w/ M.} & \multicolumn{1}{c|}{w/ M.R.} & \multicolumn{1}{c}{w/o M.} & \multicolumn{1}{c}{w/ M.} & \multicolumn{1}{c}{w/ M.R.} \\ 
 \midrule
Sephora        & 0.621& \textbf{0.661} & \underline{0.634} & 0.602& \underline{0.612} & \textbf{0.647} & 0.569& \underline{0.587} & \textbf{0.616} & 0.601& \underline{0.603} & \textbf{0.628} \\
Dianping       & 0.739& \underline{0.749} & \textbf{0.769} & 0.426& \underline{0.457} & \textbf{0.512} & 0.531& \underline{0.542} & \textbf{0.578} & \underline{0.521}& \textbf{0.527} & 0.505 \\
WikiRevision   & 0.673& \underline{0.682} & \textbf{0.699} & 0.533& \underline{0.579} & \textbf{0.658} & 0.578& \underline{0.608} & \textbf{0.621} & 0.589& \underline{0.614} & \textbf{0.629} \\
WikiLife       & 0.542& \underline{0.555} & \textbf{0.804} & 0.501& \underline{0.518} & \textbf{0.536} & 0.541& \underline{0.553} & \textbf{0.549} & 0.510& \textbf{0.534} & \underline{0.531} \\
IMDB           & 0.601& \underline{0.616} & \textbf{0.729} & 0.511& \underline{0.522} & \textbf{0.568} & 0.502& \underline{0.557} & \textbf{0.588} & 0.521& \underline{0.535} & \textbf{0.563} \\
WeiboTech      & \textbf{0.604}& \underline{0.591} & 0.562 & 0.601& \underline{0.629} & \textbf{0.695} & 0.531& \underline{0.547} & \textbf{0.549} & 0.501& \underline{0.514} & \textbf{0.536} \\
WeiboDaily     & 0.541& \textbf{0.567} & \underline{0.555} & 0.620& \underline{0.623} & \textbf{0.638} & 0.498& \textbf{0.513} & \underline{0.503} & 0.459& \underline{0.481} & \textbf{0.502} \\
Cora           & 0.610& \underline{0.623} & \textbf{0.828} & \textbf{0.655}& \underline{0.642} & 0.635 & 0.526& \underline{0.599} & \textbf{0.634} & 0.541& \underline{0.552} & \textbf{0.572} \\

\bottomrule
\end{tabular}}
\end{table}

{
\subsection{Results of VRDAG, DG-Gen, {and TIGGER-I}}}
Comprehensive results on the graph structural and the graph embedding metrics under IDGG of GAG-General, VRDAG, DG-Gen, {and TIGGER-I} are shown in \cref{tab:idgg_vrdag_dggen}.
Since VRDAG \citep{VRDAG}, DG-Gen \citep{DGGEN}, {and TIGGER-I \citep{TIGGER}} do not support text generation, textual quality comparisons are not feasible. 
Furthermore, node/edge representations generated by VRDAG and DG-Gen are directly used as node/edge features when computing graph embeddings. 
The experimental results demonstrate that GAG-General significantly outperforms VRDAG, DG-Gen, {and TIGGER-I} in both graph structural quality and graph embedding quality. 
Notably, VRDAG {and TIGGER-I} struggles to generate high-quality datasets across all eight GDGB datasets, while DG-Gen achieves marginally better performance due to its explicit control over the size of generated graphs. 
These findings highlight the limitations of existing dynamic graph generation models in the DyTAG generation task and underscore the need for future models to effectively balance structural fidelity, temporal dynamics, and textual richness.

\begin{table}[H]
\caption{The results on the graph structural and the graph embedding metrics under IDGG of our framework and current feature-supportive dynamic graph generation models. Ours correspond to the best results of our proposed GAG-General among four LLM backbones.} 
\centering
\label{tab:idgg_vrdag_dggen}
\resizebox{0.65\textwidth}{!}{ 
\begin{tabular}{l|c|cccc}
\toprule
\textbf{Datasets}
&\textbf{Models}
&\multicolumn{1}{c}{\textbf{Ours}}

&\multicolumn{1}{c}{\textbf{VRDAG}}
&\multicolumn{1}{c}{\textbf{DG-Gen}} &
{\textbf{TIGGER-I}}\\
\midrule

\multirow{4}{*}{Sephora}
&Degree MMD $\downarrow$ &\textbf{0.370} &0.795 &0.422  &{0.622}\\
&Spectra MMD $\downarrow$ &\textbf{0.189} &0.847 &0.274  &{0.687}\\
&Power-law Validity &\ding{51} &\ding{55} &\ding{55}  &{\ding{55}} \\
&Graph Embedding $\uparrow$ &\textbf{0.661} &0.011 &0.228  &{0.085} \\

\midrule

\multirow{4}{*}{Dianping}
&Degree MMD $\downarrow$ &\textbf{0.150} &0.887 &0.167  &{0.446}\\
&Spectra MMD $\downarrow$ &0.351 &0.808 &\textbf{0.245}  &{{0.341}}\\
&Power-law Validity &\ding{51} &\ding{55} &\ding{51}  &{\ding{51}}\\
&Graph Embedding $\uparrow$ &\textbf{0.769} &0.024 &0.517  &{0.580}\\

\midrule

\multirow{4}{*}{WikiRevision}
&Degree MMD $\downarrow$ &\textbf{0.102} &0.807 &0.197  &{0.658}\\
&Spectra MMD $\downarrow$ &\textbf{0.105} &0.776 &0.159  &{0.606}\\
&Power-law Validity &\ding{51} &\ding{55} &\ding{51}  &{\ding{55}}\\
&Graph Embedding $\uparrow$ &\textbf{0.699} &0.089 &0.121  &{0.105} \\

\midrule

\multirow{4}{*}{WikiLife}
&Degree MMD $\downarrow$ &\textbf{0.078} &0.464 &0.283  &{0.418}\\
&Spectra MMD $\downarrow$ &\textbf{0.206} &0.236 &0.288  &{0.254}\\
&Power-law Validity &\ding{51} &\ding{55} &\ding{55}  &{\ding{55}} \\
&Graph Embedding $\uparrow$ &\textbf{0.804} &0.088 &0.303  &{0.156}\\

\midrule

\multirow{4}{*}{IMDB}
&Degree MMD $\downarrow$ &\textbf{0.145} &0.882 &0.373  &{0.679}\\
&Spectra MMD $\downarrow$ &\textbf{0.379} &0.710 &0.537  &{0.653}\\
&Power-law Validity &\ding{55} &\ding{55} &\ding{55}  &{\ding{55}} \\
&Graph Embedding $\uparrow$ &\textbf{0.729} &0.095 &0.293  &{0.159}\\

\midrule

\multirow{4}{*}{WeiboTech}
&Degree MMD $\downarrow$ &0.212 &0.867 &\textbf{0.211}  &{{0.313}}\\
&Spectra MMD $\downarrow$ &\textbf{0.172} &0.778 &0.311  &{0.384}\\
&Power-law Validity &\ding{51} &\ding{55} &\ding{51}  &{\ding{51}} \\
&Graph Embedding $\uparrow$ &\textbf{0.695} &0.085 &0.292  &{0.289}\\

\midrule

\multirow{4}{*}{WeiboDaily}
&Degree MMD $\downarrow$ &\textbf{0.219} &0.871 &0.259  &{0.465}\\
&Spectra MMD $\downarrow$ &\textbf{0.385} &0.791 &0.553  &{0.589}\\
&Power-law Validity &\ding{55} &\ding{55} &\ding{51}  &{\ding{55}}\\
&Graph Embedding $\uparrow$ &\textbf{0.638} &0.108 &0.294  &{0.241}\\

\midrule

\multirow{4}{*}{Cora}
&Degree MMD $\downarrow$ &\textbf{0.073} &0.877 &0.212  &{0.372}\\
&Spectra MMD $\downarrow$ &\textbf{0.181} &0.760 &0.365  &{0.389}\\
&Power-law Validity &\ding{51} &\ding{55} &\ding{55}  &{\ding{55}} \\
&Graph Embedding $\uparrow$ &\textbf{0.828} &0.053 &0.056  &{0.197}\\

\bottomrule
\end{tabular}}
\end{table}

\subsection{Results of IDGG on Utility in Data Augmentation for Inductive Learning}\label{IDGG-inductive}
We further investigate the utility of DyTAGs generated under the IDGG task for data augmentation in inductive learning scenarios, which simulate real-world cold-start problems such as recommending new items to users. We augment the training data of the original graph with the generated DyTAG (Aug.) and evaluate performance on an inductive link prediction task, where the goal is to predict future links involving nodes not observed during training. The DGNN models and the experimental settings that we used for evaluation are identical to those in \cref{TDGG-downstream}.

As shown in \cref{tab:idgg_downstream_inductive}, augmenting training data with IDGG-generated graphs consistently improves model performance across all datasets and models. The improvement in Average Precision ($\Delta AP$) ranges from 0.027 to 0.108, with particularly notable gains on WikiLife and Cora. For example, CAWN achieves a $\Delta AP$ of 0.108 on WikiLife, and DyGFormer shows a 0.046 improvement on Cora.
These results demonstrate that IDGG-generated DyTAGs provide valuable structural and textual context for unseen nodes, effectively enriching the training signal and enhancing model generalization in cold-start settings. This validates the practical value of our generative framework in improving downstream model robustness through synthetic data augmentation.

\begin{table}[ht]
\centering
\caption{The results of link prediction (inductive) performance with data augmentation from the graphs generated from IDGG.}
\label{tab:idgg_downstream_inductive}
\resizebox{0.55\textwidth}{!}{
\begin{tabular}{l|l|l|c|c}
\toprule
\textbf{Model} & \textbf{Dataset} & \textbf{Type} & \textbf{Average Precision} & $\Delta AP\uparrow$ \\
\midrule
\multirow{8}{*}{\textbf{TGN}} 
& \multirow{2}{*}{Sephora} & Aug. & 0.673 $\pm$ 0.010 & \multirow{2}{*}{0.037} \\
& & Ori. & 0.636 $\pm$ 0.002 & \\
\cmidrule{2-5}
& \multirow{2}{*}{WikiLife} & Aug. & 0.667 $\pm$ 0.020 & \multirow{2}{*}{0.099} \\
& & Ori. & 0.568 $\pm$ 0.024 & \\
\cmidrule{2-5}
& \multirow{2}{*}{Cora} & Aug. & 0.552 $\pm$ 0.030 & \multirow{2}{*}{0.034} \\
& & Ori. & 0.518 $\pm$ 0.019 & \\
\cmidrule{2-5}
& \multirow{2}{*}{WeiboTech} & Aug. & 0.741 $\pm$ 0.022 & \multirow{2}{*}{0.041} \\
& & Ori. & 0.700 $\pm$ 0.027 & \\
\midrule
\multirow{8}{*}{\textbf{CAWN}} 
& \multirow{2}{*}{Sephora} & Aug. & 0.750 $\pm$ 0.014 & \multirow{2}{*}{0.050} \\
& & Ori. & 0.700 $\pm$ 0.013 & \\
\cmidrule{2-5}
& \multirow{2}{*}{WikiLife} & Aug. & 0.741 $\pm$ 0.010 & \multirow{2}{*}{0.108} \\
& & Ori. & 0.633 $\pm$ 0.006 & \\
\cmidrule{2-5}
& \multirow{2}{*}{Cora} & Aug. & 0.651 $\pm$ 0.017 & \multirow{2}{*}{0.043} \\
& & Ori. & 0.608 $\pm$ 0.010 & \\
\cmidrule{2-5}
& \multirow{2}{*}{WeiboTech} & Aug. & 0.740 $\pm$ 0.009 & \multirow{2}{*}{0.033} \\
& & Ori. & 0.707 $\pm$ 0.009 & \\
\midrule
\multirow{8}{*}{\textbf{GraphMixer}} 
& \multirow{2}{*}{Sephora} & Aug. & 0.710 $\pm$ 0.006 & \multirow{2}{*}{0.027} \\
& & Ori. & 0.683 $\pm$ 0.025 & \\
\cmidrule{2-5}
& \multirow{2}{*}{WikiLife} & Aug. & 0.696 $\pm$ 0.004 & \multirow{2}{*}{0.031} \\
& & Ori. & 0.665 $\pm$ 0.030 & \\
\cmidrule{2-5}
& \multirow{2}{*}{Cora} & Aug. & 0.575 $\pm$ 0.003 & \multirow{2}{*}{0.040} \\
& & Ori. & 0.535 $\pm$ 0.018 & \\
\cmidrule{2-5}
& \multirow{2}{*}{WeiboTech} & Aug. & 0.720 $\pm$ 0.006 & \multirow{2}{*}{0.058} \\
& & Ori. & 0.662 $\pm$ 0.029 & \\
\midrule
\multirow{8}{*}{\textbf{DyGFormer}} 
& \multirow{2}{*}{Sephora} & Aug. & 0.691 $\pm$ 0.009 & \multirow{2}{*}{0.051} \\
& & Ori. & 0.640 $\pm$ 0.003 & \\
\cmidrule{2-5}
& \multirow{2}{*}{WikiLife} & Aug. & 0.734 $\pm$ 0.022 & \multirow{2}{*}{0.039} \\
& & Ori. & 0.695 $\pm$ 0.023 & \\
\cmidrule{2-5}
& \multirow{2}{*}{Cora} & Aug. & 0.599 $\pm$ 0.019 & \multirow{2}{*}{0.046} \\
& & Ori. & 0.553 $\pm$ 0.020 & \\
\cmidrule{2-5}
& \multirow{2}{*}{WeiboTech} & Aug. & 0.736 $\pm$ 0.008 & \multirow{2}{*}{0.040} \\
& & Ori. & 0.696 $\pm$ 0.016 & \\
\bottomrule
\end{tabular}
}
\end{table}

\subsection{Visualizations of the Hub Node Structures}
\begin{figure}[h]
  \centering
  \includegraphics[width=0.90\linewidth,page=1]{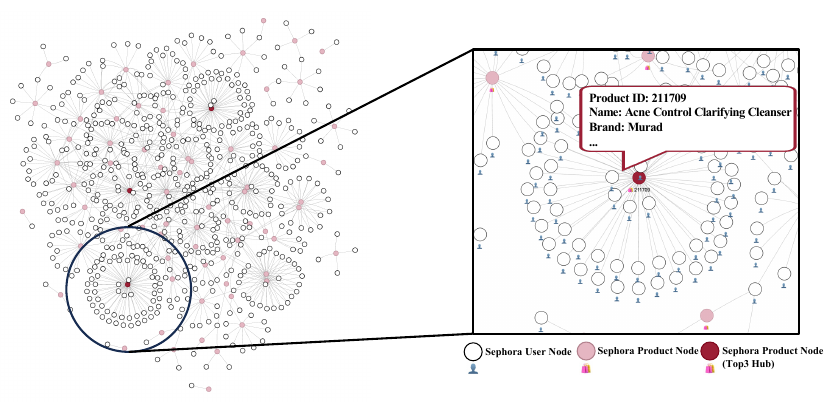}
  \caption{The visualization of the hub node structures in the ground-truth graph.}
   \vspace{-3mm}
  \label{fig:hub-nodes-gt}
\end{figure}

\begin{figure}[h]
  \centering
  \includegraphics[width=0.90\linewidth,page=2]{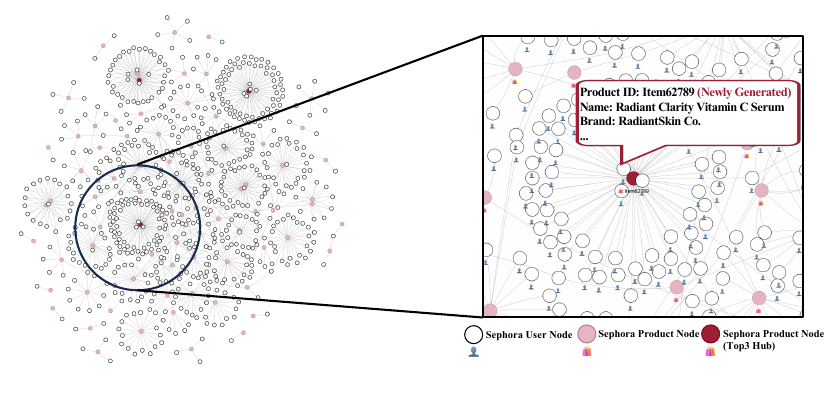}
  \caption{The visualization of the hub node structures in the generated graph.}
   \vspace{-3mm}
  \label{fig:hub-nodes-gen}
\end{figure}

The visualization of hub node structures in both the ground-truth and generated graphs from IDGG is presented in \cref{fig:hub-nodes-gt,fig:hub-nodes-gen}. The top-3 hub product nodes are highlighted in dark red.
As described in \cref{sec:exp_idgg}, both graphs exhibit similar hub node patterns, yet the generated graph contains hub nodes formed through inductive generation, featuring entirely distinct textual profiles.

This divergence arises because IDGG emulates real-world graph evolution dynamics, preserving structural fidelity while generating new nodes with plausible attributes that align with the underlying generative mechanisms of real systems.
For example, in recommendation systems, hub nodes in the generated DyTAG may represent emerging products with high virality potential, whereas those in the ground-truth graph correspond to established bestsellers.
This capability establishes DyTAG generation as a strategic tool for proactive decision-making in e-commerce and digital marketing. By identifying potential future hubs, platforms can prioritize resource allocation for product promotion, optimize advertising strategies, and anticipate market trends before they emerge in real-world data.

{
\subsection{Semantic-drift Analysis for IDGG Evaluation}
\label{subsec:semantic_drift}

To assess the long-term semantic consistency of generated DyTAGs in the IDGG task, we introduce a cross-snapshot semantic drift detection evaluation. 
Specifically, we partition the generated 10K-edge DyTAG into 10 temporal snapshots $\{\mathcal{G}_t\}_{t=1}^{10}$, each containing 1K newly added edges and corresponding newly generated nodes.
For each snapshot $\mathcal{G}_t$, we compute a global semantic representation $\mathbf{h}_t$ by encoding all textual node and edge attributes using {BERT-base-uncased}\citep{BERT} and aggregating via mean pooling:
\begin{equation}
    \mathbf{h}_t = \frac{1}{|\mathcal{V}_t| + |\mathcal{E}_t|} \left( \sum_{v \in \mathcal{V}_t} \text{BERT}(v.\text{text}) + \sum_{e \in \mathcal{E}_t} \text{BERT}(e.\text{text}) \right),
\end{equation}
where $\mathcal{V}_t$ and $\mathcal{E}_t$ denote the sets of nodes and edges introduced in the current snapshot $t$. We then measure inter-temporal semantic stability via cosine similarity between consecutive representations:
\begin{equation}
    s_t = \cos(\mathbf{h}_t, \mathbf{h}_{t+1}), \quad t = 1, \dots, 9.
\end{equation}
Thus, higher $s_t$ indicates greater semantic coherence over time. 

On the Sephora dataset, as shown in \cref{fig:semantic_drift}, BERT-based similarity metrics reveal a gradual decline after 5K edges under GAG-General—suggesting emerging semantic drift (e.g., generation of out-of-domain items like ``makeup bag''). Importantly, we further analyze the semantic drift under guided generation. As shown in \cref{fig:semantic_drift}, we find that the drift is significantly reduced when domain-specific constraints (e.g., “only generate reviews of makeup products (edge generation)/core makeup products (node generation), which are possible to be shown on Sephora platform”) are incorporated into the prompts for edge generation (\cref{prompts:sephora_interaction}) or node generation (\cref{prompts:sephora_node_generation}), demonstrating the effectiveness of guided generation in future work.

\begin{figure}[h]
  \centering
  \includegraphics[width=0.31\linewidth]{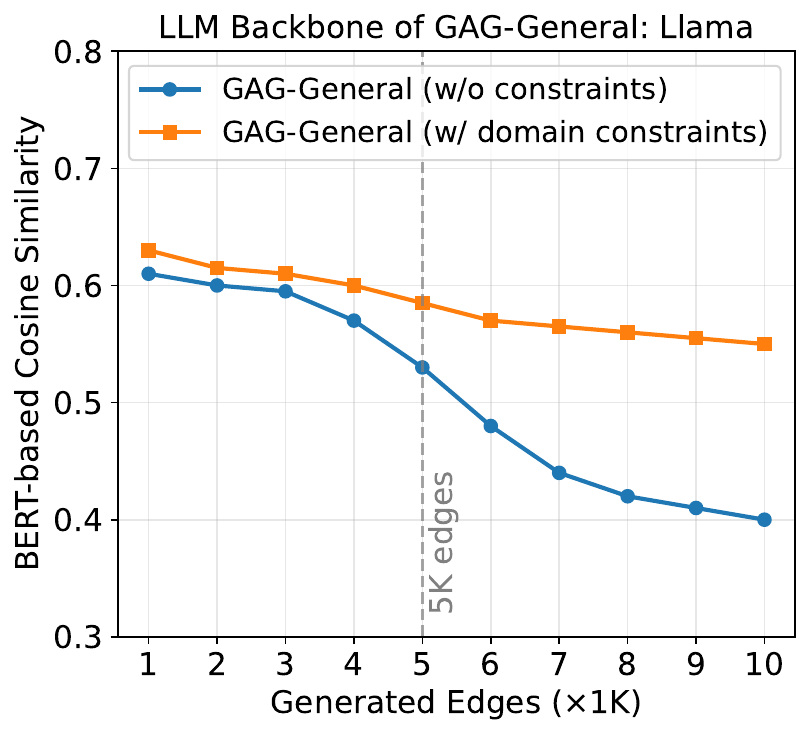}
  \includegraphics[width=0.31\linewidth]{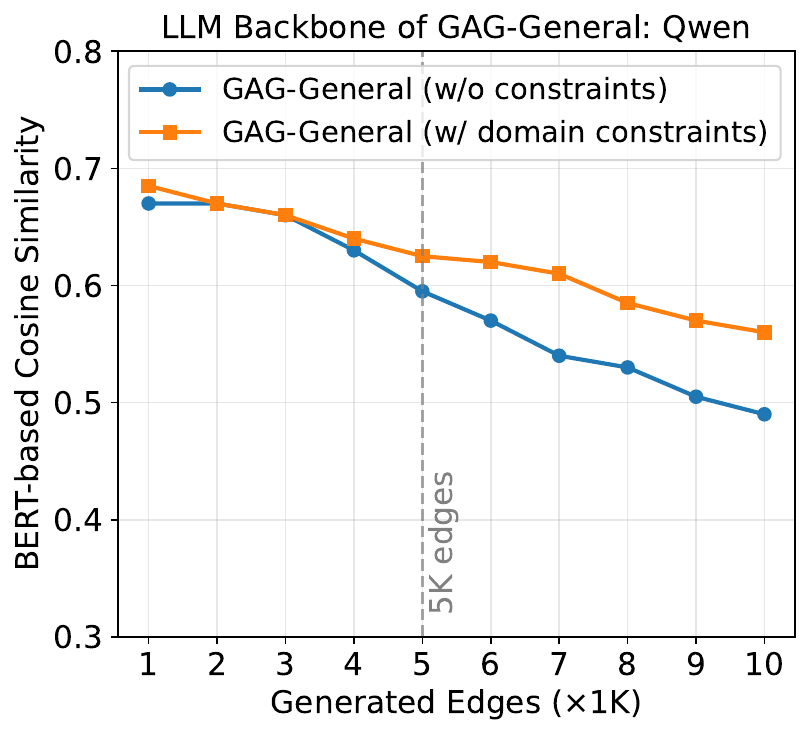}
  \includegraphics[width=0.31\linewidth]{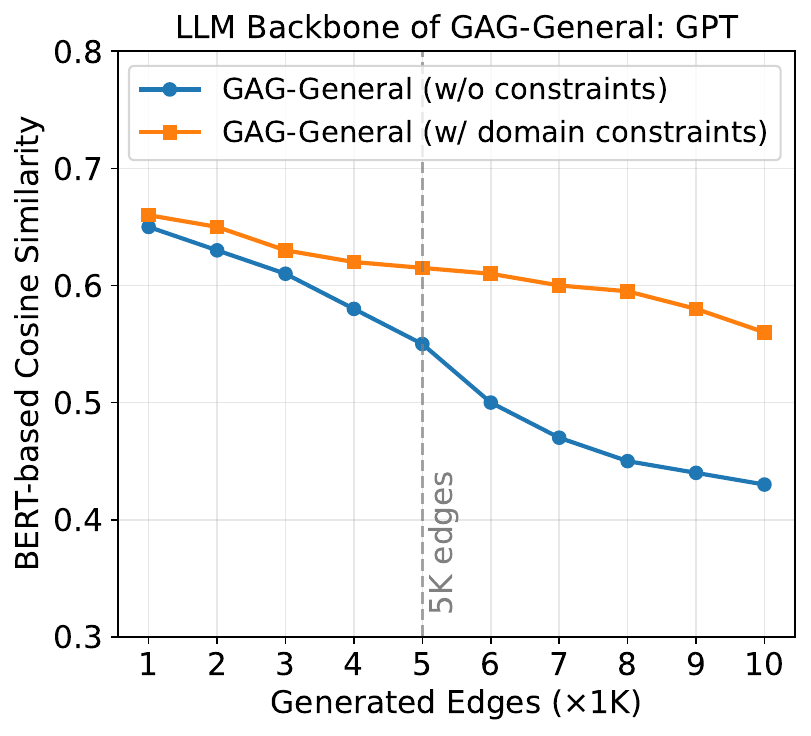}
  \caption{{Cross-snapshot semantic consistency over three LLM backbones in IDGG on Sephora.}}
   \vspace{-3mm}
  \label{fig:semantic_drift}
\end{figure}

\subsection{Human-amenity Check for IDGG Evaluation}
\label{subsec:human_amenity}

To evaluate the practical usability and naturalness of generated content, we conduct a small-scale human evaluation on the IDGG-generated DyTAGs. Five skilled annotators independently assessed 50 randomly sampled newly generated nodes and their interactions (e.g., reviews, revisions, citations, etc.). Each item was rated on a 5-point Likert scale across three dimensions: \emph{naturalness}, \emph{plausibility}, and \emph{consistency} with platform context.

The results are shown in \cref{tab:human_evaluation}. For instance, the results on Sephora with GPT as the LLM backbone show that generated nodes (users/products) achieve an average score of 3.9, while generated edges (e.g., user reviews) receive a higher average of 4.3. These scores indicate that the majority of generated content is perceived as realistic, contextually appropriate, and aligned with real-world user behavior on the target platform—supporting the human amenity of our framework in practical deployment scenarios.

\begin{table}[H]
\centering
\caption{{Human evaluation average scores (5-point Likert scale) for generated nodes and edges under IDGG across different LLM backbones and datasets. Higher scores indicate better naturalness, plausibility, and consistency.}}
\label{tab:human_evaluation}
\resizebox{0.7\linewidth}{!}{
\begin{tabular}{l|cc|cc|cc|cc}
\toprule
 & \multicolumn{2}{c|}{\textbf{DeepSeek}} & \multicolumn{2}{c|}{\textbf{Llama}} & \multicolumn{2}{c|}{\textbf{Qwen}} & \multicolumn{2}{c}{\textbf{GPT}} \\
 & Node & Edge & Node & Edge & Node & Edge & Node & Edge \\
\midrule
Sephora        & 3.7 & 4.2 & 3.6 & 4.1 & 3.8 & 4.3 & \textbf{3.9} & \textbf{4.3} \\
Dianping       & 3.5 & 4.0 & \textbf{4.0} & \textbf{4.5} & 3.6 & 4.1 & {3.8} & {4.2} \\
WikiRevision   & 3.3 & 3.7 & 3.2 & 3.6 & \textbf{3.6} & \textbf{4.0} & {3.5} & {3.9} \\
WikiLife       & 3.6 & 4.1 & 3.5 & 4.0 & 3.7 & 4.2 & \textbf{3.8} & \textbf{4.3} \\
IMDB           & \textbf{3.8} & \textbf{4.3} & 3.3 & 3.8 & 3.5 & 4.0 & {3.6} & {4.1} \\
WeiboTech      & 3.2 & 3.6 & 3.1 & 3.5 & 3.3 & 3.7 & \textbf{3.4} & \textbf{3.8} \\
WeiboDaily     & 3.3 & 3.8 & \textbf{3.8} & \textbf{4.3} & 3.4 & 3.9 & {3.5} & {4.0} \\
Cora           & 3.5 & 4.0 & \textbf{3.9} & \textbf{4.5} & 3.6 & 4.1 & {3.7} & {4.2} \\
\bottomrule
\end{tabular}}
\end{table}
}

\section{Prompts}
\label{app:prompt}
\subsection{Prompt Templates of Bipartite Graphs}

\begin{table}[H]
  \caption{The prompt template for node description in bipartite graphs (e.g. Sephora).}
  \label{prompts:sephora_node}
  \begin{tcolorbox}[colback=blue!5, 
  colframe=black,
  ]
\textbf{Node (Sephora User):}
Author ID: \{node\_id\}, Skin Tone: \{skin\_tone\}, Eye Color: \{eye\_color\}, Skin Type: \{skin\_type\}, Hair Color: \{hair\_color\}, Total Negative Feedback Count: \{total\_neg\_feedback\_count\}, Total Positive Feedback Count: \{total\_pos\_feedback\_count\}

\textbf{Node (Sephora Product):}
  Product ID: \{node\_id\}
  Product Name: \{product\_name\}
  Brand Name: \{brand\_name\}
  Primary Category: \{primary\_category\}
  Secondary Category: \{secondary\_category\}
  Ingredients of the Product: \{ingredients\}
  Loves Count of the Product: \{loves\_count\}
  Product Rating: \{rating\}
  Number of the Product Reviews: \{reviews\}
  Size of the Product: \{size\}
  Product Price (USD): \{price\_usd\}

  \end{tcolorbox}
  \end{table}

\begin{table}[H]
  \caption{The prompt template for edge description in bipartite graphs (e.g. Sephora).}
  \label{prompts:sephora_edge}
  \begin{tcolorbox}[colback=blue!5, 
  colframe=black,
  ]
\textbf{Edge (Sephora Review):}
  Review Time (yyyy-mm-dd): \{timestamp\}
  Rating: \{rating\}
  Review Title: \{review\_title\}
  Review Text: \{review\_text\}
  Total Negative Feedback Count: \{total\_neg\_feedback\_count\}
  Total Positive Feedback Count: \{total\_pos\_feedback\_count\}  
  \end{tcolorbox}
  \end{table}

\begin{table}[H]
  \caption{The prompt template for memory reflection in bipartite graphs (e.g. Sephora).}
  \label{prompts:sephora_memory}
  \begin{tcolorbox}[colback=blue!5, 
  colframe=black,
  ]
  As a customer of the Sephora online shopping platform, you can review Sephora products based on the provided information and your own situation:
  \{node\_info\}

  Here are your previous review history:
  \{node\_memory\}

  Now, based on your personal description and past review history, progressively refine your memory into a concise version, ensuring it reflects your personal preferences. 
  
  Respond.

  \end{tcolorbox}
  \end{table}

\begin{table}[H]
  \caption{The prompt template for edge generation in bipartite graphs (e.g. Sephora).}
  \label{prompts:sephora_interaction}
  \begin{tcolorbox}[colback=blue!5, 
  colframe=black,
  ]
  \textbf{Query:}
  
  You are a customer of the Sephora online shopping platform, you can review Sephora products based on the provided information and your own situation:
  \{node\_info\}

  Here's your past reviews history:
  \{node\_memory\}

  FIRST, you should Search for candidate prodcuts using the provided tools. 
  
  Respond.

\textbf{Action: }

  You are a customer of the Sephora online shopping platform, you can review Sephora products based on the provided information and your own situation:
  \{node\_info\}

  Here's your past reviews history:
  \{node\_memory\}

  Here's the candidate prodcuts you can review:
  \{node\_items\}

  Here's the example of how you should proceed with your review:
  \{interaction\_example\}
  
  You should review ONE product. You can review the chosen product with detailed text and rate it. 
  Additionally, you should predict how many positive/negative feedbacks will be received for this review. 
  The predicted time of the review should be firmly related to the time in your past reviews history (relatively later than or equal to them).
  Respond using the following detailed JSON format for ONE product:
  
  \{ 
  
  \quad "review":\{
    
        \quad \quad item\_id: (str, "The ID of the product you want to review. Be sure to be one of the Item IDs mentioned above!"),
        
        \quad \quad timestamp: (str, "The time of review (yyyy-mm-dd)"),
        
        \quad \quad rating: (int, "The overall rating given to the product (From 1 to 5)"),
        
        \quad \quad review\_title: (str, "The title of your review"),
        
        \quad \quad review\_text: (str, "Your detailed review text"),
        
        \quad \quad total\_neg\_feedback\_count: (int, "Number of negative feedback received for this review"), 
        
        \quad \quad total\_pos\_feedback\_count: (int, "Number of positive feedback received for this review")
        
    \quad \}
    
    \}
  
  Respond.

  \end{tcolorbox}
  \end{table}

\begin{table}[H]
  \caption{The prompt template for node generation in bipartite graphs (e.g. Sephora).}
  \label{prompts:sephora_node_generation}
  \begin{tcolorbox}[colback=blue!5, 
  colframe=black,
  ]
  \textbf{Node (Sephora Author):}
  
  Now we have a Sephora dataset, which records Sephora users' reviews on several Sephora products.
  Here's information of the recent active Sephora author nodes:
  \{recent\_node\_info\}

  You are expected to generate ONE new Sephora author node for the Sephora dataset, and ensure that the generated new node is somewhat different from the existing nodes.
  Respond using the following detailed JSON format for ONE new Sephora author:
  
  \{
  
    \quad "sephora\_author":\{
    
        \quad \quad node\_id: (str, "The ID of the generated author (Format: G + 6-digit random number)"),
        
        \quad \quad node\_type: sephora\_author,
        
        \quad \quad skin\_tone: (str, "The skin tone of the generated author (e.g. light, fair, mediumTan, tan, olive, etc.)"),
        
        \quad \quad eye\_color: (str, "The eye color of the generated author (e.g. brown, green, hazel, blue, etc.)"),
        
        \quad \quad skin\_type: (str, "The skin type of the generated author (e.g. oily, dry, combination, normal, etc.)"),
        
        \quad \quad hair\_color: (str, "The hair color of the generated author (e.g. brown, black, blonde, auburn, etc.)"),
        
        \quad \quad total\_neg\_feedback\_count: (int, "The number of total negative feedback received from other authors of the generated author"), 
        
        \quad \quad total\_pos\_feedback\_count: (int, "The number of total active feedback received from other authors of the generated author"),
        
    \quad \}
  
  \}
  
  Respond.

  \textbf{Node (Sephora Product):}
  
  Now we have a Sephora dataset, which records Sephora users' reviews on several Sephora products.
  Here's information of the recent active Sephora product nodes:
  \{recent\_node\_info\}

  You are expected to generate ONE new Sephora product node for the Sephora dataset, and ensure that the generated new node is somewhat different from the existing nodes.
  Respond using the following detailed JSON format for ONE new Sephora product:
  
  \{
  
    \quad "sephora\_product":\{
    
        \quad \quad node\_id: (str, "The ID of the generated product (Format: G + 5-digit random number)"),
        
        \quad \quad node\_type: sephora\_product,
        
        \quad \quad product\_name: (str, "The name of the generated product"),
        
        \quad \quad brand\_name: (str, "The name of the brand of the generated product"),
        
        \quad \quad primary\_category: (str, "The primary category of the generated product"),
        
        \quad \quad secondary\_category: (str, "The secondary category of the generated product"),
        
        \quad \quad ingredients: (str, "The ingredients of the generated product"), 
        
        \quad \quad loves\_count: (int, "The loves count from the users of the generated product"),
        
        \quad \quad rating: (float, "The avg rating from the users of the generated product"),
        
        \quad \quad reviews: (int, "The number reviews from the users of the generated product"),
        
        \quad \quad size: (str, "The size the generated product"),
        
        \quad \quad price\_usd: (float, "The price the generated product"),
        
    \quad \}
    
    \}
  
  Respond.

  \end{tcolorbox}
  \end{table}

\subsection{Prompt Templates of Non-bipartite Graphs}

\begin{table}[H]
  \caption{The prompt template for node description in non-bipartite graphs (e.g. WeiboTech).}
  \label{prompts:weibotech_node}
  \begin{tcolorbox}[colback=blue!5, 
  colframe=black,
  ]
\textbf{Node (Weibo User):}
  User ID: \{node\_id\}
  User Name: \{user\_name\}
  User Source: \{user\_source\}
  User Gender: \{user\_gender\}
  User Location: \{user\_location\}
  Number of the User's Followers: \{user\_followers\}
  Number of the User's Followees: \{user\_friends\}
  User Description: \{user\_description\}

  \end{tcolorbox}
  \end{table}
  
\begin{table}[H]
  \caption{The prompt template for edge description in non-bipartite graphs (e.g. WeiboTech).}
  \label{prompts:weibotech_edge}
  \begin{tcolorbox}[colback=blue!5, 
  colframe=black,
  ]
\textbf{Edge (Weibo Interact):}
  Interaction Time (yyyy-mm-dd hh-mm-ss): \{timestamp\}
  Interaction Type: \{label\}
  Source User Text: \{src\_text\}
  Destination User Text: \{dst\_text\}
  \end{tcolorbox}
  \end{table}
  
\begin{table}[H]
  \caption{The prompt template for memory reflection in non-bipartite graphs (e.g. WeiboTech).}
  \label{prompts:weibotech_memory}
  \begin{tcolorbox}[colback=blue!5, 
  colframe=black,
  ]
  As a Weibo user, you can search for other Weibo users who may interact with you on the online social media Weibo platform:
  \{node\_info\}

  Here are your previous interactions history:
  \{node\_memory\}

  Now, based on your personal description and past interactions history, progressively refine your memory into a concise version, ensuring it reflects your personal preferences. 
  
  Respond.

  \end{tcolorbox}
  \end{table}

\begin{table}[H]
  \caption{The prompt template for edge generation in non-bipartite graphs (e.g. WeiboTech).}
  \label{prompts:weibotech_interaction}
  \begin{tcolorbox}[colback=blue!5, 
  colframe=black,
  ]
\textbf{Query:}

  As a Weibo user, you need to search for other Weibo users who may interact with you on the online social media Weibo platform:
  \{node\_info\}

  Here are your previous interactions history:
  \{node\_memory\}

  First, utilize the provided tools to search for potential Weibo users who may interact with you.
  
  Respond.

\textbf{Action:}

  As a Weibo user, you need to search for other Weibo users who may interact with you on the online social media Weibo platform:
  \{node\_info\}

  Here are the potential Weibo users you can choose from:
  \{node\_items\}

  Here is your previous interaction history:
  \{node\_memory\}

  Here's the example of how to interact with others:
  \{interaction\_example\}
  
  You should select ONE destination user, you're tend to select the one you have interacted with before. Respond using the following detailed JSON format:
  
  \{
    
    \quad "interact":\{
    
        \quad \quad item\_id: (str, "The ID of the destination user. Be sure to be one of the Item IDs mentioned above!")
        
    \quad \}
    
  \}

  Respond.

\textbf{Request:}

  As a Weibo user, you need to search for other Weibo users who may interact with you on the online social media Weibo platform:
  \{node\_info\}

  Here is your previous interaction history:
  \{node\_memory\}

  The chosen destination Weibo user who may interact with you is:
  \{item\_info\}

  Here's the interaction history of this chosen Weibo user:
  \{item\_memory\}

  Here's the example of how to interact with others:
  \{interaction\_example\}

  You should select ONE destination user. You should post texts as the source user and the selected destination user should interact with you in detailed text. 
  Additionally, you should label the type of the interaction (comment or repost).
  The predicted time of the interaction should be firmly related to the time in your previous interaction history (relatively later than or equal to them).
  Respond using the following detailed JSON format:
  
  \{
  
    \quad "interact":\{
    
        \quad \quad item\_id: (str, "The ID of the destination user"),
        
        \quad \quad timestamp: (str, "The time of the interaction (yyyy-mm-dd hh-mm-ss)"),
        
        \quad \quad label: (str, "The type of interaction (TWO TYPE: 1.comment, 2.repost)"),
        
        \quad \quad src\_text: (str, "The text from the source user"),
        
        \quad \quad dst\_text: (str, "The text from the destination user")
        
    \quad \}    
    
  \}

  Respond.

  \end{tcolorbox}
  \end{table}

\begin{table}[H]
  \caption{The prompt template for node generation in non-bipartite graphs (e.g. WeiboTech).}
  \label{prompts:weibotech_node_generation}
  \begin{tcolorbox}[colback=blue!5, 
  colframe=black,
  ]
\textbf{Node (Weibo User):}

  Now we have a weibo dataset, which records the interaction history between Weibo users.
  Here's information of the recent active user nodes:
  \{recent\_node\_info\}

  You are expected to generate ONE new user node for the weibo dataset, and ensure that the generated new node is somewhat different from the existing nodes.
  Respond using the following detailed JSON format for ONE new user:
  
  \{
  
    \quad "weibo\_user":\{
    
        \quad \quad node\_id: (str, "The ID of the generated user (Format: G + 5-digit random number)"),
        
        \quad \quad node\_type: weibo\_user,
        
        \quad \quad user\_name: (str, "The name of the generated user"),
        
        \quad \quad user\_source: (str, "The source(IP/location/device) of the generated user"),
        
        \quad \quad user\_gender: (str, "The gender of the generated user"),
        
        \quad \quad user\_location: (str, "The location of the generated user"),
        
        \quad \quad user\_followers: (int, "The number of the followers of the generated user"),
        
        \quad \quad user\_friends: (int, "The number of the followees of the generated user"),
        
        \quad \quad user\_description: (int, "The description of the generated user"),
        
    \quad \}
    
    \}
  
  Respond.
  \end{tcolorbox}
  \end{table}

{
\subsection{Prompt Templates of LLM-as-Evaluator in Textual Quality Metrics}
\label{app:prompt_llm_eval}

\begin{table}[H]
  \caption{{The prompt template of evaluation criteria for LLM-as-Evaluator.}}
  \label{prompts:critique}
  \begin{tcolorbox}[colback=blue!5, 
  colframe=black,
  ]
Please evaluate the role-playing ability of the ACTOR NODE based on its actions with ITEM NODES across multiple turns based on scene, ACTOR NODE information, action history and evaluation criteria.
  \newline
  
  [Environment Description]:
  \{environment\}
  \newline
  
  [ACTOR NODE Description]:
  \{actor\_node\}
\newline

  [Multi-turn Actions]:
  \{actions\}
\newline

  Strict Evaluation Criteria:
  
  1. Factual Accuracy: Identify and point out any elements that do not accurately match the historical or factual backdrop.
  
  2. Behavior Consistency: Explicitly highlight inconsistencies between the actor nodes' actions and their traits.
  
  3. Logical Coherence: Point out any logical fallacies or actions that contradict the established context or logic.
  
  4. Content Redundancy: Identify repetitions in actions that could detract from engagement and realism.
  
  5. Emotional Expression: Assess whether emotional responses and expressions are appropriate and convincingly portrayed, highlighting any discrepancies.
  
  6. Interaction Adaptability: Critique the actor node's interactions with item nodes, noting any unnatural or contextually inappropriate responses.
  
  7. Creativity and Originality: Evaluate the creativity of responses and actions, pointing out generic or unoriginal content.
  
  8. Detail Handling: Scrutinize the level of detail in scene setting and actor node enactment, marking areas lacking depth or accuracy.
  
  9. Style Consistency: Ensure that the linguistic style remains consistent, identifying any deviations.
  
  10. Fluency and Quality: Critically assess the smoothness and quality of the text, highlighting any grammatical errors or awkward phrasings.

  \end{tcolorbox}
  \end{table}
  
\begin{table}[H]
  \caption{{The prompt template of scoring criteria for LLM-as-Evaluator.}}
  \label{prompts:critique}
  \begin{tcolorbox}[colback=blue!5, 
  colframe=black,
  ]  
  [Scoring Criteria]:
  
  1. Contextual Fidelity:
  
    - 1 Point: Responses are often incorrect or irrelevant, significantly inconsistent with the actor node's background.
    
    - 3 Points: Responses are generally accurate, though there are occasional errors or some details are not very relevant to the actor node's background.
    
    - 5 Points: Responses are always accurate and highly relevant to the actor node's historical or professional background, demonstrating deep knowledge and skills.
    \newline
    
  2. Personality Depth:  
  
    - 1 Point: The displayed personality traits often conflict with or lack consistency with the actor node's setup.
    
    - 3 Points: Personality traits generally match the actor node's setup, though there are occasional inconsistencies.
    
    - 5 Points: Consistently displays behavior and language choices that match the core personality traits of the actor node, showcasing the actor node's uniqueness.
\newline

  3. Dynamic Adaptability:  
  
    - 1 Point: The actor node struggles to adapt to new scenarios over time, producing rigid or illogical responses.  
    
    - 3 Points: Adaptation occurs in most situations but occasionally falters in handling unexpected turns or maintaining coherence.  
    
    - 5 Points: The actor node fluidly adapts to novel contexts and introduces innovative solutions and maintains consistency even under challenging conditions.
\newline

  4. Immersive Quality:  
  
    - 1 Point: Actor node portrayal is often inconsistent, making it difficult for users to immerse or understand the actor node.
    
    - 3 Points: Actor node is mostly consistent, but occasional contradictions slightly affect immersion.
    
    - 5 Points: Actor node portrayal is always consistent, enhancing user immersion and effectively showing self-awareness and actor node limitations.
\newline

  5. Content Richness:  
  
    - 1 Point: Output is sparse, lacking depth, detail, or meaningful interaction with item nodes. 
    
    - 3 Points: Content is adequate but occasionally superficial, missing opportunities for nuanced or layered interactions.  
    
    - 5 Points: Output is exceptionally dense with relevant details, creative ideas, and meaningful engagement with item nodes. It showcases a high level of expertise and creativity, providing users with an abundance of valuable information and interactive elements that significantly enhance the overall experience.

  \end{tcolorbox}
  \end{table}

\begin{table}[H]
  \caption{{The prompt template of evaluation steps and response formats for LLM-as-Evaluator.}}
  \label{prompts:critique}
  \begin{tcolorbox}[colback=blue!5, 
  colframe=black,
  ]     
  [Evaluation Steps]:
  
  1. Contextual Understanding: Examine the actor node's profile and background information thoroughly to fully grasp the nuances of their context, motivations, and historical background.
  
  2. Behavioral Observation: Monitor how the actor node reacts across different scenarios, paying special attention to their decisions and interactions.
  
  3. Criteria-Based Assessment: Strictly analyze each observed behavior using the above criteria to systematically evaluate the consistency and effectiveness of the actor node's portrayal. 
  \newline
  
  Your response must follow the format provided below. Please note that only when the content quality is extremely good can 5 Points be given.
\newline

  [Response Format]:
  
  Contextual Fidelity: [1-5] 
  
  Personality Depth: [1-5]  
  
  Dynamic Adaptability: [1-5]  
  
  Immersive Quality: [1-5]  
  
  Content Richness: [1-5]  
\newline

  [Response Format Example]:
  
  Contextual Fidelity: 3
  
  Personality Depth: 3
  
  Dynamic Adaptability: 3
  
  Immersive Quality: 3  
  
  Content Richness: 3 
  \newline

  [Response]:

  \end{tcolorbox}
  \end{table}
  
\begin{table}[H]
  \caption{{The prompt template of dataset descriptions for LLM-as-Evaluator.}}
  \label{prompts:datasetdescription}
  \begin{tcolorbox}[colback=blue!5, 
  colframe=black,
  ]
  \textbf{Sephora:}

  General Description: This is a dataset recording users' reviews of products on the Sephora e-commerce platform, including information about each user's age, skin type, eye color, information about each product, and users' reviews of each product.
  Main Actors: The users who review products on the Sephora e-commerce platform.
  Main Activities of Actors: Sephora users review various products, rating products, and providing additional user experiences.
  
\textbf{Dianping:} 

  General Description: This is a dataset recording users' reviews of businesses on the Dianping platform, including information about each user, information about each business, and users' reviews of each business.
  Main Actors: The users who review businesses on the Dianping platform.
  Main Activities of Actors: Dianping users review various businesses, rating businesses, and providing additional user consumption experiences.
  
  \textbf{WikiRevision:}
  
  General Description: This is a dataset recording users' revisions on Wikipedia pages, including information about each wiki page and comments submitted by users during revision.
  Main Actors: The users who revise the wiki pages.
  Main Activities of Actors: Wiki users revise various wiki pages (correcting errors in wiki pages, adding new content, removing redundant content, and summarizing the wiki pages) with a comment or brief summary in text.
  
    \textbf{WikiLife:}
    
    General Description: This is a dataset recording the life trajectories of several people (celebrities), including basic information about each person, information about the location of each life trajectory, and the events happened at those locations for each person.
  Main Actors: The people who physically present in some locations for various categories of life trajectories.
  Main Activities of Actors: People (celebrities) physically present in the specified location for various life trajectories, such as being born, studying, working, getting married, having children, and passing away.

\textbf{IMDB:}
  
  General Description: This is a dataset recording the collaboration relationships of movie actors/actresses, including information about each movie actor/actress, information about the movies they collaborated on, and the roles they played in those movies.
  Main Actors: The movie actors/actresses who collaborate with each other.
  Main Activities of Actors: Movie actors/actresses collaborate with others and playing different roles in movies.

\textbf{WeiboTech/WeiboDaily:}
  
  General Description: This is a dataset recording users' interaction on the Weibo platform, including information about each user and the text information attached during user interactions such as comments and reposts.
  Main Actors: The users who interact with each other on the Weibo platform.
  Main Activities of Actors: Weibo users make multiple posts, and other users interacting with them by commenting or reposting the posts.

\textbf{Cora:}

  General Description: This is a dataset recording paper citation relationships, including information about each paper and the specific text attached during citation in a specific section of the paper.
  Main Actors: The academic papers which cite other related papers.
  Main Activities of Actors: Academic papers cite other related papers with specific citation sentences in the text.

  \end{tcolorbox}
  \end{table}
}

\section{Use of Large Language Models (LLMs)}
LLMs were used in two specific aspects of this work. 
First, we employ LLMs to polish the manuscript text and refine figure captions for clarity and presentation quality. 
Second, our proposed generative framework, GAG-General, is built upon an LLM-based multi-agent architecture that coordinates multiple specialized agents to generate structurally and semantically coherent DyTAGs. 
The LLMs in this framework are used to model textual attributes and support agent reasoning during graph generation. 
No other parts of the research, including problem formulation, method design, or analysis, involved significant LLM assistance.

\end{document}